\documentclass[10pt,journal,compsoc]{IEEEtran}

\ifCLASSOPTIONcompsoc
  \usepackage[nocompress]{cite}
\else
  \usepackage{cite}
\fi
\ifCLASSINFOpdf
\else
\fi
\usepackage{amsmath,amsfonts}
\usepackage{array}
\usepackage{textcomp}
\usepackage{stfloats}
\usepackage{url}
\usepackage{verbatim}
\usepackage{animate}
\usepackage{multirow}
\usepackage[utf8]{inputenc}
\usepackage{mathtools}
\usepackage{amsmath}
\usepackage{array}
\usepackage{graphicx} 
\usepackage{subfigure} 
\usepackage{amssymb}
\usepackage{amsmath}
\usepackage{color}
\usepackage{lipsum}
\usepackage{booktabs}
\usepackage{amsmath}
\usepackage{adjustbox}
\usepackage{etoolbox}

\usepackage{lipsum,xcolor}
\newcommand\dunderline[3][-1pt]{{%
  \sbox0{#3}%
  \ooalign{\copy0\cr\rule[\dimexpr#1-#2\relax]{\wd0}{#2}}}}
  
\usepackage{amsfonts}
\usepackage{float}
\usepackage{color}
\usepackage{multirow}

\usepackage{float}
\usepackage{caption}
\usepackage{hyperref}

\hypersetup{
  colorlinks=true, 
  linkcolor=blue, 
  citecolor=blue, 
  filecolor=magenta, 
  urlcolor=blue, 
  pdftitle={Your Title}, 
  pdfpagemode=UseNone, 
  pdfborder={0 0 0} 
}

\usepackage{algorithm}%
\usepackage{algorithmicx}%
\usepackage{algpseudocode}%
\usepackage{listings}%

\renewcommand{\algorithmicrequire}{\textbf{Input}}

\newcounter{note}

\usepackage{ragged2e}

\begin{document}
%

\title{\huge Fine-Tuning Adaptive Stochastic Optimizers: Determining the Optimal Hyperparameter $\epsilon$ via Gradient Magnitude Histogram Analysis}

\author{Gustavo~Silva, ~and~
        Paul~Rodriguez
\IEEEcompsocitemizethanks{\IEEEcompsocthanksitem Gustavo Silva and Paul Rodriguez are with the Department of Electrical Engineering, Pontificia Universidad Cat\'olica del Per\'u, Lima, Per\'u. \\
E-mails: gustavo.silva@pucp.edu.pe and prodrig@pucp.edu.pe.}
}

\IEEEtitleabstractindextext{%
\begin{abstract}
\justifying{
Stochastic optimizers play a crucial role in the successful training of deep neural network models. To achieve optimal model performance, designers must  carefully select both model and optimizer hyperparameters. However, this process is frequently demanding in terms of computational resources and processing time. While it is a well-established practice to tune the entire set of  optimizer hyperparameters for peak performance, there is still a lack of clarity regarding the individual influence of hyperparameters  mislabeled as  “low priority”, including the safeguard factor $\epsilon$ and decay rate $\beta$, in leading adaptive stochastic optimizers like the Adam optimizer.  In this manuscript, we introduce a new framework based on the empirical probability density function of the loss' gradient magnitude, termed as the “gradient magnitude histogram”, for a thorough analysis of adaptive stochastic optimizers and the safeguard hyperparameter $\epsilon$. This framework reveals and justifies valuable relationships and dependencies among hyperparameters in connection to optimal performance across diverse tasks, such as classification, language modeling and machine translation. Furthermore, we propose a novel algorithm  using  gradient magnitude histograms to automatically estimate a refined and accurate search space for the optimal safeguard hyperparameter $\epsilon$, surpassing the conventional trial-and-error methodology by establishing a worst-case search space that is two times narrower.}
\end{abstract}

}

\maketitle

\IEEEdisplaynontitleabstractindextext

%
\IEEEpeerreviewmaketitle

\IEEEraisesectionheading{\section{Introduction}\label{sec:introduction}}

\begin{figure*}[h!]
\centering
\includegraphics[width=0.95\textwidth]{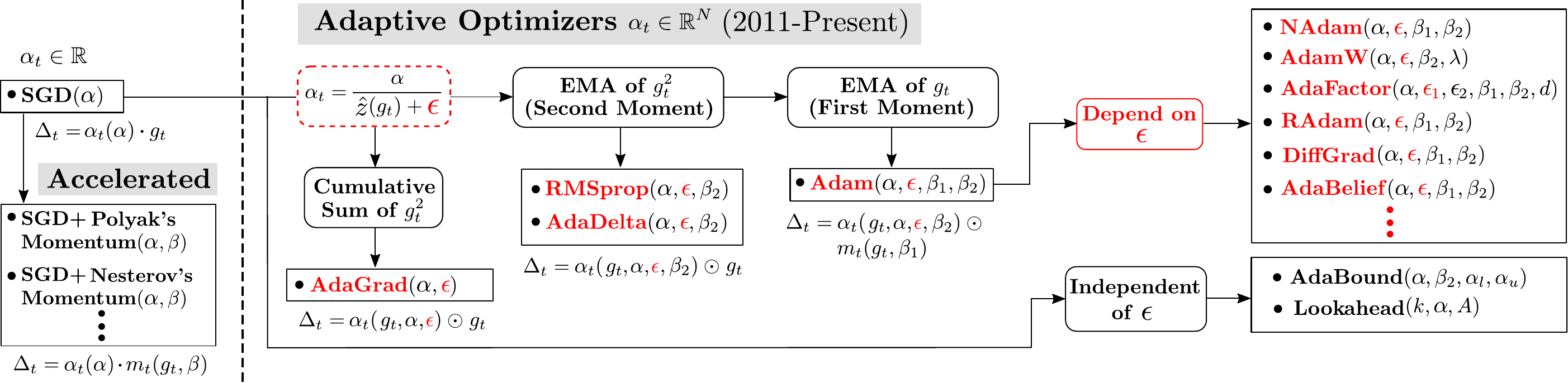} 
\vspace{1mm}
\caption{Taxonomy of stochastic optimizers; those containing the safeguard hyperparameter $\epsilon$ in their formulation are highlighted in red. This hyperparameter is used to avoid division overflow issues. The descent step in Eq. \ref{eq:updateeq} is simplified as $\theta_{t+1} = \theta_{t} - \Delta_t$.
\label{subfig:Taxonomy} }
\end{figure*}

\IEEEPARstart{I}{n} the last decade, neural networks (NN) have emerged as an exceptional framework for multiple real-world applications, such as  image classification \cite{he2016deep}, language processing \cite{manning2014stanford}, autonomous systems \cite{mnih2015human}, and so on. As NN models have become deeper and more complex, optimization algorithms, commonly known as optimizers, have continuously evolved to ensure the effective training of these sophisticated models. 
Stochastic gradient descent-based
optimizers \cite{bottou2012stochastic},  classified as accelerated and adaptive variants, iteratively minimize an objective function by updating parameters in the opposite direction of the gradient. Specifically, the iterative update step of adaptive stochastic optimizers can be expressed as:

\begin{equation}
\theta_{t+1} = \theta_{t} - \alpha_t(\alpha, hyper)\odot  {m}_t (g_t, hyper), 
\label{eq:updateeq}
\end{equation}

\noindent where $\alpha_t(\cdot) \in \mathbb{R}^N$ denotes the adaptive learning rate modified at each time step $t$, and $\alpha = \alpha_0$ serves as the initial learning rate hyperparameter. Here, $g_t$  represents a noisy approximation of the loss function's gradient, while  $m_t(\cdot) \in \mathbb{R}^N$ is the smoothed gradient approximation obtained via the application of a momentum or moving average technique.  Unlike the stochastic gradient descent (SGD) and its classical accelerated variants, adaptive optimizers  involves  a greater number of  hyperparameters. These include learning rate $\alpha$, safeguard factor $\epsilon$, decay rates ($\beta_1$ and $\beta_2$) for the first and second moments, and hyperparameters related to the learning rate schedule (such as decay steps and learning rate decay). Among them, the learning rate $\alpha$ is recognized as an important hyperparameter \cite{Bengio2012, wilson2017marginal} for preventing divergence and achieving good performance. 

\begin{figure}[h!]
\centering
\subfigure[VGG11 - CIFAR10]{\includegraphics[width=0.227\textwidth]{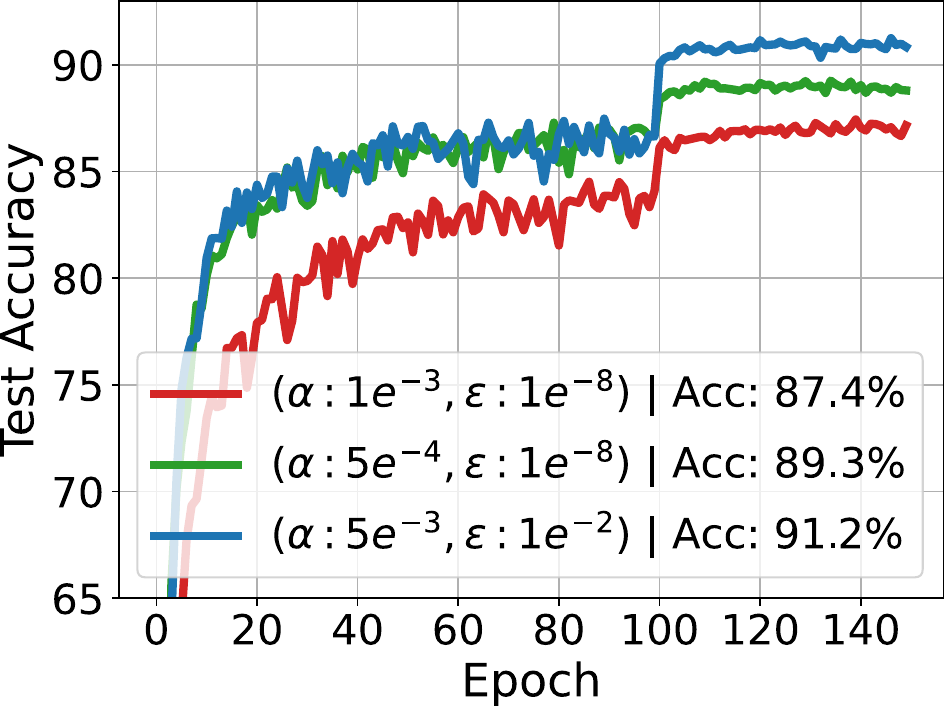} \label{subfig:classif-default1a}}
\subfigure[VGG11 - Tiny ImageNet]{\includegraphics[width=0.227\textwidth]{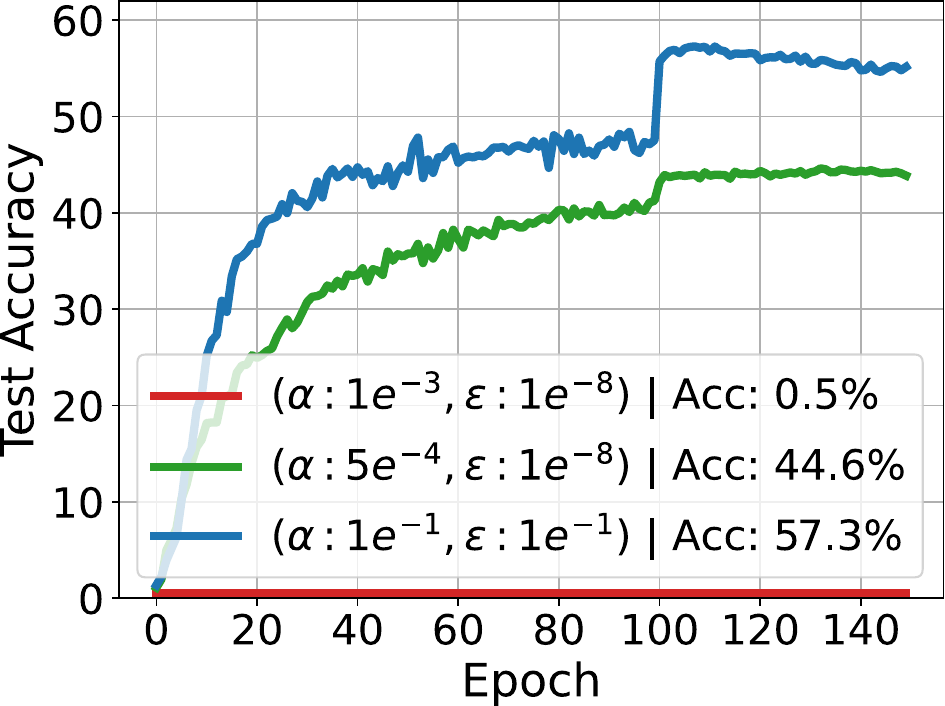} \label{subfig:classif-default1b}} 
\caption{Adam optimizer performance using default hyperparameters ($\alpha= 1e^{-3}$ and $\epsilon= 1e^{-8}$) versus tuned hyperparameters for training the VGG11 model on CIFAR-10 and Tiny ImageNet datasets.\label{subfig:classif-default}}
\vspace{-5mm}
\end{figure}

In recent years, there has been increased attention on exploring additional hyperparameters, such as the decay
rates  ($\beta_1$ and $\beta_2$) and  safeguard factor $\epsilon$, as a means to enhance practical performance.  Traditionally, $\epsilon$ is incorporated into most adaptive stochastic optimizers (as depicted in Fig. \ref{subfig:Taxonomy}) to prevent division overflow during the computation of the adaptive learning rate:

\begin{equation}
\alpha_t = \frac{\alpha}{\hat{z}(g_t) + \epsilon } , \label{Eq1-lr}   
\end{equation}

\noindent where $\hat{z}(\cdot)$ represents a weighted moving average function. However,  beyond its conventional role, this hyperparameter $\epsilon$ also controls spatial adaptability, allowing performance maximization (as evidenced by the blue curves in Fig. \ref{subfig:classif-default}).
In \cite{choi2019empirical}, supported by the outcomes from other studies  
\cite{szegedy2016rethinking,tan2019mnasnet,tan2019efficientnet, zaheer2018adaptive,liu2019variance}, the authors emphasized the importance of accurately selecting the safeguard hyperparameter $\epsilon$, whose optimal value can differ significantly from the default value depending on the deep learning problem. Furthermore, empirical findings reported in \cite{choi2019empirical} suggest that when fine-tuning the complete set of optimizer hyperparameters, adaptive stochastic optimizers can provide equivalent or superior effectiveness compared to  basic and accelerated SGD methods.
Unfortunately, the search process for such hyperparameters, not only involve a recursive trial-and-error methodology to define a useful and accurate search space but also entails significant computational effort, even more when dealing with large-scale problems. 

\begin{figure}[h!]
\centering
\includegraphics[width=0.4\textwidth]{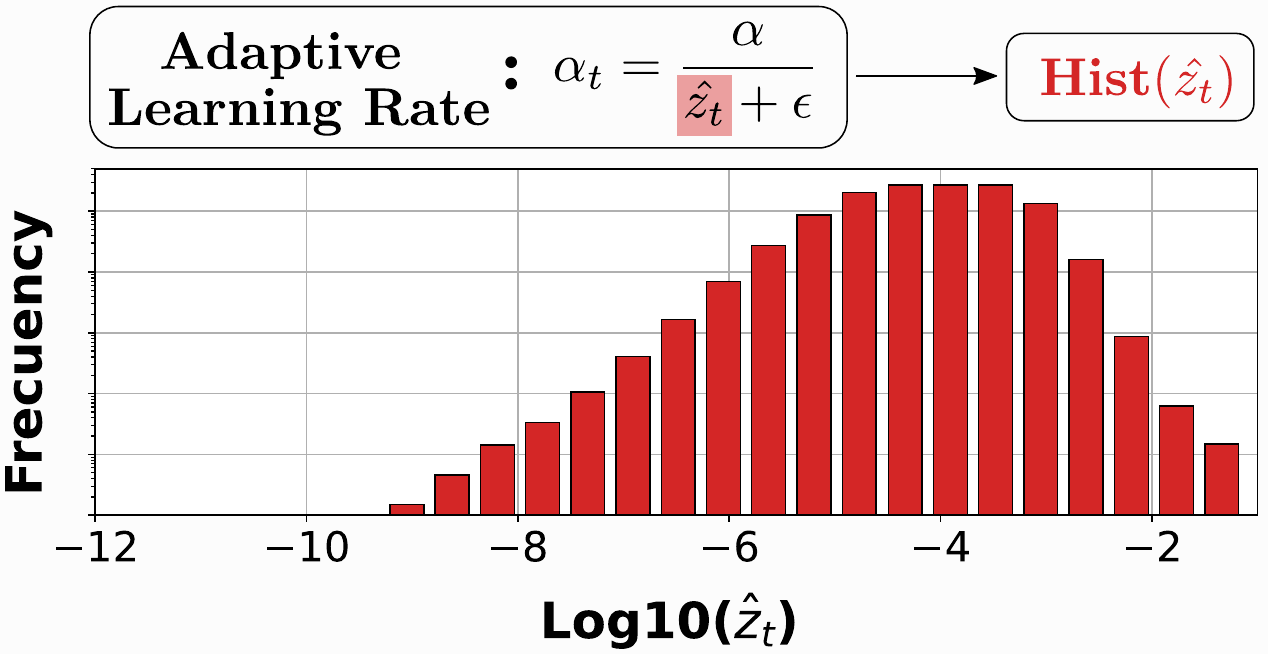} 
\vspace{1mm}
\caption{Computing the gradient magnitude histogram for adaptive stochastic optimizers that depend on safeguard hyperparater $\epsilon$.\label{subfig:AlgoHist} }
\end{figure}

In comparison with \cite{choi2019empirical}, we introduce a novel framework based on the empirical probability density function of the loss' gradient magnitude, referred to as the “gradient magnitude histogram” for a thorough analysis of the individual-level effects of the safeguard hyperparameter $\epsilon$ on adaptive stochastic optimizers. Furthermore, this framework enables the automation of the process for determining the optimal value of the hyperparameter $\epsilon$. At a high level, the proposed framework consists in evaluating histograms of the element-wise adaptive component $\hat{z}_{t} = \hat{z}(g_t)$ (refer to Fig. \ref{subfig:AlgoHist}) in stochastic optimizers\footnote{For a wide range of adaptive stochastic optimizers, the specific mathematical formulations for $\hat{z}_{t}$ are summarized in Table \ref{tab:FullCases} of Appendix \ref{appendixA}. For instance, in the RMSprop optimizer, $\hat{z}_t = \sqrt{v_{t}}$, where $v_{t}$ denotes the second-order momentum.} to examine the influence of the safeguard hyperparameter $\epsilon$ on adaptability (additive relation between $\hat{z}_{t}$ and $\epsilon$). For example, if $\epsilon = 1\times10^{-2}$ in the Fig. \ref{subfig:AlgoHist}, there would be a low level of adaptability, as the value of the safeguard hyperparameter $\epsilon$ predominates over the majority of elements in $\hat{z}_{t}$.

Now, we will outline the specific contributions of this research:
\begin{enumerate}
 \item \textbf{Novel framework for analyzing adaptive stochastic optimizers:} Introducing a pioneering framework grounded in gradient magnitude histograms to analyze adaptive stochastic optimizers, revealing relationships and dependencies among optimizer hyperparameters ($\alpha$, $\epsilon$ and $\beta_2$) linked to optimal performance.

 \item \textbf{Automated algorithm for selecting the safeguard hyperparameter \scalebox{1.35}{$\mathbf{\epsilon}$}:}
 Instead of depending on trial and error methodology or prior knowledge to establish a search space for finding the optimal safeguard hyperparameter $\epsilon$, a new algorithm based on gradient magnitude histograms is developed to automatically determine a reduced and accurate search range for this hyperparameter, thus simplifying the selection of optimal values.

\item \textbf{Reconsideration of evaluation tasks for new adaptive stochastic optimizers:} Providing empirical and theoretical evidence challenging the suitability of well-established tasks, such as image classification, for evaluating new proposals of adaptive stochastic optimizers. The study demonstrates that, when the optimal safeguard hyperparameter $\epsilon$ is set, these adaptive optimizers fundamentally match an SGD-based optimizer with a nearly constant learning rate from the start of training. Consequently, this leads to a comparison of a single optimizer.

\end{enumerate}

The structure of the remaining manuscript is organized as follows: 
Section \ref{Non-adaptiveOptimizersSec}  presents a brief summary of stochastic optimizers and their hyperparameters. Section \ref{sec:proposedframework} introduces  a new framework based on gradient magnitude histograms in the context of adaptive optimizers and their associated safeguard hyperparameter $\epsilon$. Experimental results are reported in Section \ref{sec:results}. Finally,  conclusions are stated in Section \ref{Conclusion}.



\section{Stochastic Optimizers}
\label{Non-adaptiveOptimizersSec}

Let an empirical risk minimization problem with cost function of the form
\begin{align}
    F(\theta) &= \frac{1}{M}  \sum_{k=1}^M f(\theta, x_k), \label{eq:sgd1} \\
    \nabla F(\theta) &= \frac{1}{M}\sum_{k=1}^M \nabla f(\theta, x_k) \label{eq:sgd2}
\end{align} 
where $\theta \in \mathbb{R}^N$ is a set of parameters and $\{x_1, \dots,x_M\}$ is the training data. Gradient descent is a basic optimization algorithm that computes the gradient Eq. \ref{eq:sgd2} of the cost function over the entire dataset. However, it can require 
a substantial computational cost for large dataset, where the number of data samples ($M$) falls within  the range of million or billions. Instead, stochastic gradient descent (SGD) \cite{robbins1951stochastic} randomly selects a small portion of dataset at each iteration to estimate the gradient with which parameter update is performed, i.e.
\begin{align}
     g = \nabla \widetilde{F}(\theta) &= \frac{1}{|\mathcal{B}|}  \sum_{k \in \mathcal{B}} \nabla f(\theta, x_k) \label{eq-grad-minibatch}
\end{align}
where $\mathcal{B} \subset \{1, \dots,M\}$ is a subset of data (called mini-batch) with size of $|\mathcal{B}| \ll M $. 
The convergence of this last optimization algorithm can be improved by incorporating Polyak's momentum (heavy ball method) \cite{qian1999momentum}, Nesterov’s accelerated gradient \cite{nesterov1983method} or a combination of both methods. The choice of method depends on the type of optimization problem, deterministic or stochastic. The SGD+Momentum optimizer can be expressed as:
\begin{align}
&b_t = \mu \cdot b_{t-1} + (1-\gamma) \cdot g_{t} \label{eq:Mom1}\\ 
&\theta_{t+1} = \theta_{t} - \alpha_t \cdot b_t \label{eq:Mom2}
\end{align}
\noindent where $g_t =  \nabla \widetilde{F}(\theta_{t})$, see Eq. \ref{eq-grad-minibatch}, is the estimated gradient from a mini-batch at time step $t$, $\mu$ is the momentum hyperparameter with typical value $\mu = 0.9$, $\gamma$ is the dampening hyperparameter and $\alpha_t=\alpha$ is a single constant learning rate. If the momentum and dampening hyperparameters are equal ($\mu = \gamma$), then  Eq. \ref{eq:Mom1} reduces to the exponential moving average (EMA) of the gradients, generally contained in state-of-the-art adaptive stochastic optimizers. Other accelerated algorithms include aggregated momentum \cite{lucas2018aggregated}, quasi-hyperbolic
momentum \cite{ma2018quasi}, SGD-$\ell_\infty$ \cite{rodriguez2023improving}, stochastic two-step \cite{rodriguez2023assessment}, etc.

\subsection{Adaptive Optimizers}
Adaptive optimization algorithms iteratively adjust the learning rate throughout the training process based on gradient information. In contrast to vanilla SGD and its accelerated variants (see previous section), adaptive optimizers can indeed exhibit faster convergence and good generalization ability across a wider range of tasks \cite{dosovitskiy2021an, shen2015deepcontour,keskar2016large}, such as image recognition, image segmentation and natural language processing.

\begin{algorithm}[H]
\caption{Generic framework of adaptive optimization methods}\label{alg:generic}
\begin{algorithmic}[1]
\algrenewcommand\algorithmicrequire{\textbf{Input}}
\Require $\bullet$  Learning rate hyperparameter $\alpha$,    $\bullet$ safeguard hyperparameter $\epsilon$,  $\bullet$ decay rate  hyperparameters  $\beta_1$ and $\beta_2$, and  $\bullet$  sequence of functions $\{\phi_t, \psi_t\}^T_{t=1}$  
\For{$t=1$ to T} 
        \State $m_t = \phi_t(g_1,...,g_t, \beta_1),$
        \State $v_t = \psi_t(g_1,...,g_t, \beta_2), ~~~~ \alpha_t = \alpha \cdot \mathcal{W}(v_t, \epsilon)$
        \State $\theta_{t+1} = \theta_{t} - \alpha_t \odot {m}_t $
\EndFor
\end{algorithmic}
\end{algorithm}

Several attractive adaptive optimizers, including AdaGrad \cite{duchi2011adaptive}, RMSprop \cite{tieleman2012lecture}, Adam\cite{kingma2014adam}, AdaBelief \cite{zhuang2020adabelief}, and others, can be synthesized using the generic structure presented in Algorithm \ref{alg:generic}. 
In this structure, $\phi(\cdot)$ and $\psi(\cdot)$ commonly represent  exponential moving averaging functions, $\mathcal{W}(\cdot,\epsilon)$ defined as $\frac{1}{\sqrt{\cdot + \epsilon} }$ or $\frac{1}{\sqrt{\cdot} + \epsilon}$ is an essential element-wise adaptive component, where the original purpose of $\epsilon$ is to prevent division overflow, and $\alpha_t$ is the adaptive learning rate.

In particular, the hyperparameters for this optimizer family can be categorized into two groups: those related to gradient direction and those related to adaptive learning rates, that are linked to $m_t$ and $\alpha_t$, respectively (as depicted in lines 2 and 3 of the Algorithm \ref{alg:generic}). For Adam-based optimizers (as highlighted in Fig. \ref{subfig:Taxonomy}), estimation of gradient direction is performed from an exponential moving average, as pointed out in Eq.  \ref{eq:firstMoment}, 
where $\beta_1$ is the exponential decay rate for the first moment, reducing noise variance in gradients at each time step.

\begin{equation}
 m_t = \beta_1 \cdot m_{t-1} + (1-\beta_1)\cdot g_t \label{eq:firstMoment}
\end{equation}

The adaptive learning rate in these optimizers is primarily governed by three key hyperparameters: learning rate $\alpha$,  decay rate hyperparameter $\beta_2$, and safeguard factor $\epsilon$.
The first hyperparameter $\alpha$ is a positive scalar value that controls the length of step by which the model parameters are updated. In convex landscapes, a smaller or larger learning rate hyperparameter generates a slower or faster convergence, respectively. Nevertheless, choosing larger learning rates can lead to situations where the optimization process overshoots and oscillates around the optimal solution. 
While manual tuning of  the learning rate hyperparameter is necessary to enhance performance, as indicated in \cite{wilson2017marginal, smith2019super}, from a theoretical point of view  \cite{robbins1971convergence}, decreasing the learning rate is required to ensure the convergence of any SGD variant \cite{robbins1971convergence}. In practice, this reduction is implemented through learning rate schedules, which may follow linear, polynomial, exponential and some other types of adjustments at specific epochs and may integrate additional hyperparameters. Various learning rate schedules have been investigated in studies such as \cite{he2016deep, smith2017cyclical}. 

The second hyperparameter $\beta_2$, also referred to as the exponential decay rate for the second moment: 
\begin{equation}
v_t = \beta_2 \cdot v_{t-1} + (1-\beta_2)\cdot g_t^2, \label{eq:secondMoment}
\end{equation}

\noindent determines the contribution of the exponential moving average of squared gradients. In essence, it defines the amount of previous squared gradient samples that are observed over a window of approximately 1/(1-$\beta_2$) time samples, where the regular value for $\beta_2$ is 0.999.

Finally,  hyperparameter $\epsilon$,  known as safeguard hyperparameter, is a small value  conventionally used to prevent division by zero  in the computation of the adaptive learning rate, as illustrated by the formula\footnote{Specifically, Eq. \ref{eq:stepsize1} is part of the ADAM algorithm. Nonetheless, with some minor algebraic differences, such structure is observed in a large number of algorithms.}:
\begin{equation}
\alpha_t = \frac{\alpha}{\sqrt{\hat{v}_t} + \epsilon }  \label{eq:stepsize1}
\end{equation}
where $\hat{v}_t$ is the bias-corrected estimate of the second moment, given by:
\begin{equation}
\hat{v}_t = \frac{v_t}{(1-\beta^t_2)}.
\end{equation}
\noindent In the case of  RMSprop and Adam optimizers, the default values for the safeguard hyperparameter $\epsilon$ are $10^{-6}$ and $10^{-8}$, respectively. However,  their optimal values in practice  can  differ depending on the characteristics of the problem, optimizer and dataset.
Further details regarding the safeguard hyperparameter are presented  in  Section \ref{sec:safeguardsec}.

\subsection{Safeguard Hyperparameter $\epsilon$}
\label{sec:safeguardsec}

Recently, \cite{choi2019empirical} shows the practical importance of the intrinsic inclusion hierarchy among stochastic optimizers for determining their performance ranking. According to  \cite{choi2019empirical}, if the formulation of a general optimizer integrates another optimizer as its structural basis (forming an inclusion relationship\footnote{For example, the formulation of SGD (base optimizer) is encompassed within the formulation of SGD+Momentum (generic optimizer). This can be verified by setting  $\mu = \gamma = 0$ in SGD+Momentum (see Eq. \ref{eq:Mom1} and Eq.\ref{eq:Mom2}), which results in SGD.}), general optimizers like RMSprop and ADAM never perform worse than basic optimizers, such as SGD and SGD+Momentum, as long as  a certain set of optimizer hyperparameters is carefully configured. However,  \cite{choi2019empirical} also states that obtaining the values of the hyperparameters that enable ADAM to match or surpass the performance of SGD+Momentum might not be easily accessible. Furthermore, this study reiterates the importance of adjusting the safeguard hyperparameter $\epsilon$, often overlooked in practice.

In particular, literature evidence suggests that the default value of safeguard hyperparameter $\epsilon$ can be suboptimal for achieving peak  performance. In fact, the optimal hyperparameter $\epsilon$ can be significantly larger, differing by orders of magnitude from the default setting. For example, in \cite{szegedy2016rethinking}, the authors   selected $\epsilon=1.0$  as the optimal safeguard hyperparameter when training their proposed Inception-V2 model on the ImageNet dataset with the RMSprop optimizer. In the same context but with $\epsilon = 10^{-3}$, \cite{tan2019mnasnet} and \cite{tan2019efficientnet} examined the performance of the MnasNet and EfficientNet models, respectively. 
Given the optimal safeguard value $\epsilon = 10^{-3}$, \cite{zaheer2018adaptive} compared the YOGI and Adam optimizers on the CIFAR-10 dataset for ResNet20, ResNet50 and DenseNet models. In neural machine translation, specifically for a transformer-based model trained on the  WSLT’14 dataset, \cite{liu2019variance} showed that ADAM with $\epsilon = 10^{-4}$ outperformed ADAM with its default value $\epsilon = 10^{-8}$ in terms of convergence rate and  minimum stable loss. For reinforcement learning task, \cite{hessel2018rainbow} employed the Adam optimizer with a tuned safeguard value $\epsilon = 1.5\times 10^{-4}$. 

The hyperparameter $\epsilon$ has been subject to some interpretations beyond its conventional role as a safeguard factor aimed at preventing divisions by zero. For instance, \cite{choi2019empirical} lists two alternative interpretations of this hyperparameter. First, by viewing ADAM as a diagonal approximation of natural gradient descent, $\epsilon$ can function as a damping factor that enhances the condition of Fisher \cite{martens2015optimizing}. Second, the hyperparameter $\epsilon$ can be seen as a trust region radius that regulates the effect of adaptability term $v_t$. In line with this second interpretation, \cite{savarese2021domain} indicates that large $\epsilon$ can result in less adaptability; however, the aforementioned research focused 
on proposing a new adaptive stochastic optimizer 
designed to mitigate sensitivity to the hyperparameter
$\epsilon$. 

In comparison with \cite{choi2019empirical}  and \cite{savarese2021domain}, we will explore both the theoretical and experimental impact of this hyperparameter $\epsilon$ on adaptive  stochastic optimizers. Additionally, we will develop an algorithm for estimating suitable search ranges for this hyperparameter.

\section{A New Perspective based on Gradient Magnitude histograms}
\label{sec:proposedframework}

Over the past few years, there has been a substantial increase in the development of various adaptive stochastic optimizers that depend on the hyperparameter $\epsilon$. Therefore, gaining an in-depth understanding of how this hyperparameter works is undoubtedly highly relevant to the field of machine learning.
In this section, we introduce an innovative framework, based on the gradient magnitude histograms, that enables to conduct an integral analysis of adaptive optimizers and the safeguard hyperparameter  $\epsilon$.  

\begin{figure}[h!]
\centering
\includegraphics[width=0.475\textwidth]{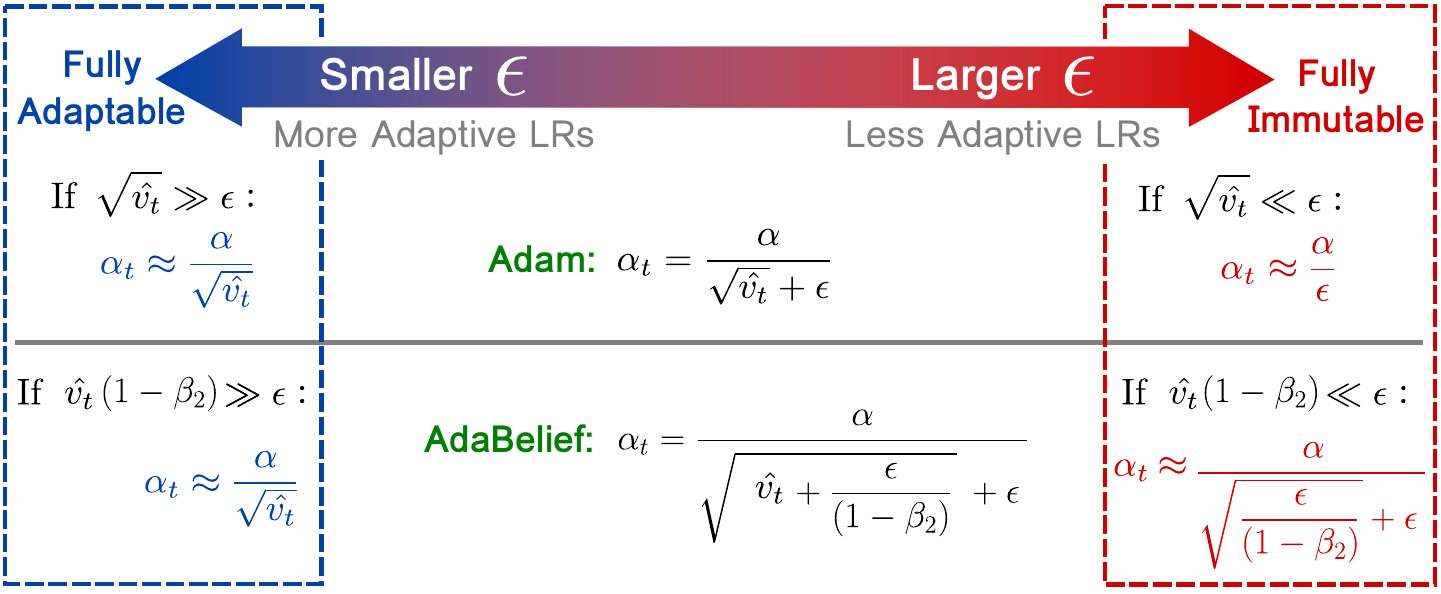} 
\caption{Influence of the safeguard hyperparameter $\epsilon$, renamed as  immutability hyperparameter since it inversely affects adaptability of optimizers. For a comprehensive summary of the two extreme cases (fully adaptable and fully immutable) in some other adaptive stochastic optimizers, please review Table \ref{tab:FullCases} in Appendix \ref{appendixA}.\label{subfig:InfluenceEpsilon}}
\end{figure}

Before delving into our proposed approach, it is essential to establish, as evidenced in Fig. \ref{subfig:InfluenceEpsilon}, that
in the case of an adaptive optimizer like
ADAM optimizer,  where the adaptive learning rate is $\alpha_t = \alpha / (\sqrt{\hat{v}_t} + \epsilon )$ (refer to Eq. \ref{eq:stepsize1}),
a very small $\epsilon$ facilitates the adaptability of $\hat{v}_t$ up to the point of having a fully adaptive learning rate $\alpha_t \approx \alpha/\sqrt{\hat{v}_t}$, where $\epsilon$ is omitted due to its negligible impact. In the opposite situation, where $\epsilon$ is significantly large, the adaptability of $\hat{v}_t$ is attenuated, and in an extreme case, the adaptive learning rate is approximately  a constant value 
$\alpha_t \approx \alpha/\epsilon$.

Taking into account that a large hyperparameter $\epsilon$ mitigates the adaptability offered by $\hat{v}_t$, we now label $\epsilon$ as immutability hyperparameter which can be opposed to adaptability, e.g. smaller immutability means higher adaptability of $\hat{v}_t$ and higher immutability means less adaptability of $\hat{v}_t$.

\begin{figure}[h!]
    \includegraphics[width=0.47\textwidth]{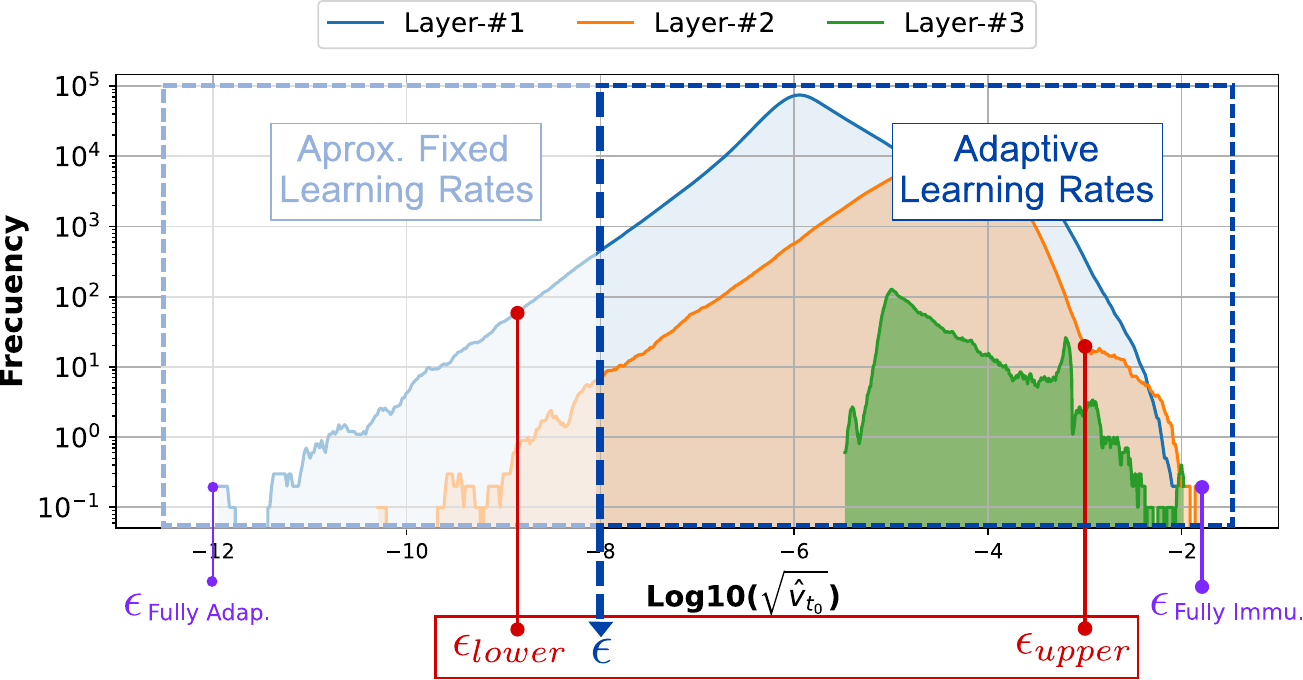} 
\caption{Gradient magnitude histograms for each layer in the 1-Layer LSTM model  at time $t_0$. Each layer can be composed by more than one update parameter such as weights, biases, mean and variance of batch normalization, etc. Assuming an immutability value $\epsilon = 10^{-8}$ (vertical dashed blue line), gradients on the right-side end up generating adaptive learning rates $\alpha_t$, while gradients on the left-side, whose value are lower  than $\epsilon$, results in approximately constant learning rates $\alpha_t$.}\label{subfig:hist_grad} 
\end{figure}



 With the above information in consideration, it seems natural to ask how large or small the immutability hyperparameter $\epsilon$ must be to lie in any of the two extreme cases or to achieve the best optimizer performance. In order to have a reliable measure of this hyperparameter we propose to inspect  the gradient magnitude histograms, that  in particular for the Adam algorithm such histograms are computed from the adaptive term $\sqrt{\hat{v}_t}$. Using the gradient magnitude histograms, we can analyze the number of individual gradients that contribute to the adaptability of the optimizer during the training process. For example, in Fig. \ref{subfig:hist_grad}, we show, at time $t_0$, the gradient magnitude histograms of a 1-Layer LSTM model \cite{ma2015long} trained with the Adam optimizer and a chosen immutability hyperparameter $\epsilon$. 
As can be seen, a large number of gradients are greater than the selected value of $\epsilon$ (vertical dashed blue line), which  produce a large amount of dominant adaptive learning rates, i.e. the optimizer exhibit a high adaptability level at time $t_0$. 
However, in our experimental results, by monitoring the histograms, we also distinguish low and almost zero adaptability levels for Adam-based optimizers.

Furthermore, as detailed in next Section \ref{section:algo-immutability},  from an algorithm based  on the gradient magnitude histogram, we can automatically estimate an accurate  and narrowed search range for the immutability hyperparameter  $\epsilon$ of the adaptive optimizers, simplifying the process of finding the optimal value.

\subsection{Computing Optimal Search Space for the Hyperparameter $\epsilon$}

\label{section:algo-immutability}

Although manual search with a predefined range based on previous knowledge or random search with a broad range are  common methodologies \cite{bergstra2012random}, they have inherent limitations related to computational cost and inaccuracy when applied to new cases. Particularly, the optimal range for the immutability hyperparameter $\epsilon$ can differ by many orders of magnitude depending on the type of tasks, optimizers, models and datasets. Instead of empirically establishing a range for the immutability hyperparameter $\epsilon$ or requiring previous research experience to narrow it down, as reported in \cite{choi2019empirical, zhuang2020adabelief, wang2021rethinking}, we can automatically estimate a reduced search space based on the gradient magnitude histograms during the first epoch. To avoid outliers such as zero gradient values, our proposed algorithm\footnote{Our code is publicly available at the following link: \href{https://github.com/Gustavo-SO15/Hyperparameter-epsilon.git}{https://github.com/Gustavo-SO15/Hyperparameter-epsilon.git}\label{refnotecode}}, described in  Algorithm \ref{alg:adamdist},  computes the upper and lower bounds across all layers from the lowest 2nd-percentile and the highest 98th-percentil of the gradient magnitude histograms to ensure that the natures of adaptive optimizers are approximately fully adaptable and fully immutable. For the Adam optimizer, the gradient magnitude histograms are calculated from $\hat{z_t} = \sqrt{\hat{v_t}}$, where $\hat{z_t}$ can be interpreted as a proxy to estimate the
immutability hyperparameter. For other adaptive optimizers, you can find the $\hat{z_t}$ values summarized in Table \ref{tab:FullCases} of Appendix \ref{appendixA}.

\begin{algorithm}[h!]
\caption{Computing search space for the immutability hyperparameter $\epsilon$.} \label{alg:adamdist}
\begin{algorithmic}
\algrenewcommand\algorithmicrequire{\textbf{Note}}
\algrenewcommand\algorithmicensure{\textbf{Output}}
\Require $\phi_1(\cdot)$ and $\phi_2(\cdot)$ are functions that respectively return the 2nd and 98th percentiles of a histogram of gradients. 
\Ensure $\epsilon_{lower}$ and $\epsilon_{upper}$ 
\For{$t=1$ to T= № of iterations in 1 epoch}
\State $g_{t,k}= \nabla f_{t,k} $ 
\State  $Compute~ v_{t,k} ~accordingly~ Table ~ \ref{tab:FullCases}$
    \For{$k=1$ to K=№ of updated variables}
            \If{$t==T$}
                \State $Compute~ \hat{z}_{t,k} ~accordingly~ Table ~ \ref{tab:FullCases}$
                \State $Compute~ Histogram~ of~  \hat{z}_{t,k} $
                \State $\epsilon_{lower} = min\{\epsilon_{lower}, ~\phi_1(Hist_{\hat{z}_{t,k}} )\}$
                \State $\epsilon_{upper} = max\{\epsilon_{upper}, ~\phi_2(Hist_{\hat{z}_{t,k}} )\}$
            \EndIf
    \EndFor
\EndFor
\State $\epsilon_{lower} = 10^{round(log10(\epsilon_{lower}))}$
\State $\epsilon_{upper} = 10^{round(log10(\epsilon_{upper}))}$
\end{algorithmic}
\end{algorithm}

In addition, there are some supplementary algorithmic details to consider, including:
(i) As the computation of bounds is an iterative procedure, the initial values of  $\epsilon_{lower}$ and $\epsilon_{upper}$ are set to the maximum value for the float32 data type and zero, respectively. (ii) For simplicity and less computational cost, our implementation uses a sort function to find the percentiles (boundaries) instead of directly calculating a histogram function.


\section{Experimental Results} 
\label{sec:results}

\subsection{Simulation Scenario}
\label{subsect-settings}
We conduct three set of experiments
to analyze the performance of adaptive optimizers from their specific gradient magnitude histograms, which are related with the immutability hyperparameter $\epsilon$. Particularly, this new framework based on the gradient magnitude histograms helps to better understand adaptive algorithms (Section \ref{Analysis_optimizers}) and to automatically estimate a precise  and narrowed search space for the mentioned hyperparameter (Section \ref{optimal-search-space}). The experiments are  structured as follows (see also Table \ref{table:2cases}):

\begin{itemize}
\item \textbf{NNs on image classification:} VGG11 \cite{simonyan2014very}, ResNet34 \cite{he2016deep}, DenseNet121 \cite{huang2017densely},  AlexNet \cite{krizhevsky2017imagenet} and Multi Layer Perceptron (MLP) \cite{bishop1995neural} models were trained on the CIFAR-10 \cite{krizhevsky2009learning}, Tiny ImageNet \cite{le2015tiny} and Fashion-MNIST \cite{greeshma2019hyperparameter} datasets. It is important to note that not all datasets were used for training all models. For more details, please refer to Table \ref{table:2cases}. {\color{black} Similar to \cite{zhuang2020adabelief},  we trained the models with a batch size of 128 (except for the Tiny ImageNet dataset, which used a batch size of 32) and a weight decay of $5 \times 10^{-4}$. However, unlike \cite{zhuang2020adabelief}, where the models were trained for 150 epochs, we used fewer epochs\footnote{For the classification task, and within the optimal range of the optimizer’s hyperparameters, the number of epochs per model was chosen to achieve an accuracy gap of approximately $5\%$, while minimizing potential increases in test loss. \label{refnote-epochs} } to avoid overfitting (see Table \ref{table:2cases}). The corresponding accuracy gaps  (difference between training accuracy and test accuracy) and test loss progression are summarized in Fig. \ref{Fig18} of Appendix \ref{appendixA}.}
For CIFAR-10 and Tiny ImageNet, we performed data augmentation that consists of  random horizontal flipping, random cropping and z-score normalization, see code\footref{refnotecode} for numerical details.

\item   \textbf{LSTM on language modeling}: 1-Layer and 2-Layer LSTM models were trained from the Penn TreeBank \cite{marcus1993building} dataset.  {\color{black}In line with the configurations presented in \cite{zhuang2020adabelief},  we  trained the models for 200 epochs with a batch size of 128, a weight decay of $1.2 \times 10^{-6}$, and applied a learning rate decay of 0.1 at the 100-th and 145-th epochs. The resulting test loss, which consistently decreases over the 200 epochs, is detailed in Fig. \ref{Fig19}  and Fig. \ref{Fig20} of Appendix \ref{appendixA}.}

\item  \textbf{Transformer on natural machine translation}:
Analogous to \cite{zhuang2020adabelief}, we employed the fairseq library to train a transformer model \cite{vaswani2017attention} on the IWSTL’14 dataset. The training configurations consisted of 50 epochs, a weight decay of $10^{-4}$, and a decay rate $\beta_2$ of $0.98$. The corresponding plots for this task are only presented in Appendix \ref{appendixA} to improve the readability of the experimental results.
\end{itemize}

\begin{table}[h!]
\centering
\caption{\color{black} Summary of experimental details.} \label{table:2cases}
\renewcommand*{\arraystretch}{1.4}
\footnotesize
\begin{tabular}{cccc}
\hline 
\textbf{Task} & \textbf{Dataset} & \textbf{Architecture} & \textbf{Epochs}\footref{refnote-epochs} \tabularnewline
\hline 
\hline 
\multirow{8}{*}{\begin{tabular}[c]{@{}c@{}}\textbf{Image}\vspace{-1mm} \\ \textbf{Classification}\end{tabular}} & \multirow{4}{*}{CIFAR10} & AlexNet, & 25\tabularnewline
 &  & VGG11, &45\tabularnewline
  &  & ResNet34, &60\tabularnewline
 &  & DenseNet121& 60 \tabularnewline
\cmidrule{2-4}
 & \multirow{2}{*}{Tiny ImageNet} & AlexNet,&25\tabularnewline
 &  & VGG11 &45\tabularnewline
\cmidrule{2-4}
 & \multirow{2}{*}{Fashion-MNIST} & MLP& 25\tabularnewline
 &  & AlexNet &  25\tabularnewline
\hline 
\multirow{2}{*}{\begin{tabular}[c]{@{}c@{}}\textbf{Language}\vspace{-1mm} \\ \textbf{Modeling}\end{tabular}}& \multirow{2}{*}{Penn TreeBank} & 1-Layer LSTM &  200\tabularnewline
  &  & 2-Layer LSTM& 200\tabularnewline
\hline 
\multirow{2}{*}{\begin{tabular}[c]{@{}c@{}}\textbf{Machine}\vspace{-1mm} \\ \textbf{Translation}\end{tabular}}& \multirow{2}{*}{IWSTL’14} & \multirow{2}{*}{Transformer} &  \multirow{2}{*}{50}\tabularnewline
  &  &  &\tabularnewline
\hline 
\end{tabular}
\end{table}

\begin{figure}[h!]
\centering
\subfigure[Classification]{\includegraphics[width=0.228\textwidth]{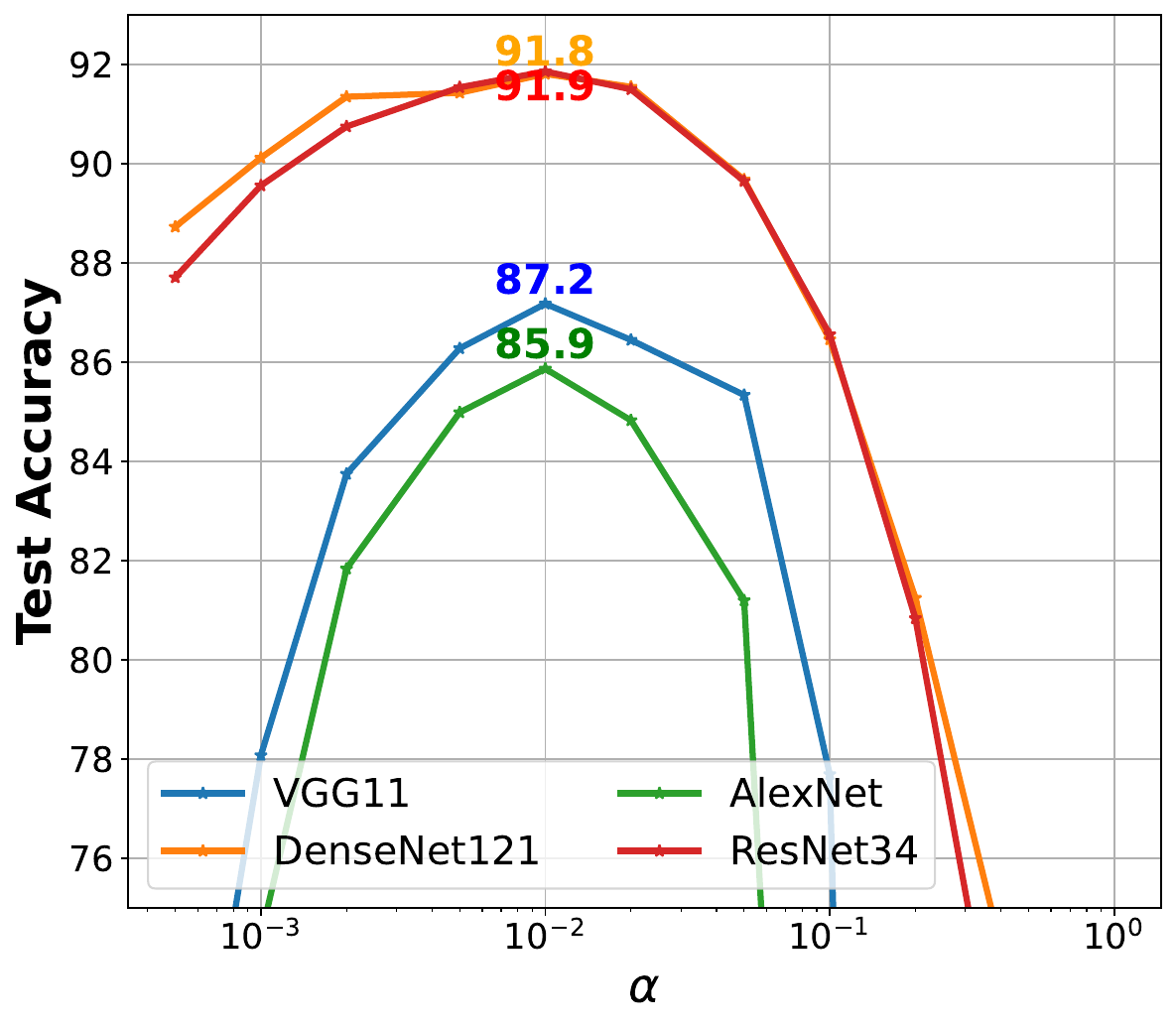} \label{subfig:classif1a-SGD}} \hfill
\subfigure[Language modeling]{\includegraphics[width=0.225\textwidth]{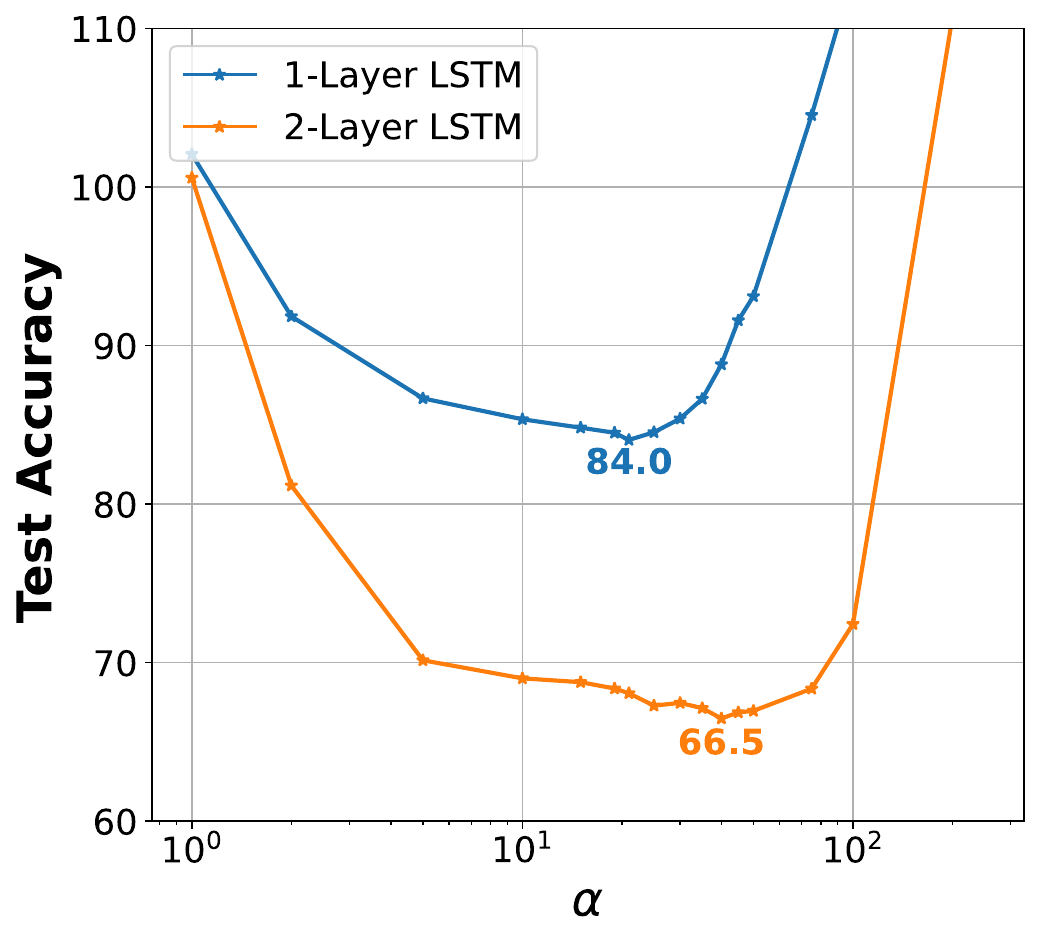} \label{subfig:classif1b-SGD}} 
\caption{Performance of  deep learning models trained on the CIFAR-10 \textbf{(left)} and Penn TreeBank \textbf{(right)} datasets with SGD+Momentum, varying learning rate hyperapareter $\alpha$.  For classification (a) higher accuracy equals better performance, whereas for language modeling (b) lower perplexity equals better performance.} \label{subfig:classif1-SGD}
\vspace{-2.5mm}
\end{figure}

Moreover, for clarity of the experimental analysis, we only examine the Adam optimizer with respect to the SGD+Momentum optimizer. In Fig. \ref{subfig:classif1-SGD}, we show the maximum performance achieved with the SGD+Momentum algorithm, which will be crucial for the final remarks in Section \ref{sec:final-imu}. Evaluation of other adaptive stochastic algorithms, including RSMprop, AdaBelief and AdaMomentum, may be found in Appendix \ref{appendixA}. The optimizer hyperparameters ($\alpha$, $\epsilon$, $\beta_2$) linked to adaptive learning rates $\alpha_t$ are carefully inspected and tuned. Conversely, the decay rate hyperparameter $\beta_1$ for the first momentum, whose impact is not explored in this study, is set to its default value  $\beta_1 = 0.9$. In order establish  a reliable reference for the optimal search range of the Adam optimizer, we performed a grid search over an extensive range, varying the hyperparameters $\alpha$, $\epsilon$ and  $(1-\beta_2)$ in different powers of 10.

\subsection{Analysis of Adaptive Optimizers using Gradient Magnitude Histograms}
\label{Analysis_optimizers}
  
\setcounter{figure}{6} 
\begin{figure}[h!]
\centering
\subfigure[VGG11]{\includegraphics[width=0.227\textwidth]{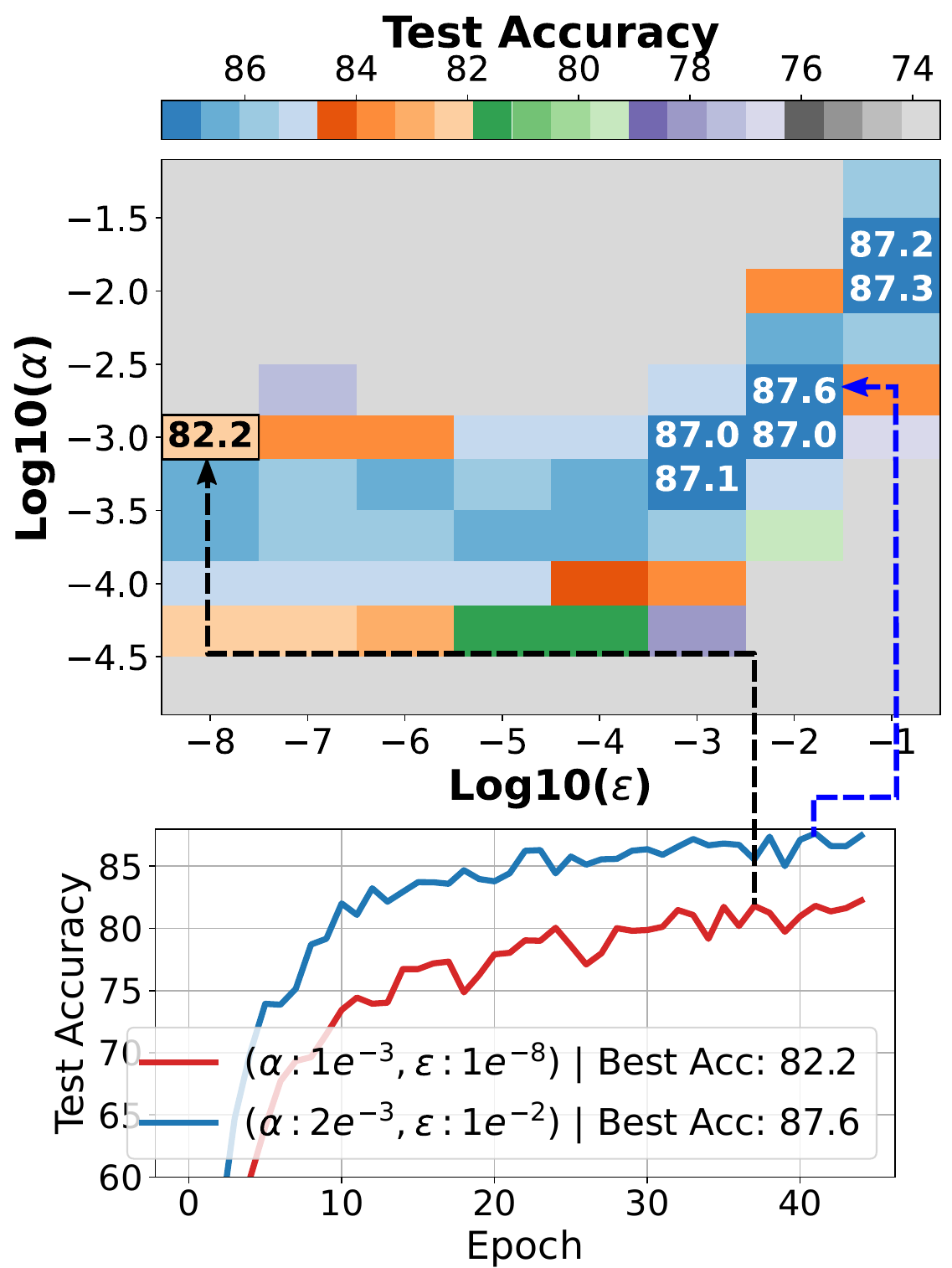} \label{subfig:classif1a}} 
\subfigure[ResNet34]{\includegraphics[width=0.227\textwidth]{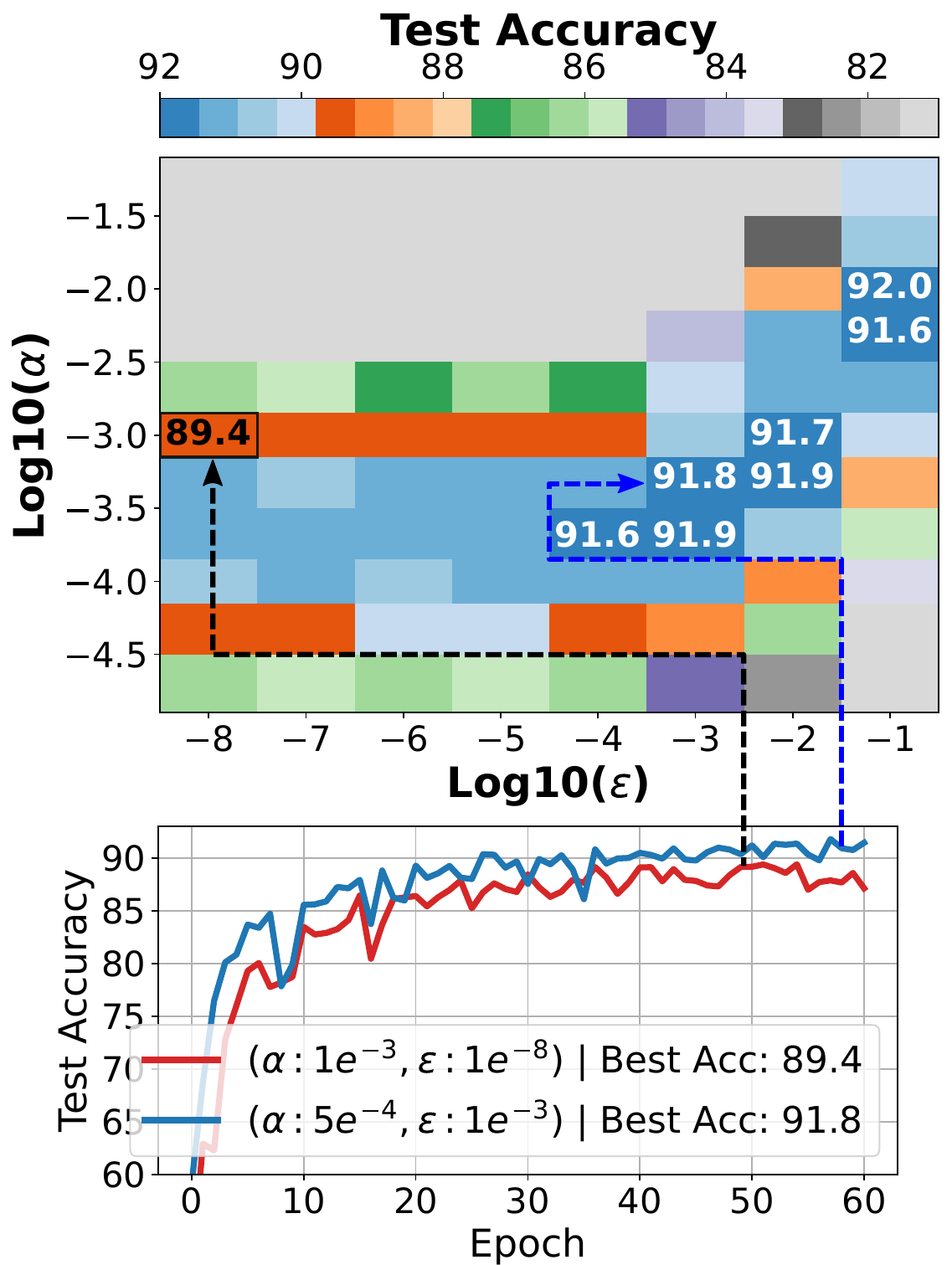} \label{subfig:classif1b}} 
\caption{Classification task: Test accuracy of VGG11 (trained for
45 epochs) and ResNet34 (trained for 60 epochs)
on CIFAR-10 dataset with
Adam optimizer, varying learning rate $\alpha$ and  immutability $\epsilon$. For a given pair of hyperparameter values, each heatmap contains the highest accuracy  achieved across all epochs. \label{subfig:classif1}}
\end{figure}

\begin{figure}[h!]
\centering
\subfigure[1-Layer LSTM]{\includegraphics[width=0.227\textwidth]{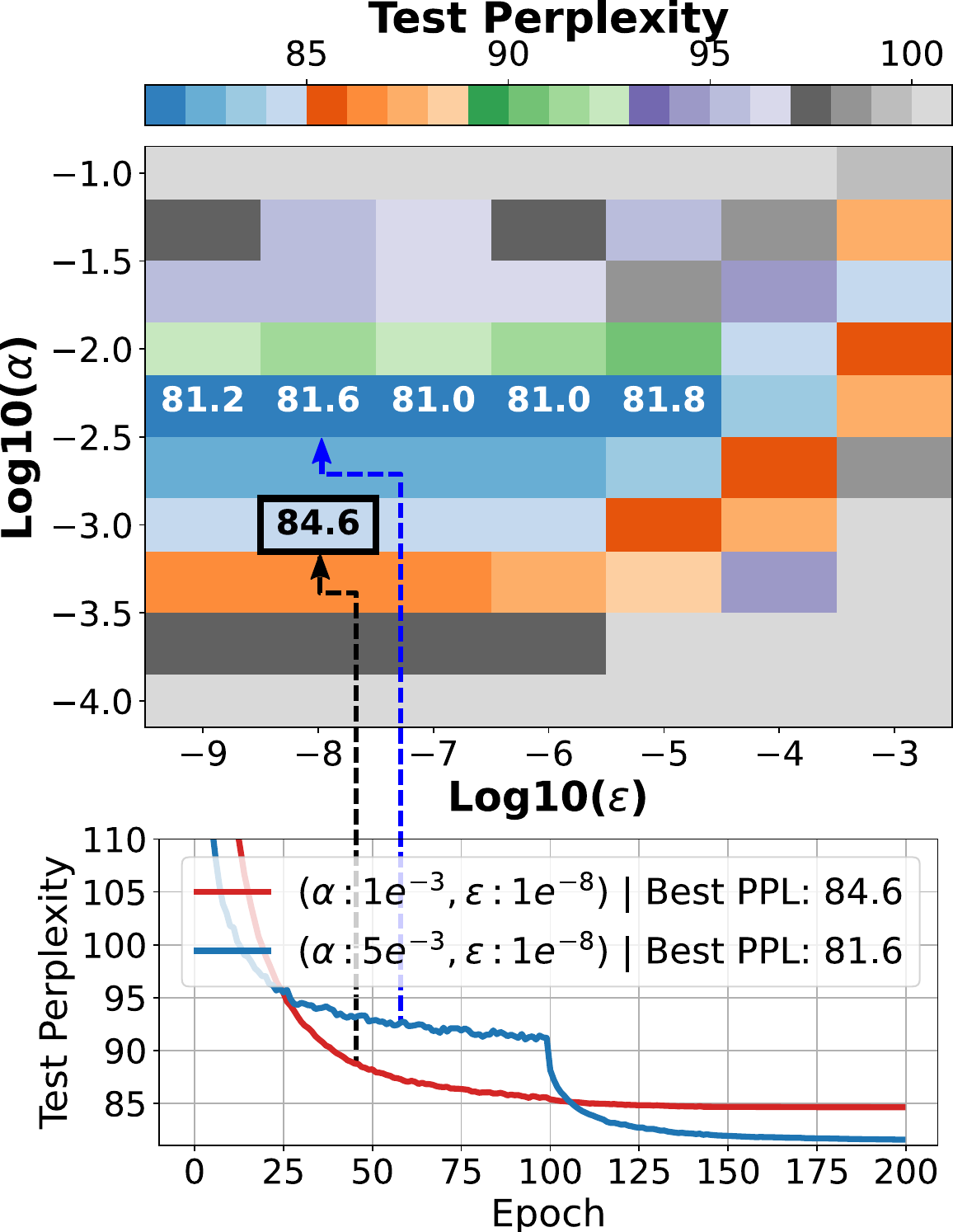} \label{subfig:NLP1a}}  
\subfigure[2-Layer LSTM]{\includegraphics[width=0.227\textwidth]{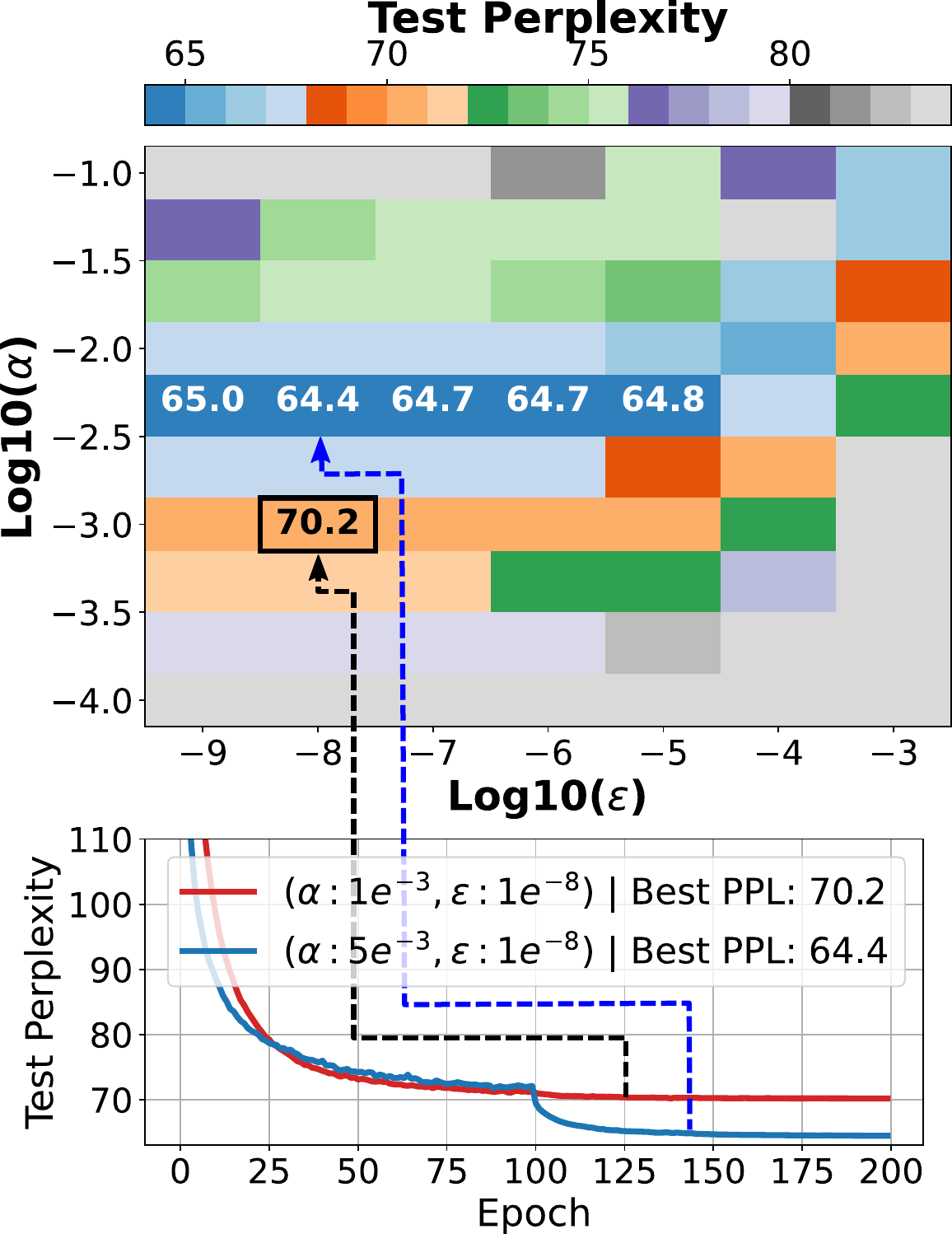} \label{subfig:NLP1b}}  
\caption{Language modeling task: Test perplexity of different LSTM models  (trained for 200 epochs) on the Penn TreeBank dataset with Adam optimizer, varying the learning rate $\alpha$ and  immutability $\epsilon$. For a given pair of hyperparameter values, each heatmap contains the lowest perplexity achieved across all epochs, where better performance is indicated by lower perplexity. \label{subfig:NLP1}}
\end{figure}

As a baseline, for the classification and language modeling tasks, we present the performance of Adam optimizer when varying the learning rate $\alpha$ and immutability $\epsilon$, illustrated in Fig. \ref{subfig:classif1} and Fig. \ref{subfig:NLP1}  in the form of heatmaps.

In both figures, we can observe that there is a subset of optimal learning rate $\alpha_{opt}$ and  optimal immutability $\epsilon_{opt}$ that jointly maximize the optimizer performance, whose optimal immutability value  can vary by orders of magnitude depending on the task. For example, in the Fig. \ref{subfig:classif1} and  $\text{Fig. \ref{subfig:NLP1}}$, the optimal immutability values are $\epsilon_{opt} \geq 10^{-4}$ and $\epsilon_{opt} \leq 10^{-5}$ for classification task and language modeling task, respectively. Based on the information that can be extracted in these figures, it becomes challenging to address questions such as: Why do the optimal values of both hyperparameters ($\alpha$ and $\epsilon$) exhibit different linear relationships across different tasks? Why is a higher or lower value of immutability required to get the best results? Additionally, in Fig. \ref{subfig:beta-eps}, we show the performance obtained when varying decay rate hyperparameter $\beta_2$ in relation to the immutability hyperparameter. This results in different types of patterns between them, which will be explained later in  Section  \ref{sub:influ_beta2}.

\subsubsection{Influence of the immutability hyperparameter $\epsilon$}

\setcounter{figure}{8} 
\begin{figure*}[h]
\centering
\subfigure[VGG11 | From Epoch 1 to Epoch 45]{\includegraphics[width=0.95\textwidth]{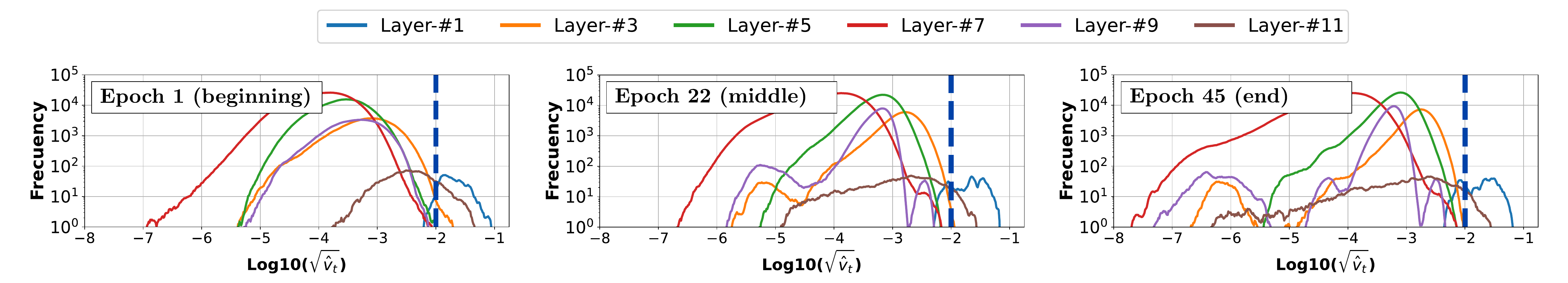} \label{subfig:fig3}}
\\
\subfigure[ResNet34 | From Epoch 1 to Epoch 60]{\includegraphics[width=0.95\textwidth]{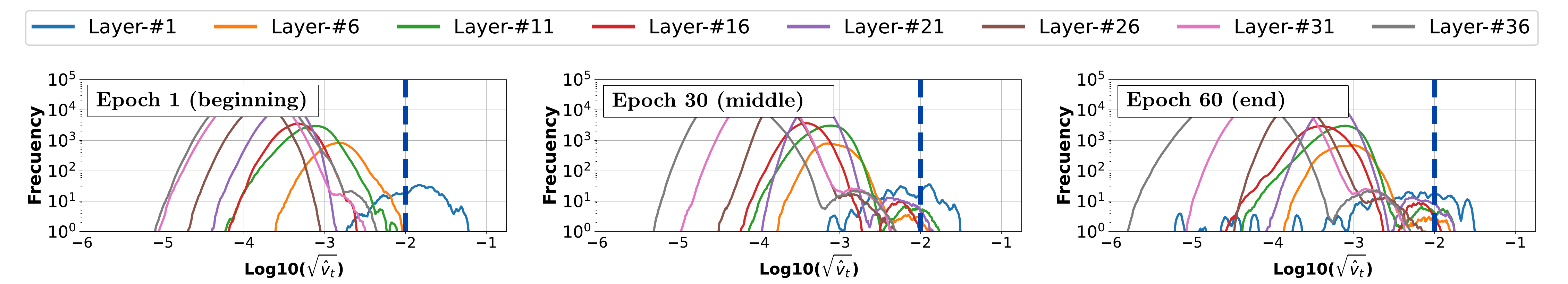} \label{subfig:fig3}}
\caption{Progress of gradient magnitude histograms of the VGG11 \textbf{(top)} and ResNet34 \textbf{(bottom)} classifiers  trained on the CIFAR-10 dataset with Adam and an immutability hyperparameter $\epsilon = 10^{-2}$.  Vertical dashed blue line (chosen immutability hyperparameter $\epsilon$) marks the boundary of discarded elements, where gradients on the left-side are attenuated by $\epsilon$, resulting in approximately constant learning rates $\alpha_t \approx \alpha/ \epsilon$, and gradients on the right-side generate adaptive learning rates.} 
\label{subfig:classif-hist}
\end{figure*}

\setcounter{figure}{9}  
\begin{figure*}[h!]
\centering
\subfigure[1-Layer LSTM | From epoch 1  to epoch 200]{\includegraphics[width=0.95\textwidth]{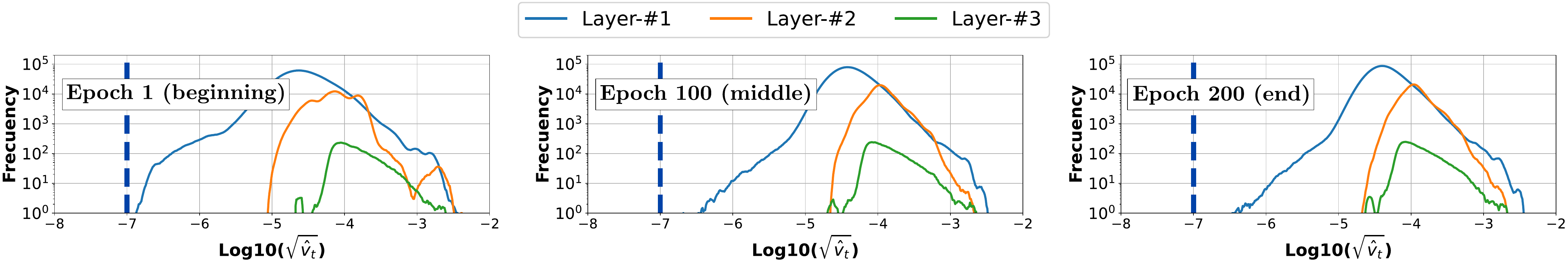} \label{subfig:fig3}}  
\\
\subfigure[2-Layer LSTM | From epoch 1  to epoch 200]{\includegraphics[width=0.95\textwidth]{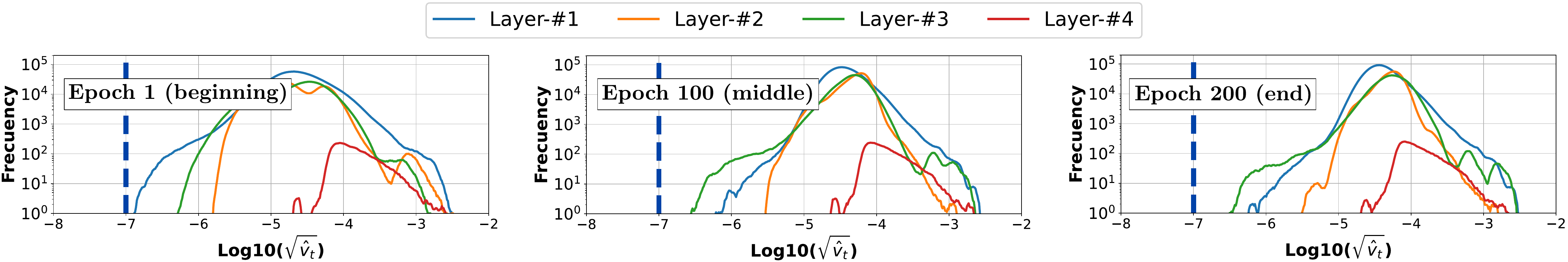} \label{subfig:fig3}}  
\vspace{1mm}
\caption{Progress of gradient magnitude histograms of the 1-Layer LSTM \textbf{(top)} and 2-Layer LSTM \textbf{(bottom)} models  trained on the Penn TreeBank dataset with Adam and an immutability hyperparameter $\epsilon = 10^{-7}$.  Vertical dashed blue line (chosen immutability hyperparameter $\epsilon$) marks the boundary of discarded elements, where gradients on the left-side are attenuated by $\epsilon$, resulting in approximately constant learning rates $\alpha_t \approx \alpha/ \epsilon$, and gradients on the right-side generate adaptive learning rates. }
\label{subfig:NLP-hist}
\end{figure*}

By analyzing the gradient magnitude histograms, presented in Fig. \ref{subfig:classif-hist} and Fig. \ref{subfig:NLP-hist}, we reveal novel and valuable information such as influence of the immutability hyperparameter $\epsilon$ and  justification of the observed relationship between the $\alpha$ and $\epsilon$ hyperparameters, both insights will help to gain an improved comprehension of the behavior of adaptive stochastic optimizers. For the classification task, if an adaptive algorithm like Adam uses a large value of immutability hyperparameter, for example $\epsilon = 10^{-2}$ as in the Fig. \ref{subfig:classif-hist}, along with its corresponding optimal learning rate that guarantee the best performance, from the initial epoch only at most  $0.33\%$ of elements in $\sqrt{\hat{v}_t}$ are greater than selected immutability hyperparameter (vertical dashed blue line), see  Fig. \ref{subfig:classif-hist} and Table \ref{porcentage}, meaning that $99.67\%$ of the elements in the denominator ($\sqrt{\hat{v}_t}+\epsilon$) of the adaptive optimization algorithm are approximately constant, since $\epsilon$ is the dominant term. In other words, given the large optimal value of immutability hyperparameter, the Adam algorithm at the beginning is basically similar to the SGD+Momentum algorithm\footnote{In addition to the provided explanation, the slight structural difference between both optimizers arises from the bias correction on $m_t$, which disappears after a few epochs.} (Eq. \ref{eq:Mom1}, where $\mu = \gamma = \beta_1$) with a near-constant adaptive learning rate $\alpha_t \approx \alpha/ \epsilon$, and it consistently becomes more like SGD+Momentum after a some epochs, as the values of $\hat{v_t}$ get even smaller (evidenced by the evolution of the histograms in the Fig. \ref{subfig:classif-hist}). This pattern is also detected by training other well-known models such as AlexNet, or even using larger datasets such as Tiny ImageNet, where the classification task can be more complex, see also  Fig. \ref{subfig:lr-eps-DenseNet} to Fig.  \ref{subfig:lr-eps-TinyImageNet} in the Appendix \ref{appendixA}. From this new experimental evidence, it is clear that the linear relation between $\alpha$ and $\epsilon$ observed for certain range of optimal values in the Fig. \ref{subfig:classif1} is due to the fact that for said range, the Adam optimizer results to be similar to SGD+Momentum with approximately single learning rate $\alpha_t \approx \alpha/ \epsilon$, where increasing $\alpha$ in the same proportion as $\epsilon$ gives the same  $\alpha_t$.
 {\color{black} Even when training for 150 epochs (see Fig. \ref{Fig21} of Appendix \ref{appendixA}) as in \cite{zhuang2020adabelief} or using early stopping (see Fig. \ref{Fig22} of Appendix \ref{appendixA}), we observe a consistent linear pattern where Adam behaves similarly to SGD+Momentum with an approximately single learning rate.}

\begin{table}
\centering
\caption{Percentage of adaptive values $\sqrt{\hat{v_t}}$ greater than the chosen $\epsilon$ after the first epoch of training using Adam. \label{porcentage}}
\centering
\addtolength{\tabcolsep}{-1pt}
\renewcommand*{\arraystretch}{1.5}
\footnotesize
\begin{tabular}{cccccc}
\hline
\multirow{2.5}{*}{\textbf{Dataset}}                                                 & \multirow{2.5}{*}{\textbf{Model}} & \multicolumn{4}{c}{\textbf{ Hyperparameter \scalebox{1.35}{$\mathbf{\epsilon}$}}}                                                                                       \\ \cmidrule{3-6} 
                                                                                  &                                 & \multicolumn{1}{c}{\textbf{$\mathbf{10^{-5}}$}} & \multicolumn{1}{c}{\textbf{$\mathbf{10^{-4}}$}} & \multicolumn{1}{c}{\textbf{$\mathbf{10^{-3}}$}} & \textbf{$\mathbf{10^{-2}}$} \\ \hline \hline
\multirow{2.5}{*}{\textbf{CIFAR-10}}                                                 & \textbf{VGG11}                  & \multicolumn{1}{c}{96.60\%}             & \multicolumn{1}{c}{61.50\%}             & \multicolumn{1}{c}{4.59\%}             & \multicolumn{1}{c}{0.33\%}                  \\ \cmidrule{2-6} 
                                                                                  & \textbf{ResNet34}            & \multicolumn{1}{c}{100.00\%}             & \multicolumn{1}{c}{78.51\%}              & \multicolumn{1}{c}{1.99\%}              & 0.03\%               
                                                                                        
                                                                               \\ \hline \hline
\multirow{2.5}{*}{\textbf{Dataset}}                                                 & \multirow{2.5}{*}{\textbf{Model}} & \multicolumn{4}{c}{\textbf{Hyperaparameter \scalebox{1.35}{$\mathbf{\epsilon}$}}}                                                                                       \\ \cmidrule{3-6} 
                                                                                  &                                 & \multicolumn{1}{c}{\textbf{$\mathbf{10^{-7}}$}} & \multicolumn{1}{c}{\textbf{$\mathbf{10^{-6}}$}} & \multicolumn{1}{c}{\textbf{$\mathbf{10^{-5}}$}} & \textbf{$\mathbf{10^{-4}}$} \\ \hline \hline
\multirow{2.5}{*}{\textbf{\begin{tabular}[c]{@{}c@{}}Penn \vspace{-1mm}\\ TreeBank\end{tabular}}} & \textbf{1-Layer LSTM}           & \multicolumn{1}{c}{100\%}              & \multicolumn{1}{c}{99.85\%}            & \multicolumn{1}{c}{91.88\%}            & $9.14\%$             \\ \cmidrule{2-6} 
                                                                                  & \textbf{2-Layer LSTM}           & \multicolumn{1}{c}{$100\%$}                & \multicolumn{1}{c}{$99.89\%$}                & \multicolumn{1}{c}{$81.83\%$}                & $3.62\%$               \\ \hline
\end{tabular}
\vspace{-1mm}
\end{table}

 In this work, we additionally examine other adaptive optimizers, such as RMSprop \cite{tieleman2012lecture}, AdaBelief \cite{zhuang2020adabelief} and AdaMomentum \cite{wang2021rethinking}, alongside the reported immutability hyperparameters $\epsilon$ from their respective publications, as presented in Fig. \ref{subfig:lr-eps-AdaBelief} and Fig.\ref{subfig:NLP-optimizer} of Appendix \ref{appendixA}. It is important to highlight that the majority of adaptive stochastic optimizers proposed in the literature, including but not limited to RMSprop \cite{tieleman2012lecture}, DiffGrad \cite{dubey2019diffgrad},  RAdan \cite{liu2019variance}, AdaBelief \cite{zhuang2020adabelief}, AdaMomentum \cite{wang2021rethinking}, have been extensively evaluated in the context of classification tasks, specifically on the CIFAR-10, CIFAR-100 and ImageNet datasets. In the Fig. \ref{subfig:lr-eps-AdaBelief}, for the classification task, it is evident that the reported optimal values of the immutability hyperparameter $\epsilon$ also inhibit the adaptability of the other evaluated optimizers from the first epoch onward. This implies that the other adaptive stochastic optimizers would be equivalent to the SGD+Momentum optimizer from the beginning for the aforementioned task.  Due to this fact, to ensure a fair and meaningful basis for evaluating new adaptive optimizers in future research, while mitigating the risk of misranking among optimizers, we recommend assessing these optimizers in alternative applications where their performance relies on their inherent adaptability. For this purpose, the following language modeling and neural machine translation tasks can be more appropriate options to contemplate.

For language modeling task, plotted in the Fig. \ref{subfig:NLP1}, the optimal values of immutability hyperparameter $\epsilon$ result to be small. By examining the gradient magnitude histograms in the Fig. \ref{subfig:NLP-hist}, it is clear that all adaptive elements in  $\sqrt{\hat{v}_t}$ are greater than the used optimal adaptability hyperparameter $\epsilon = 10^{-7}$ (vertical dashed blue line), producing 100\% adaptive learning rates that contribute to the minimization, in contrast to the near-constant learning rates observed in the previous classification task. In essence, for this new case, the superior performance of the Adam optimizer can be attributed to the full adaptability of its components. Similar patterns are identified for the natural machine translation task in Fig. \ref{subfig:Transformer} of Appendix \ref{appendixA}. Furthermore, for the language modeling task, the largest optimal value of the immutability hyperparameter ($\epsilon \leq 10^{-5}$) is associated with a high percentage of adaptive components (a high level of adaptability), the exact percentages are described in  Table \ref{porcentage}. In this second case, concerning the relationship among hyperparameters, we can conclude that the optimal learning rate hyperparameter is independent of the immutability hyperparameter, as long as the immutability is less than adaptive term $\sqrt{\hat{v}_t}$.

\subsubsection{Influence of the decay rate hyperparameter $\beta_2$}
\label{sub:influ_beta2}

Now, we will briefly examine the impact of the decay rate hyperparameter $\beta_2$ on the two previous tasks (classification and language modeling).
As expected, in the classification task with the optimal learning rate and immutability settings, and while varying the decay rate as depicted in Fig. \ref{subfig:beta-eps1} and \ref{subfig:beta-eps11}, the Adam optimizer demonstrates independence from the decay rate value $\beta_2$. As mentioned earlier, this is because, in this case, the Adam optimizer structurally resembles  an  SGD+Momentum optimizer, where each element of adaptive learning rate  ($\alpha_t \approx \alpha/ \epsilon$) remains unaffected by $v_t$ and $\beta_2$.  Clearly, tuning $\beta_2$ has no discernible impact on this task.
Conversely, for the language modeling task (as shown in Fig. \ref{subfig:beta-eps2}  and \ref{subfig:beta-eps21}), where we know that $\sqrt{\hat{v}_t}>\epsilon$  and adaptability influences on the performance, we can see that within the optimal immutability range, there is an optimal range of decay rate hyperparameter $\beta_2$. This latter range can be easily determined by ensuring that the optimal value of the decay rate hyperparameter $\beta_2$ permits the averaging of all batches of the dataset. This is crucial for obtaining $\sqrt{\hat{v_t}}$, which approximates the gradient distribution across the entire dataset.  For instance, in the Fig. \ref{subfig:beta-eps2}, training is done using approximately 668 batches of the Penn TreeBank dataset, given $N = 1/(1-\beta_2)$, where $N$ is the number of averaged samples, so $668 \leq  1/(1-\beta_2) \implies 0.9985 \leq \beta_2 $ to contemplate all dataset. Furthermore, as stated in \cite{reddi2019convergence}, $\beta_2$ must be greater than $\beta_1$ and sufficiently distant to prevent convergence failure in general stochastic problems. Otherwise, Adam's updates would be those of a sign stochastic gradient method.

\setcounter{figure}{10}  
\begin{figure}[h!]
\centering
\subfigure[VGG11 -  $\{\alpha= 10^{-3}\}$]{\includegraphics[width=0.228\textwidth]{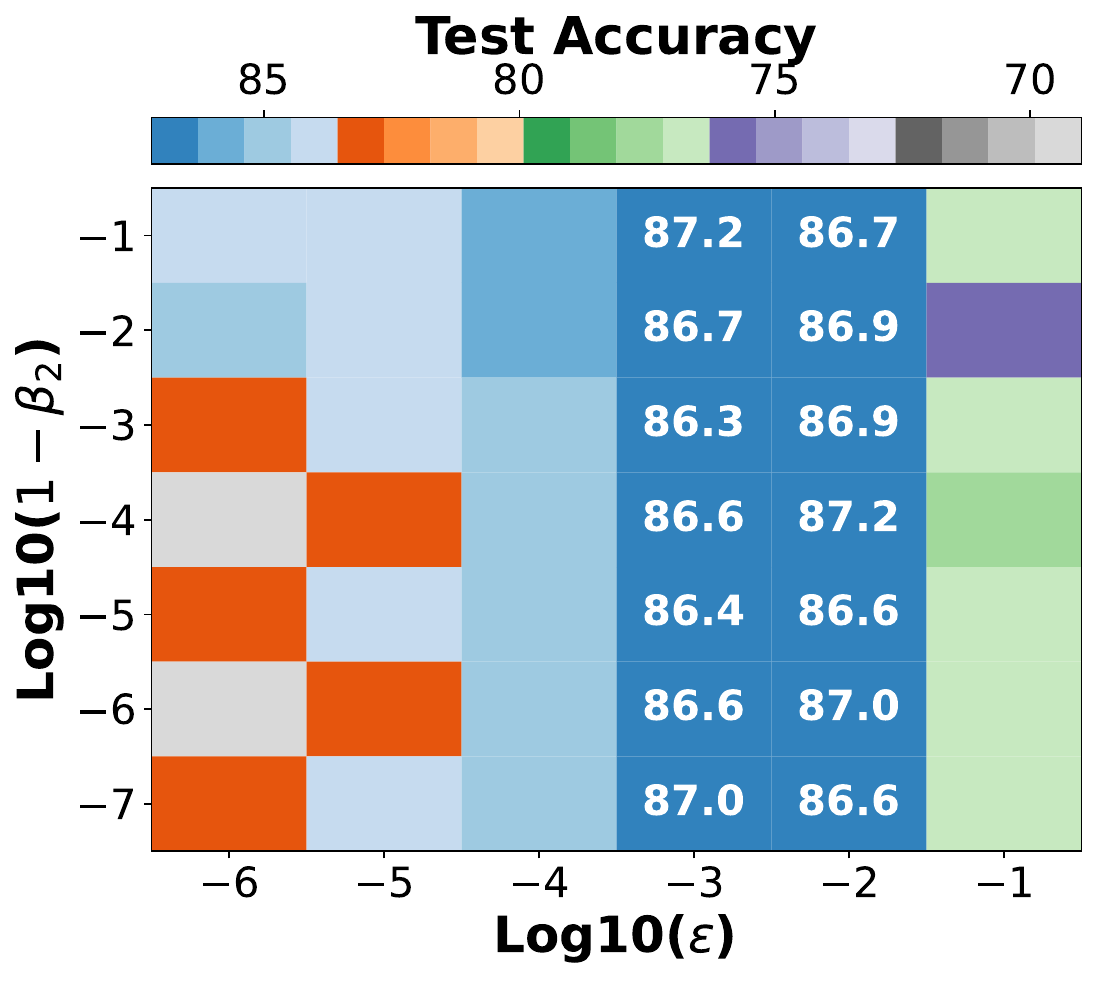} \label{subfig:beta-eps1}}  \hspace{-2mm}
\subfigure[ResNet34 -  $\{\alpha= 5 \times 10^{-4}\}$]{\includegraphics[width=0.228\textwidth]{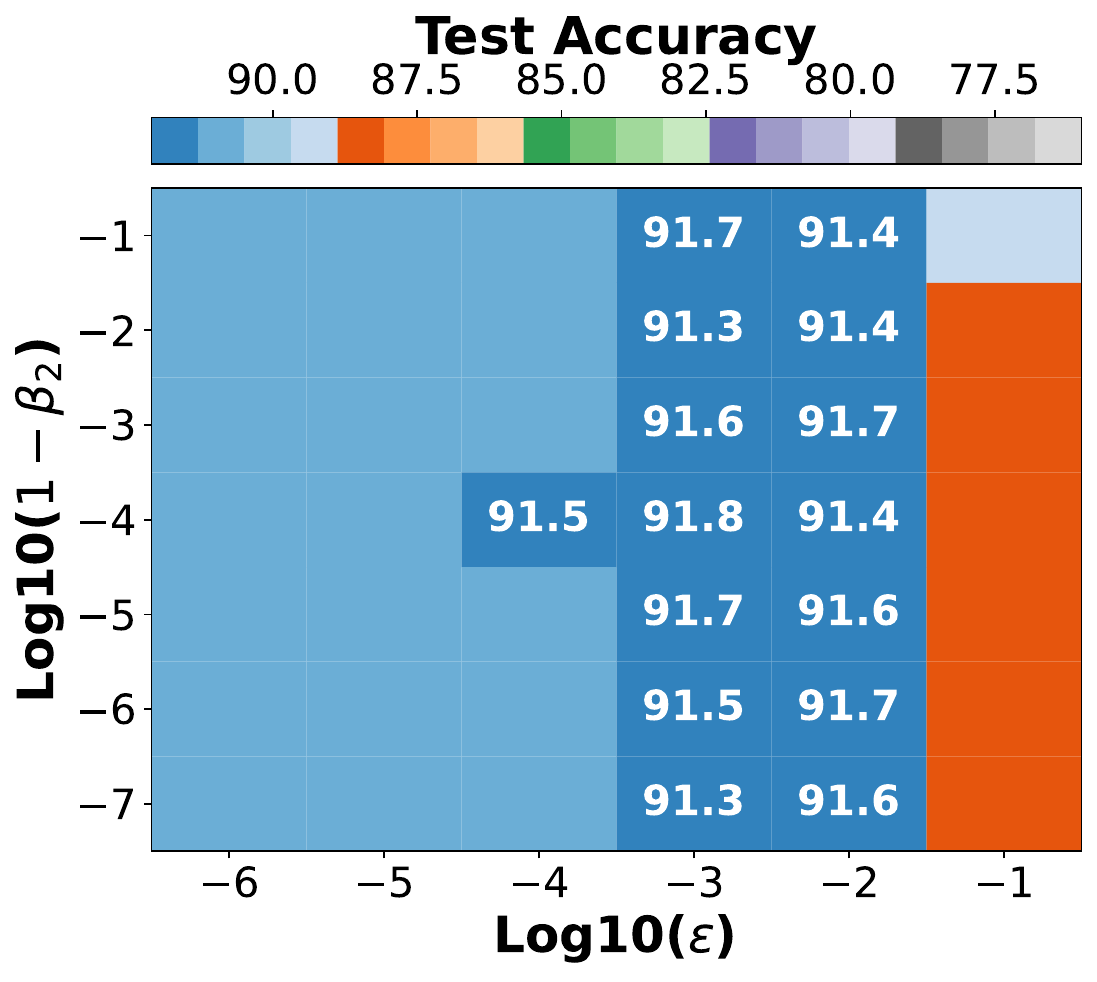} \label{subfig:beta-eps11}}  \\ \vspace{-1mm}
\subfigure[\begin{tabular}{c}
      \vspace{-2mm}\\
     1-Layer LSTM  - \\
     $\{\alpha= 5 \times 10^{-3}\}$  \vspace{-2mm}
  \end{tabular}
]{\includegraphics[width=0.228\textwidth]{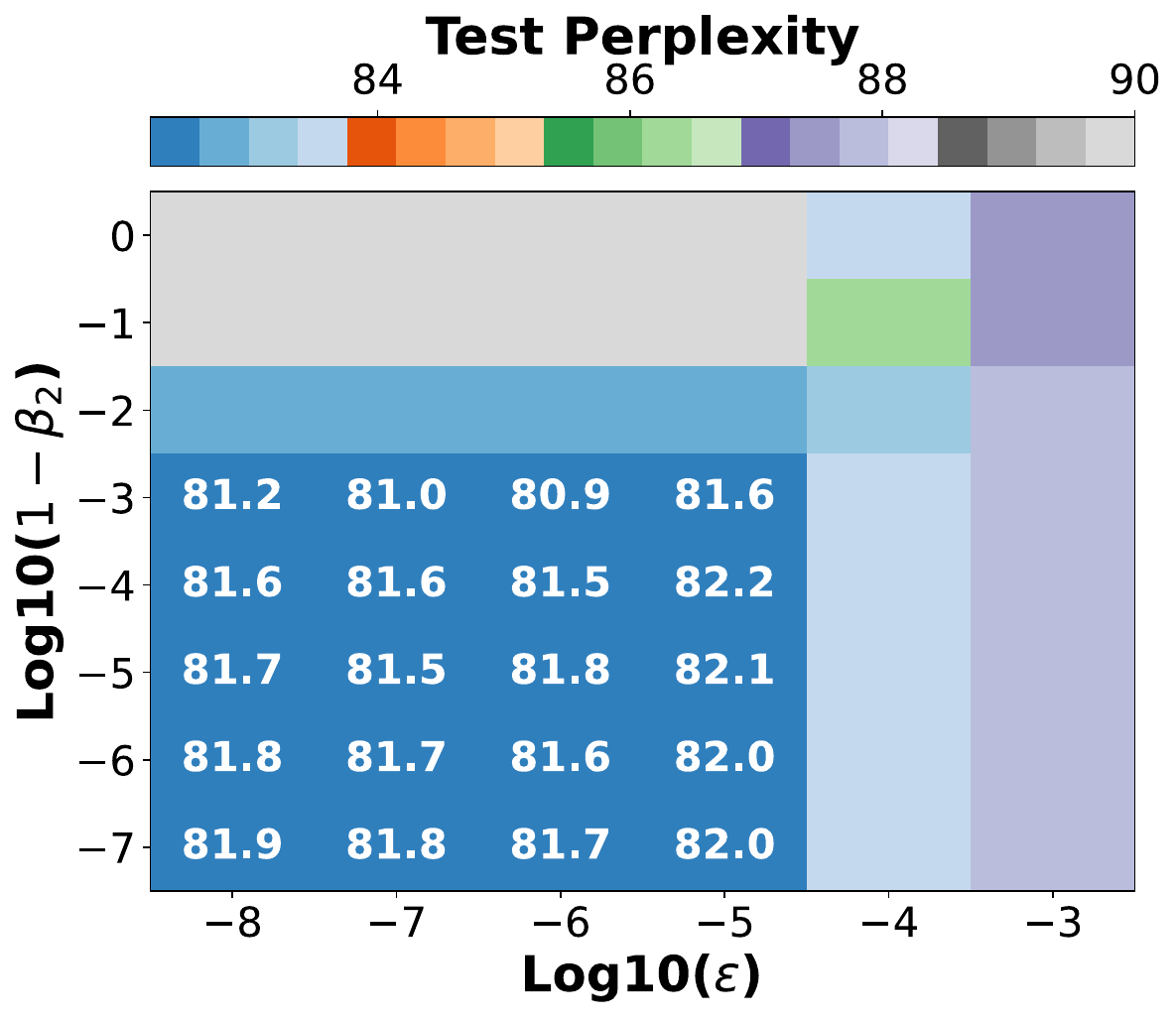} \label{subfig:beta-eps2}}
  \hspace{-2mm}
\subfigure[\begin{tabular}{c}
      \vspace{-2mm}\\
     2-Layer LSTM  - \\
     $\{\alpha= 5 \times 10^{-3}\}$  \vspace{-2mm}
  \end{tabular}]{\includegraphics[width=0.228\textwidth]{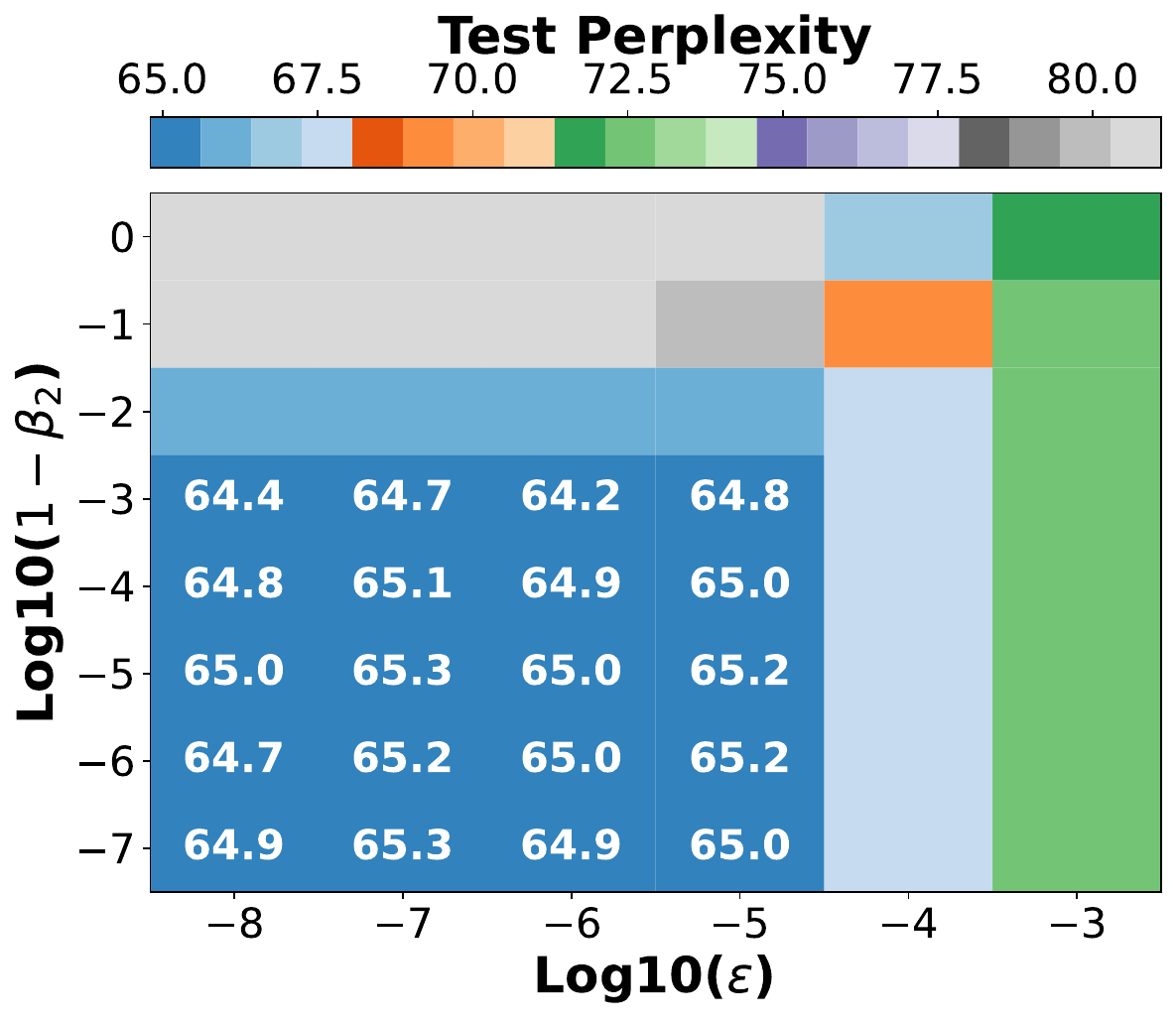} \label{subfig:beta-eps21}} 
  \vspace{2mm}
\caption{Classification \textbf{(top)} and language modeling \textbf{(bottom)} tasks: Performance of VGG11 (trained for 45  epochs) and ResNet34 (trained for 60 epochs) classifiers and 1-layer and 2-layer LSTM models (trained for 200 epochs) on the CIFAR-10 and Penn TreeBank datasets, respectively, using the Adam optimizer. Performance is evaluated while varying the second-order momentum $\beta_2$ and the immutability $\epsilon$. Lower perplexity equals better performance.\label{subfig:beta-eps}
}
\end{figure} 

\subsubsection{To be fully immutable or not?}
\label{sec:final-imu}

\begin{figure}[h!]
\centering
\subfigure[MLP]{\includegraphics[width=0.23\textwidth]{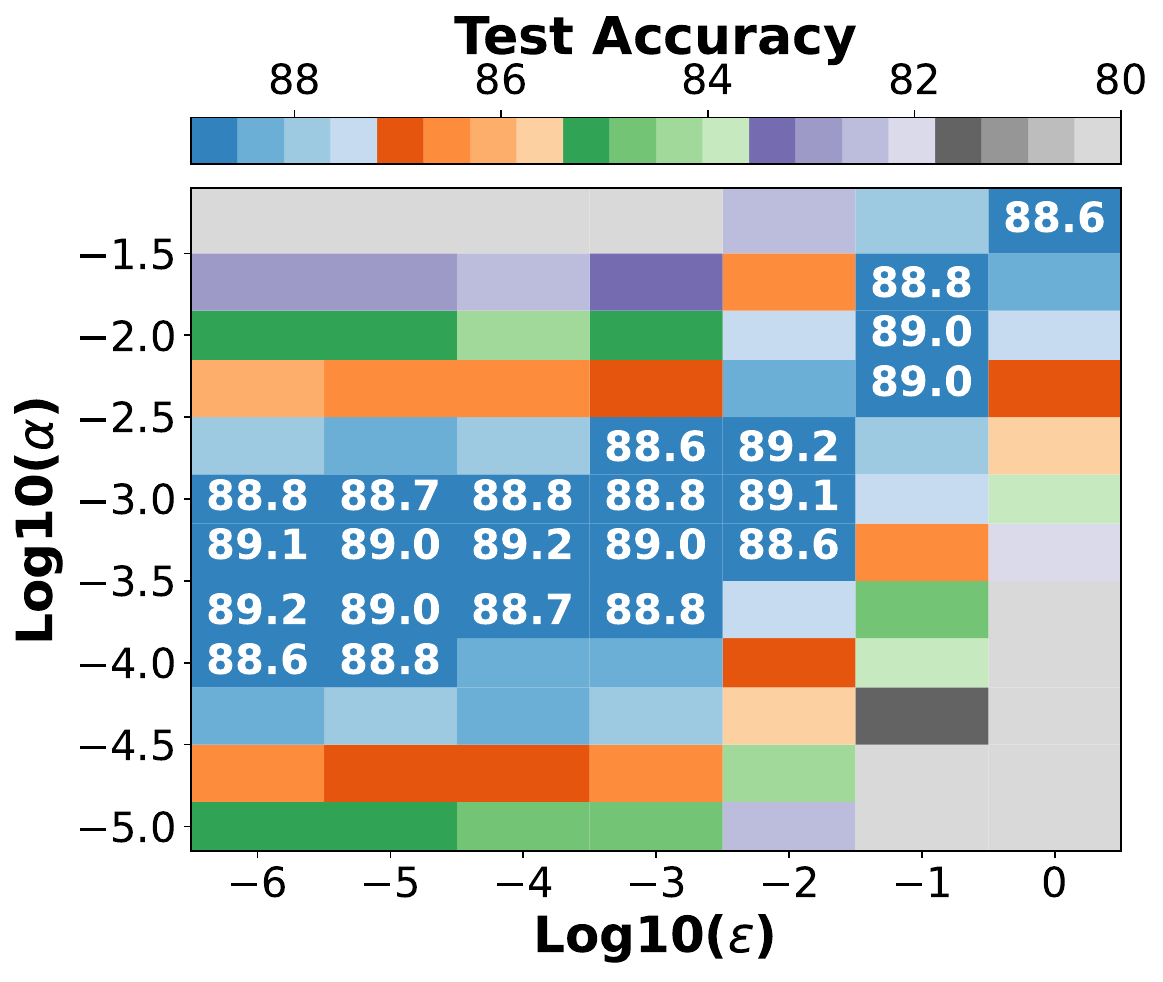} \label{subfig:fig3}} \hspace{-3mm}
\subfigure[AlexNet]{\includegraphics[width=0.23\textwidth]{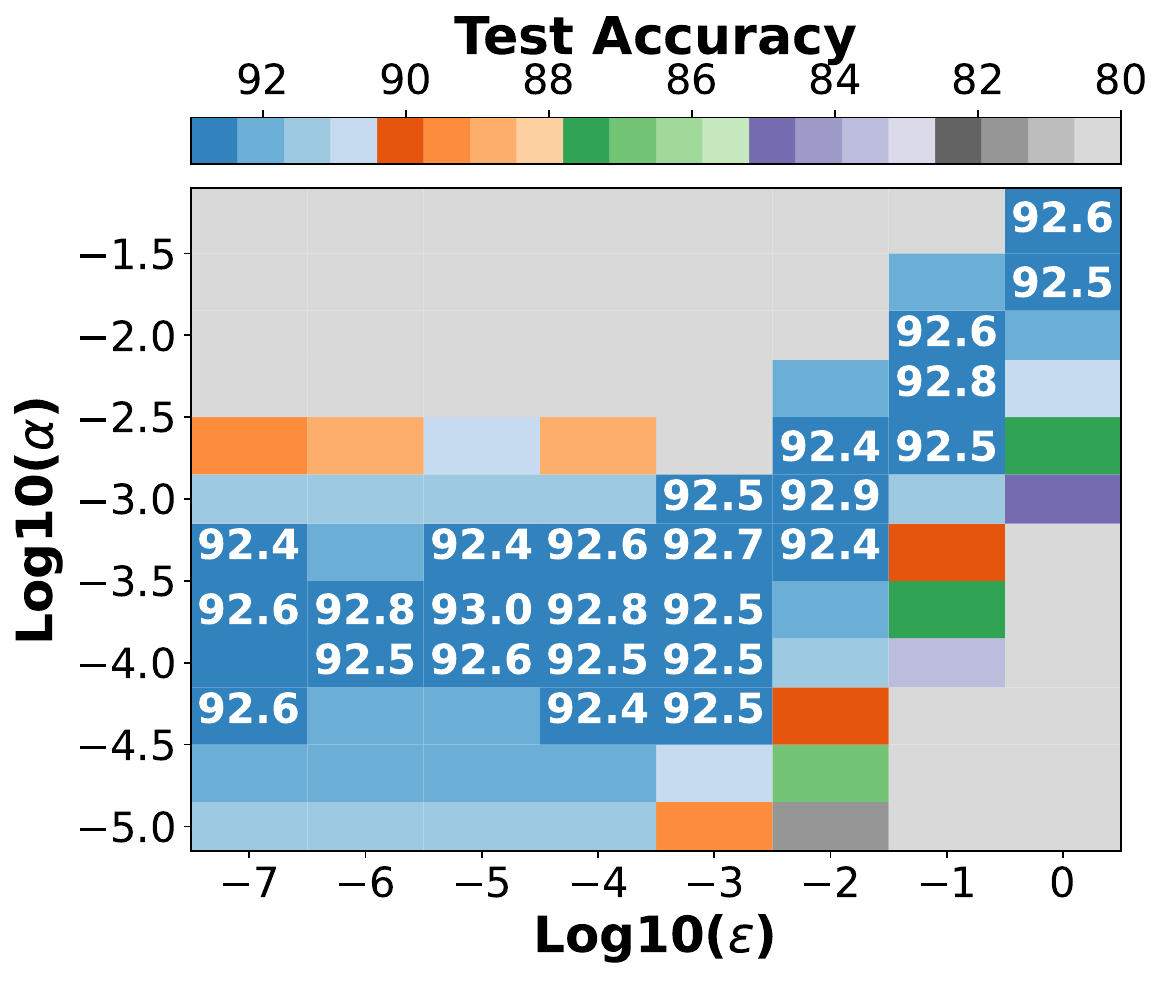} \label{subfig:fig3}}  
\caption{Classification task: Test accuracy of different NN models (trained for 25 epochs) on the Fashion-MNIST dataset with Adam optimizer, varying the learning rate $\alpha$ and  immutability $\epsilon$.}
\label{subfig:ClassMNIST}
\end{figure}

\begin{figure}[h!]
\centering
\subfigure[MLP model | Adam optimizer with $\epsilon = 10^{-2}$]{\includegraphics[width=0.47\textwidth]{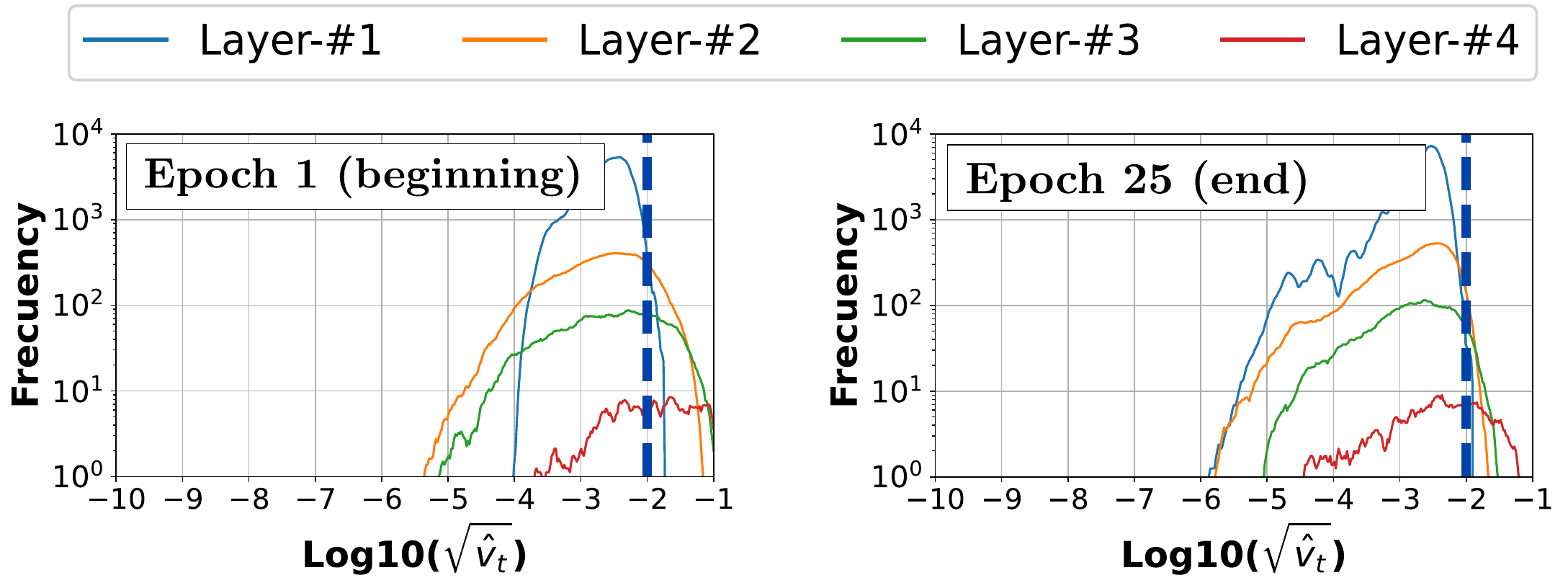} \label{subfig:fig3}} 
 \\ \vspace{-2mm}
\subfigure[MLP model | Adam optimizer with $\epsilon = 10^{-5}$]{\includegraphics[width=0.47\textwidth]{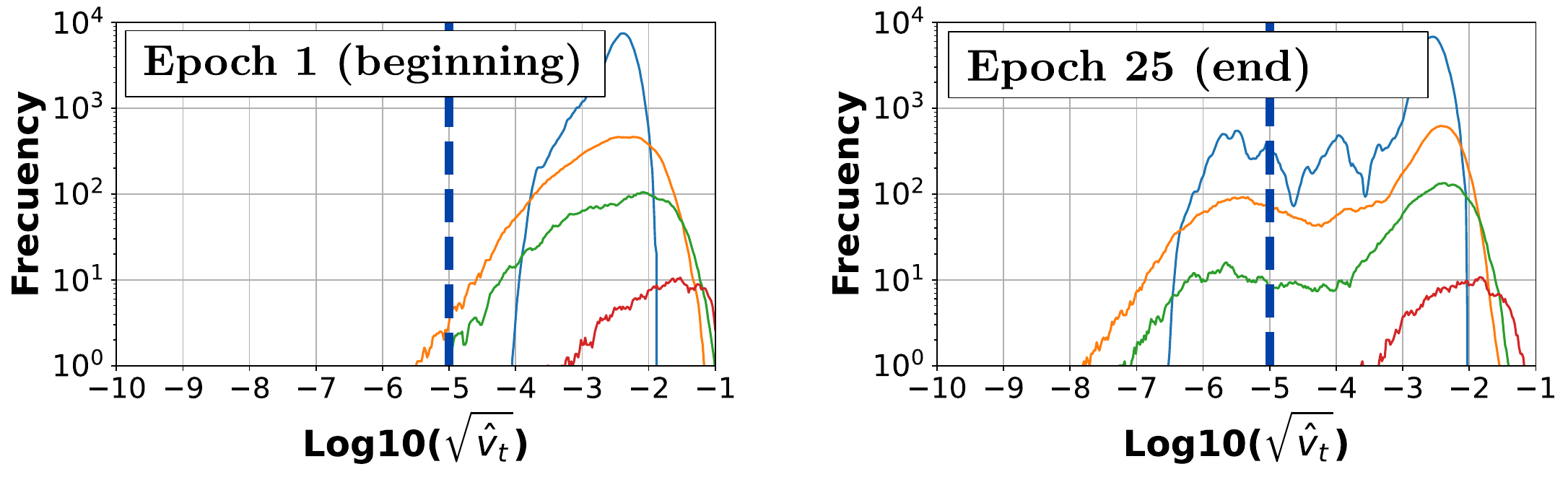} \label{subfig:fig3}} 
  \vspace{1mm}
\caption{Comparison of gradient magnitude histograms of the MLP classifier  trained on the FashionMNIST dataset with Adam, for two values of immutability hyperparameter.}
\label{subfig:distClassMNIST}
\end{figure}

Although the first two cases (classification and language modeling) show that adaptive optimizers like Adam can turn into either SGD+Momentum algorithm or fully adaptive algorithm, we do not rule out the possibility that there is a case where only partial adaptability associated with a moderate level of immutability, for example $50\%$ of dominant adaptive elements in $\sqrt{\hat{v}_t}$, can be essential to promote the best performance, while other higher or lower levels are not. Using the fashion-MNIST dataset, which is simpler compared to the CIFAR-10 and Tiny ImageNet datasets, for the classification task, we found, see Fig. \ref{subfig:ClassMNIST} and Fig. \ref{subfig:distClassMNIST}, that the best optimizer performance is attained over a wide range of immutability hyperparameter values, implying a fully immutable (similar to SGD+Momentum), partially adaptable or fully adaptable behavior in the Adam optimizer without any practical difference between them.

Reflecting on these three tested cases (classification on CIFAR-10, classification on Fashion-MNIST, and language modeling on Penn TreeBank), we wonder why the adaptive stochastic optimizers must be either fully immutable, fully adaptable, or of unimportant nature, the latter characteristic for the Fashion-MNIST case, to achieve the best performance? Before addressing this question, it is important to clarify that most of the adaptive learning rates can be interpreted as adaptive scale factors with normalization properties.
In mathematical optimization field, an element-wise normalized gradient method \cite{watt2020machine}, akin to RMSprop-based methods, is able to provide a fast progress on functions that contain flat regions, such as saddle points, and to be more useful than standard normalized gradient method when the flat regions are in some coordinates. From deep learning literature \cite{keskar2016large,choromanska2015loss, dauphin2014identifying,  zhang2021understanding}, it is reasonable to assume that the effectiveness of an adaptive optimizer with specific characteristic, such as fully immutability (approximately SGD+Momentum) or fully adaptability, is intricately connected to  the convexity of the landscape.

Based on our previous experiments and the concept presented in \cite{dauphin2014identifying}, which proposes that critical points in high-dimensional optimization problems are more likely to be saddle points than local minima, we infer that the landscape in the evaluated language modeling task  is probably non-convex with flat regions. Therefore, it is expected that an adaptive stochastic optimizer with the fully adaptive  attribute (Fig. \ref{subfig:NLP1} and Fig. \ref{subfig:NLP-hist}) outperforms an SGD+Momentum algorithm (Fig. \ref{subfig:classif1b-SGD}), which has slow crawling issues to pass through flat regions of saddle points. However, for the classification task, where the results showed that an adaptive  optimizer closely resembling SGD+Momentum optimizer (Fig. \ref{subfig:classif1}) or the SGD+Momentum optimizer (Fig. \ref{subfig:classif1a-SGD})  yields the best performance,  the landscape might be convex or approximately convex. In order to support this last assumption, we can take into account that \cite{wang2021hidden} has theoretically proved that for a 2-layer network with ReLU activation, the landscape has no spurious local minima, which means a convex surface.  In an approximately convex landscape, particularly near to the minima where the gradients are small, an adaptive stochastic optimizer can generate large learning rates, causing to be halt at higher point compare to the SGD+Momentum algorithm with smaller single learning rate, as  observed in our first evaluated classification case (Fig. \ref{subfig:classif1} and 
Fig.  \ref{subfig:classif-hist}). Nevertheless, when the convex function has a sharper minima, the gradients in a neighbor close to the minimum are higher, so the adaptive learning rates can be small enough to match in performance as the SGD+Momentum algorithm without forcing large immutability factor. This statement must be happening in our last classification case  (Fig.\ref{subfig:ClassMNIST} and Fig.
\ref{subfig:distClassMNIST}).

\subsection{Finding Optimal Search Space for the Immutability Hyperparameter $\epsilon$}
\label{optimal-search-space}

In this final experimental section, we will validate the effectiveness of our straightforward algorithm in determining accurate search ranges for the immutability hyperparameter $\epsilon$ of the Adam optimizer and other adaptive stochastic optimizers.

\begin{table}[h]
\caption{Comparison of immutability hyperparameter ranges (in log scale) for the Adam optimizer in the classification and language modeling tasks. VGG, ResNet, AlexNet and DenseNet classifiers trained on CIFAR-10. LSTM models trained on Penn TreeBank, except for \cite{choi2019empirical}, 2-layer LSTM model trained on Tolstoy’s War and Peace.\label{tabel:bounds}}
\addtolength{\tabcolsep}{-1.5pt}
\renewcommand*{\arraystretch}{1.75}
\centering
\footnotesize
\begin{tabular}{ccccc}
\hline
                        \textbf{Model}                                                             & \textbf{\begin{tabular}[c]{@{}c@{}}Optimal \vspace{-2mm} \\  value \end{tabular}}        & \textbf{\begin{tabular}[c]{@{}c@{}}Tested \vspace{-2mm} \\   Range \cite{wang2021rethinking} \end{tabular}} & \textbf{\begin{tabular}[c]{@{}c@{}}Tested  \vspace{-2mm} \\  Range \cite{choi2019empirical}\protect\footnotemark \end{tabular}}   & \textbf{\begin{tabular}[c]{@{}c@{}}  Our \vspace{-2mm} \\  Estimated \vspace{-1.5mm} \\  Range\end{tabular}} \\ 
                        \hline \hline
\multicolumn{1}{c}{\textbf{VGG}}                                                    
& $ 10^{-3} \leq \epsilon_{} $                                                  
& \dunderline[-3.5pt]{0.8pt}{\color{black} $[-16, -5]$}                                                     
& $[-10, +10]$                       
&\textbf{\color{black} $\mathbf{[-6, -2]}$}                            
\\[1ex] \hline
\multicolumn{1}{c}{\textbf{ResNet}}                                                 
&  $10^{-4} \leq \epsilon_{} $                                                
& \dunderline[-3.5pt]{0.8pt}{\color{black}$[{-16}, {-5}]$}                                                    
& $[{-10}, {+10}]$    
& \textbf{\color{black} $\mathbf{[{-4}, {-1}]}$}     
\\ [1ex]  \hline
\multicolumn{1}{c}{\textbf{AlexNet}}                                                    
&   $10^{-2} \leq \epsilon_{} $                                                
& |                                                    
& |                                   
&\textbf{\color{black} $\mathbf{[{-6}, {-2}]}$}     
\\ \hline
\multicolumn{1}{c}{\color{black}\textbf{DenseNet}}                                                 
& \color{black} $10^{-8} \leq \epsilon_{} \leq 1 $                                                   
&\color{black}$[{-16}, {-5}]$                                                    
& |  
& \textbf{\color{black} $\mathbf{ [{-4}, {-1}]}$}   
\\ [1ex]  \hline
\multicolumn{1}{c}{\textbf{\begin{tabular}[c]{@{}l@{}}1-Layer\vspace{-2mm}\\ LSTM\end{tabular}}} 
&   $10^{-5} \geq \epsilon_{} $                                                
& $[{-16}, {-5}]$                                                    
& |         
& \textbf{\color{black} $\mathbf{[{-6}, {-2}]}$}  
\\ \hline
\multicolumn{1}{c}{\textbf{\begin{tabular}[c]{@{}l@{}}2-Layer\vspace{-2mm}\\  LSTM\end{tabular}}} 
&  $10^{-5} \geq \epsilon_{} $                                           
& $[{-16}, {-5}]$                                                    
& $[{-10}, {+10}]$    
& \textbf{\color{black} $\mathbf{[{-7}, {-2}]}$}      
\\ \hline
\multicolumn{1}{c}{\textbf{\begin{tabular}[c]{@{}l@{}}Trans-\vspace{-2mm}\\  former\end{tabular}}} 
& $10^{-5}  \geq \epsilon_{} $                                           
& $[{-16}, {-5}]$                                                    
& |    
& \textbf{\color{black} $\mathbf{[-11, -2]}$}      
\\ \hline
 \end{tabular}
\end{table}

\footnotetext{\cite{choi2019empirical} initially tested a very large search range, but ended up by recommending a range of shorter length as additional knowledge for future works, where the same models and datasets are employed.}

Given the Adam optimizer, Table \ref{tabel:bounds} shows the distinct search ranges used for the immutability hyperparameter $\epsilon$, where our predicted boundaries are highlighted in bold. As can be seen without prior information of how to pick limits, this is not our case, the reported search ranges in the literature might not contain the optimal immutability hyperparameter as observed in underlined ranges, resulting in an inferior performance for the Adam optimizer. For example, in the VGG11 model trained on CIFAR-10, if we employ the recommended range ($\epsilon \in [10^{-16},10^{-5}]$) {\color{black} and 150 epochs} as suggested in \cite{wang2021rethinking}, the  maximum achieved accuracy is $89.3\%$ (see Fig. \ref{Fig21a} of Appendix \ref{appendixA}),  which is $2.3\%$ lower than the accuracy achieved using our estimated range. This performance deficiency can be even greater (8.3 percentage points) when dealing with more complex classification tasks, as illustrated in Fig. \ref{subfig:vgg-imagenet} of Appendix \ref{appendixA}. Furthermore, it  is worth noting that for the different trained models, our algorithm estimates more accurate and compact ranges, which contain approximately 6 immutability combinations for testing instead of 12 to 21 combinations as reported in the literature \cite{wang2021rethinking, choi2019empirical}. In Table \ref{tabel:boundsOtherOptimizers}, we also show the success of our algorithm in estimating search spaces for other adaptive stochastic optimizers, consistently including the optimal immutability value that can differ depending on the task, model and optimizer employed.

Considering our estimated ranges and previous experimental findings, in order to get the best performance and to further reduce computational costs, we suggest evaluating only the two extreme values of the estimated immutability boundaries, or even trying the SGD+Momentum optimizer, which spends less training time, instead of using the Adam optimizer with the upper bound, as it is structurally and behaviorally similar to an SGD+Momentum optimizer. In both situations, with only one or two values for the immutability hyperparameter $\epsilon$, we can now focus on varying the learning rate hyperparameter $\alpha$ to maximize performance. When choosing to train models with the SGD+Momentum optimizer (the second recommendation), note that its search space of the optimal learning rate hyperparameter $\alpha$ can vary drastically depending on the tasks (Fig. \ref{subfig:classif1-SGD}), while the optimal learning rate values of an adaptive stochastic optimizer are more centralized (Fig. \ref{subfig:classif1} and Fig. \ref{subfig:NLP1}), i.e. these have less variance across tasks.

\begin{table}
\caption{Summary of the estimated immutability hyperparameter ranges (in log scale) for RMSProp, AdaBelief and AdaMomentum optimizers. VGG, LSTM and Transformer models were trained on CIFAR-10, Penn TreeBank, and IWSTL’14  datasets, respectively.\label{tabel:boundsOtherOptimizers}}
\renewcommand*{\arraystretch}{1.5}
\centering
\footnotesize
\begin{tabular}{ccccc}
\hline
\textbf{Task}                                                                               & \textbf{Model}                                                                         & \textbf{Optimizer}                                                              &
 \textbf{\begin{tabular}[c]{@{}c@{}}  Optimal \vspace{-1mm} \\ Value\protect\footnotemark\end{tabular}} 
 & \textbf{\begin{tabular}[c]{@{}c@{}}  Our \vspace{-1mm} \\ Estimated \vspace{-1mm} \\  Range\end{tabular}} \\ \hline \hline
\multirow{3.5}{*}{\begin{tabular}[c]{@{}c@{}}\textbf{Classifi-}\vspace{-1mm} \\ \textbf{cation}\end{tabular}}                                                           & \multirow{3.5}{*}{\textbf{VGG}}                                                                   & \textbf{RMSProp}                                                                
& $10^{-4}\leq \epsilon_{}$       
& $\mathbf{[{-6},{-2}]}$     
\\ \cmidrule{3-5} 
                                                                                            &                                                                                        & \textbf{AdaBelief}                                                              
& $10^{-8}\leq \epsilon_{}$      
& $\mathbf{[{-17},{-6}]}$     \\ \cmidrule{3-5} 
                                                                                            &                                                                                        & \textbf{\begin{tabular}[c]{@{}c@{}}AdaMo-\vspace{-2mm}\\ mentum\end{tabular}} 
  & $10^{-12}\leq \epsilon_{}$     
  & $\mathbf{[{-16},{-7}]}$    
  \\ \hline
\multirow{3.5}{*}{\begin{tabular}[c]{@{}c@{}}\textbf{Language}\vspace{-1mm} \\ \textbf{Modeling}\end{tabular}} & \multirow{3.5}{*}{\begin{tabular}[c]{@{}c@{}}\textbf{1 Layer}-\vspace{-1mm}\\ \textbf{LSTM}\end{tabular}} & \textbf{RMSProp}                                                                
& \color{black}$10^{-5}\geq \epsilon_{}$     
& $\mathbf{[{-6},{-2}]}$     
\\ \cmidrule{3-5} 
                                                                                            &                                                                                        & \textbf{AdaBelief}                                                              
& $10^{-14}\geq \epsilon_{}$     
& $\mathbf{[{-15},{-8}]}$     \\ \cmidrule{3-5} 
                                                                                            &                                                                                        & \textbf{\begin{tabular}[c]{@{}c@{}}AdaMo-\vspace{-2mm}\\ mentum\end{tabular}} 
&  $10^{-14}\geq \epsilon_{}$     
& $\mathbf{[{-16},{-7}]}$     \\ \hline
\multirow{4}{*}{\begin{tabular}[c]{@{}c@{}}\textbf{Language}\vspace{-1mm} \\ \textbf{Trans-}\vspace{-1mm} \\ \textbf{lation}\end{tabular}} & \multirow{3.5}{*}{\begin{tabular}[c]{@{}c@{}}\textbf{Trans}-\vspace{-1mm}\\ \textbf{former}\end{tabular}} & \textbf{RMSProp}                                                                
& \color{black}$10^{-5}\geq \epsilon_{}$     
& $\mathbf{\color{black} [-11, -2]}$     
\\ \cmidrule{3-5} 
                                                                                            &                                                                                        & \textbf{AdaBelief}                                                              
&\color{black}  $10^{-16}\geq \epsilon_{}$     
& $\mathbf{\color{black} [-25,-6]}$     \\ \cmidrule{3-5} 
                                                                                            &                                                                                        & \textbf{\begin{tabular}[c]{@{}c@{}}AdaMo-\vspace{-2mm}\\ mentum\end{tabular}} 
& \color{black}  $10^{-16}\geq \epsilon_{}$     
& $\mathbf{\color{black} [{-25},{-6}]}$     \\ \hline
\end{tabular}
\end{table}
\footnotetext{The optimal values were found from a logarithmic grid search as illustrated in the first column Fig. \ref{subfig:lr-eps-AdaBelief} and Fig. \ref{subfig:NLP-optimizer} of Appendix \ref{appendixA}.}

\section{Conclusion}
\label{Conclusion}

In this manuscript, we introduced a pioneering framework based on gradient magnitude histograms for the analysis of adaptive stochastic optimizers from an alternative perspective. Our primary objective was to uncover relevant insights into adaptive stochastic optimizers and to automate and refine the estimation of safeguard hyperparameter $\epsilon$ to achieve optimal performance. Notably, the safeguard hyperparameter $\epsilon$  is redefined as the  “immutability hyperparameter” in recognition of its inherent opposition to the adaptability property.

Within this innovative framework, we developed an algorithm designed to estimate an precise and narrowed search space for the optimal immutability hyperparameter $\epsilon$  of diverse state-of-the-art adaptive stochastic optimizers, including AdaGrad, RMSprop, Adam, AdaBelief, Adamomentum, and others.  In contrast to prevalent predefined search spaces in the literature, which are often erroneously fixed and broad, encompassing 12 to 20 values for evaluating, our algorithm consistently identified efficient search spaces that are independent of the optimizer, model, data, or task. These estimated high-precision search spaces comprise approximately less than half of the values, indicating significantly reduced search ranges. Complementary, our experimental evidence suggested that the evaluation of only two immutability values—the lower and upper bounds of the estimated search space— are sufficient to select the optimal immutability hyperparameter.  We subsequently recommended a secondary exploration to determine the corresponding optimal learning rate hyperparameter considering these two values.

Finally, we identified and justified particular relationships and dependencies among hyperparameters—namely, the immutability factor $\epsilon$, learning rate $\alpha$, and decay rate $\beta_2$—in connection to their optimal performance. For instance, depending on whether the optimal immutability value aligns with the upper or lower bounds of the estimated search space, the chosen value of the decay rate $\beta_2$ has negligible impact on performance or can be directly computed using the relation provided in Section \ref{sub:influ_beta2}, respectively.
Furthermore, the analysis of gradient magnitude histograms contributed to identifying applications better suited for evaluating new adaptive stochastic optimizers.

\bibliographystyle{IEEEtran}
\bibliography{bibliographyV2}

\appendices
\section{Additional Experiments}
\label{appendixA}

To provide a more comprehensive study of the classification task, Fig. \ref{subfig:lr-eps-DenseNet} to  Fig. \ref{subfig:lr-eps-TinyImageNet} illustrate the performance of the Adam optimizer when training AlexNet and DenseNet121 models on the CIFAR-10 dataset, as well as VGG11 and AlexNet on the more complex Tiny ImageNet dataset. Meanwhile, Fig. \ref{subfig:Transformer}  shows the performance of the
Adam optimizer on the Transformer model for the machine
translation task using the IWSLT’14 dataset.
{\color{black}Detailed training settings are provided in Section  \ref{subsect-settings}. Additionally, to analyze overfitting, we present the accuracy gap (the difference between training and test accuracy) and the progression of test loss for the classification task in Fig. \ref{Fig18}, Fig. \ref{Fig21}, and Fig. \ref{Fig22},  and for language modeling tasks in Fig. \ref{Fig19} and  Fig. \ref{Fig20}.}
For both classification and language modeling tasks, we also evaluate the behavior of other adaptive optimizers, including RMSprop \cite{tieleman2012lecture}, AdaBelief \cite{zhuang2020adabelief} and AdaMomentum \cite{wang2021rethinking}, with the optimal value of the immutability hyperparameter $\epsilon$, as shown in Fig. \ref{subfig:lr-eps-AdaBelief} and   Fig.  \ref{subfig:NLP-optimizer}.

\setcounter{figure}{13} 
\begin{figure}[h]
\centering
\subfigure[AlexNet]{\includegraphics[width=0.226\textwidth]{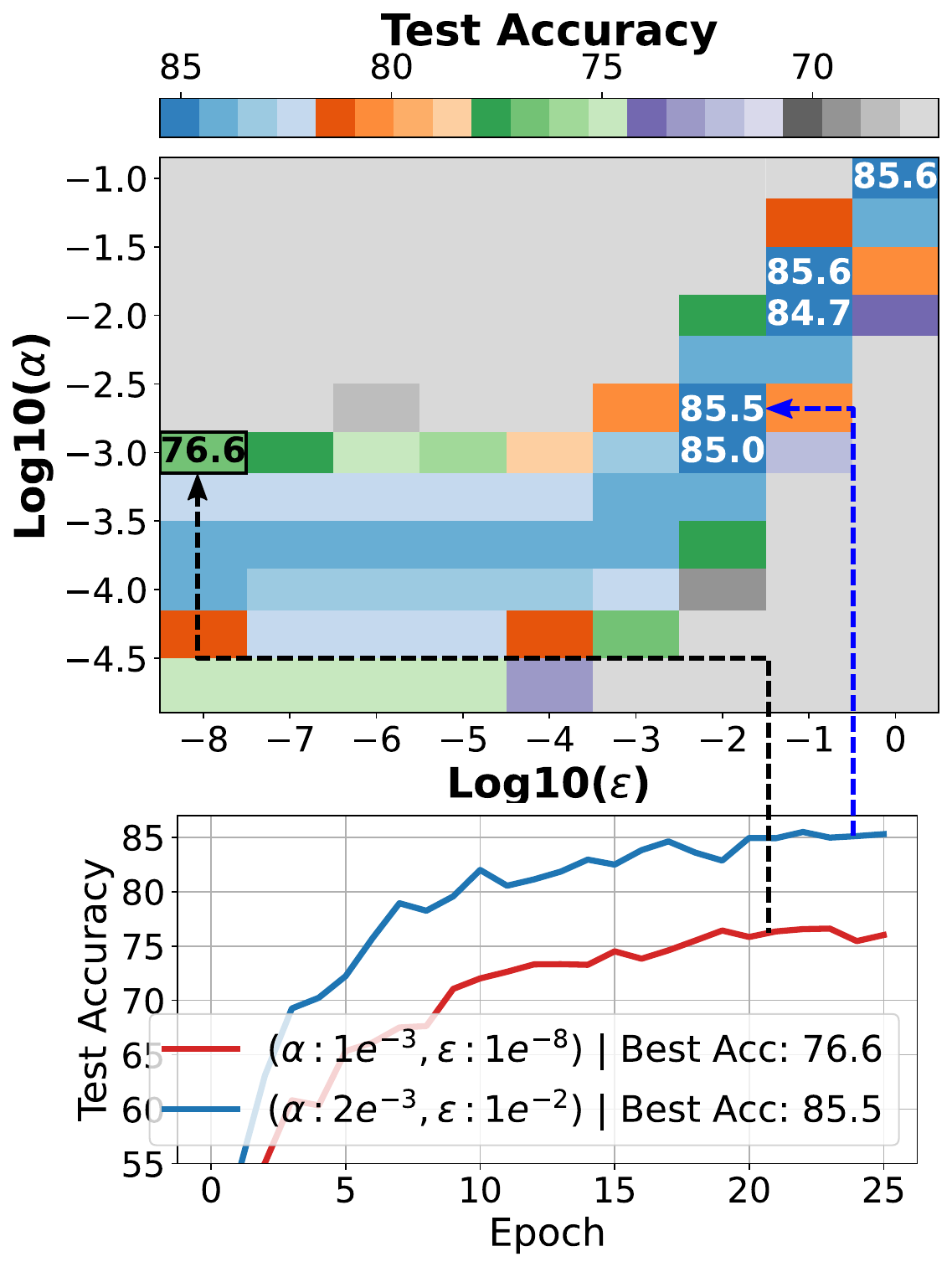} \label{subfig:fig3}}
\subfigure[DenseNet]{\includegraphics[width=0.226\textwidth]{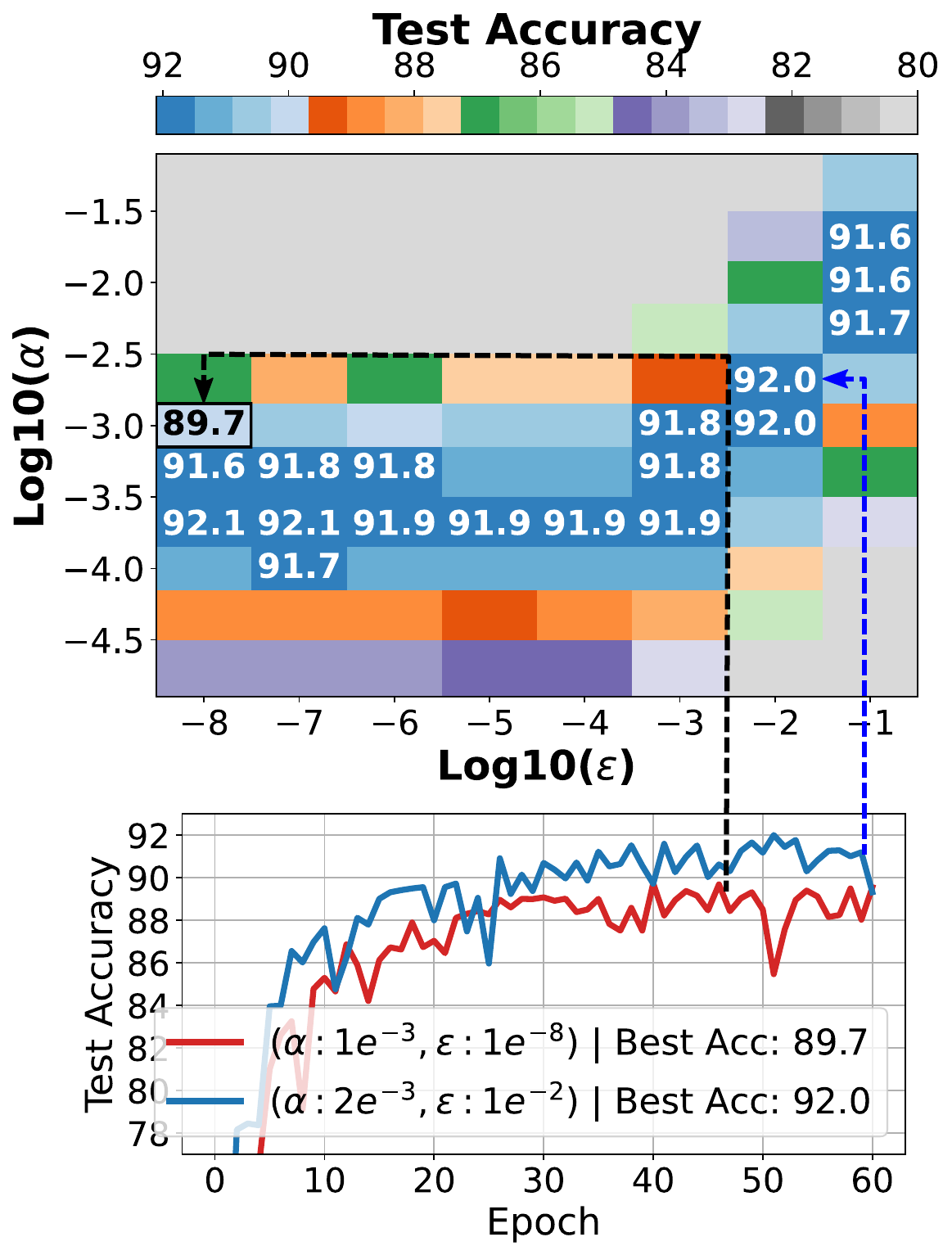} \label{subfig:fig3}}   \\ \vspace{-2.5mm}
\subfigure[{\color{black}AlexNet -  $\{\alpha= 2\times 10^{-3}\}$}]{\includegraphics[width=0.226\textwidth]{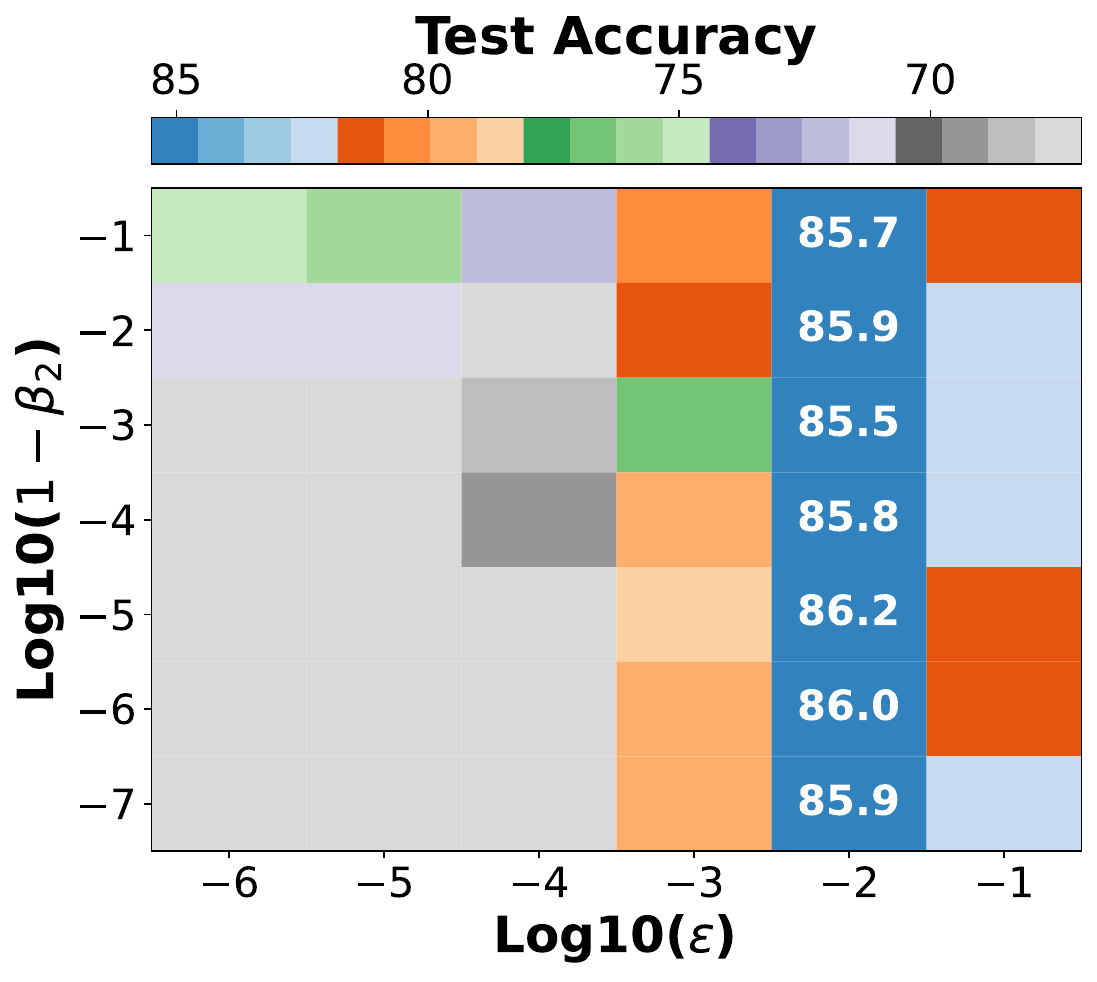} \label{subfig:beta-eps5}} 
\subfigure[DenseNet -  $\{\alpha= 2\times 10^{-3}\}$]{\includegraphics[width=0.226\textwidth]{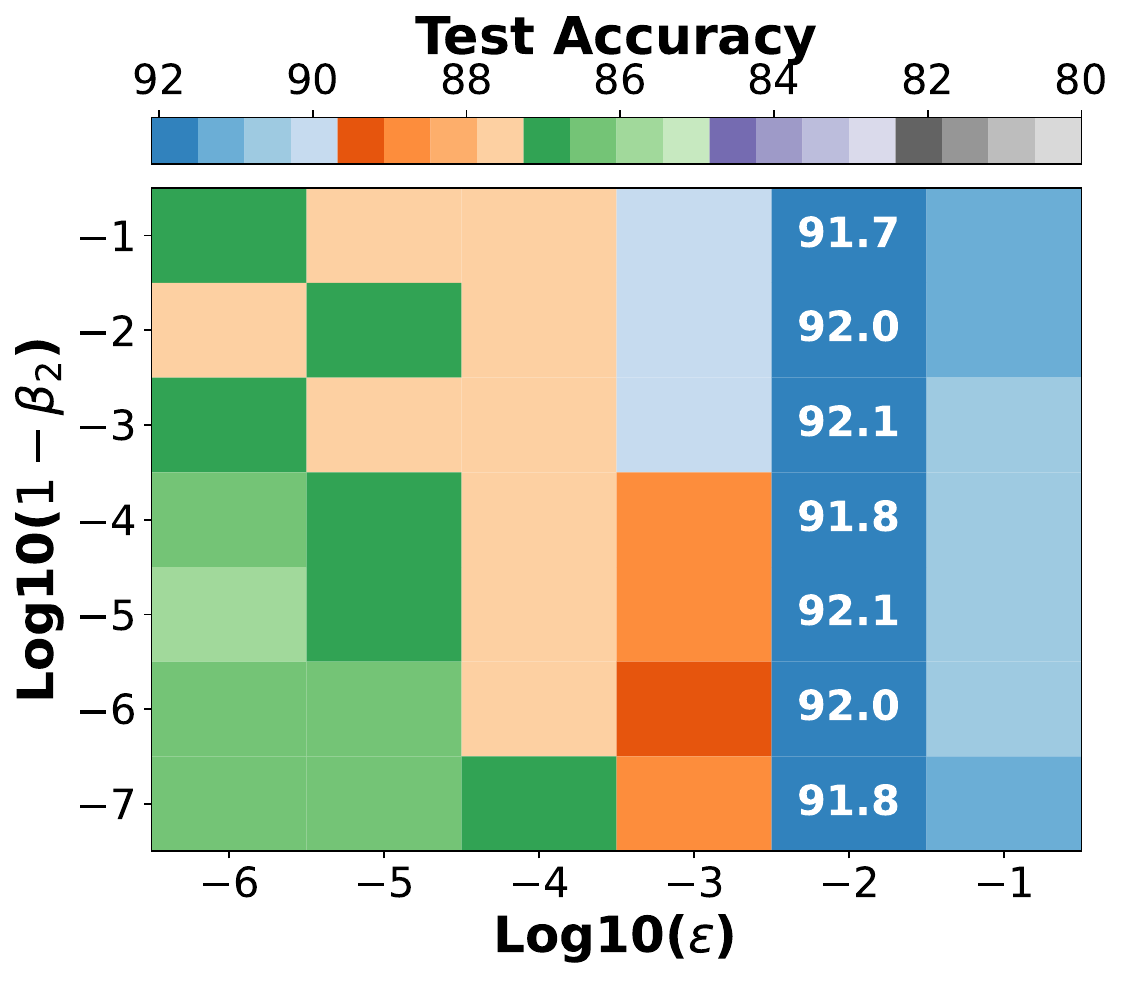} \label{subfig:beta-eps6}} 
\caption{Test accuracy of AlexNet (trained for 25 epochs) and  DenseNet121 (trained for 60 epochs) classifiers on the CIFAR-10 dataset with Adam optimizer, varying learning rate hyperparameter $\alpha$ vs immutability  hyperparameter $\epsilon$ \textbf{(top)} and  varying second-order momentum $\beta_2$ vs  immutability   $\epsilon$ \textbf{(bottom)}.\label{subfig:lr-eps-DenseNet}}
\end{figure}

\begin{figure*}
\centering
\subfigure[AlexNet | From epoch 1 to epoch 25]{\includegraphics[width=0.95\textwidth]{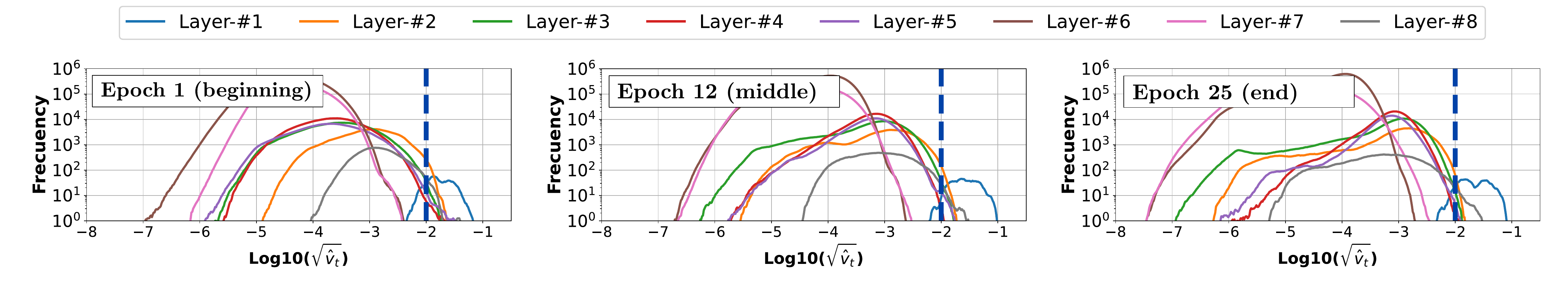} \label{subfig:fig3}} 
\\ \vspace{-1.5mm}
\subfigure[DenseNet121 | From epoch 1 to epoch 60]{\includegraphics[width=0.95\textwidth]{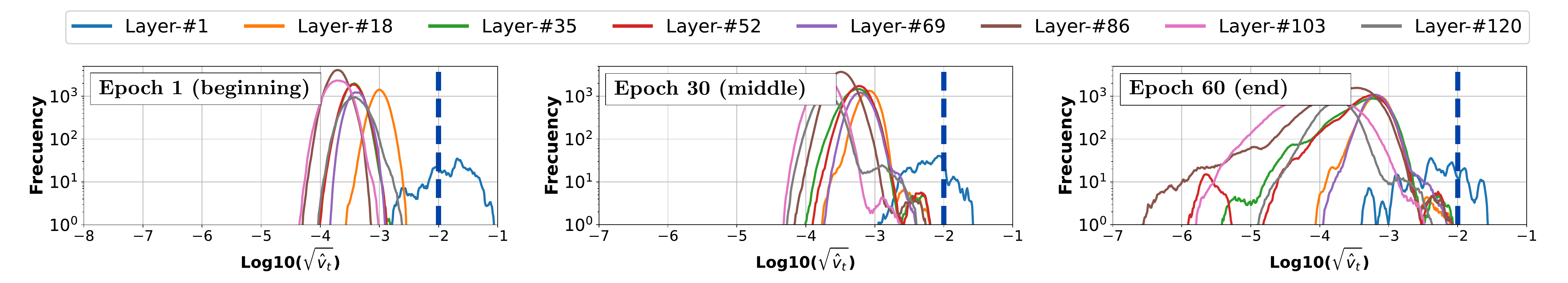} \label{subfig:fig3}} 
\vspace{-1mm}
\caption{Progress of gradient magnitude histograms of the AlexNet and DenseNet121 classifiers  trained on the CIFAR-10 dataset with Adam and an immutability hyperparameter $\epsilon = 10^{-2}$. Vertical dashed blue line (chosen immutability hyperparameter $\epsilon$) marks the boundary of discarded elements, where gradients on the left-side are attenuated by $\epsilon$, resulting in approximately constant learning rates $\alpha_t \approx \alpha/ \epsilon$, and gradients on the right-side generate adaptive learning rates.}
\label{subfig:dist-lr-eps-DenseNet}
\end{figure*}

\begin{figure*} 
\vspace{-2.5mm}
\subfigure[VGG11]{\includegraphics[width=0.228\textwidth]{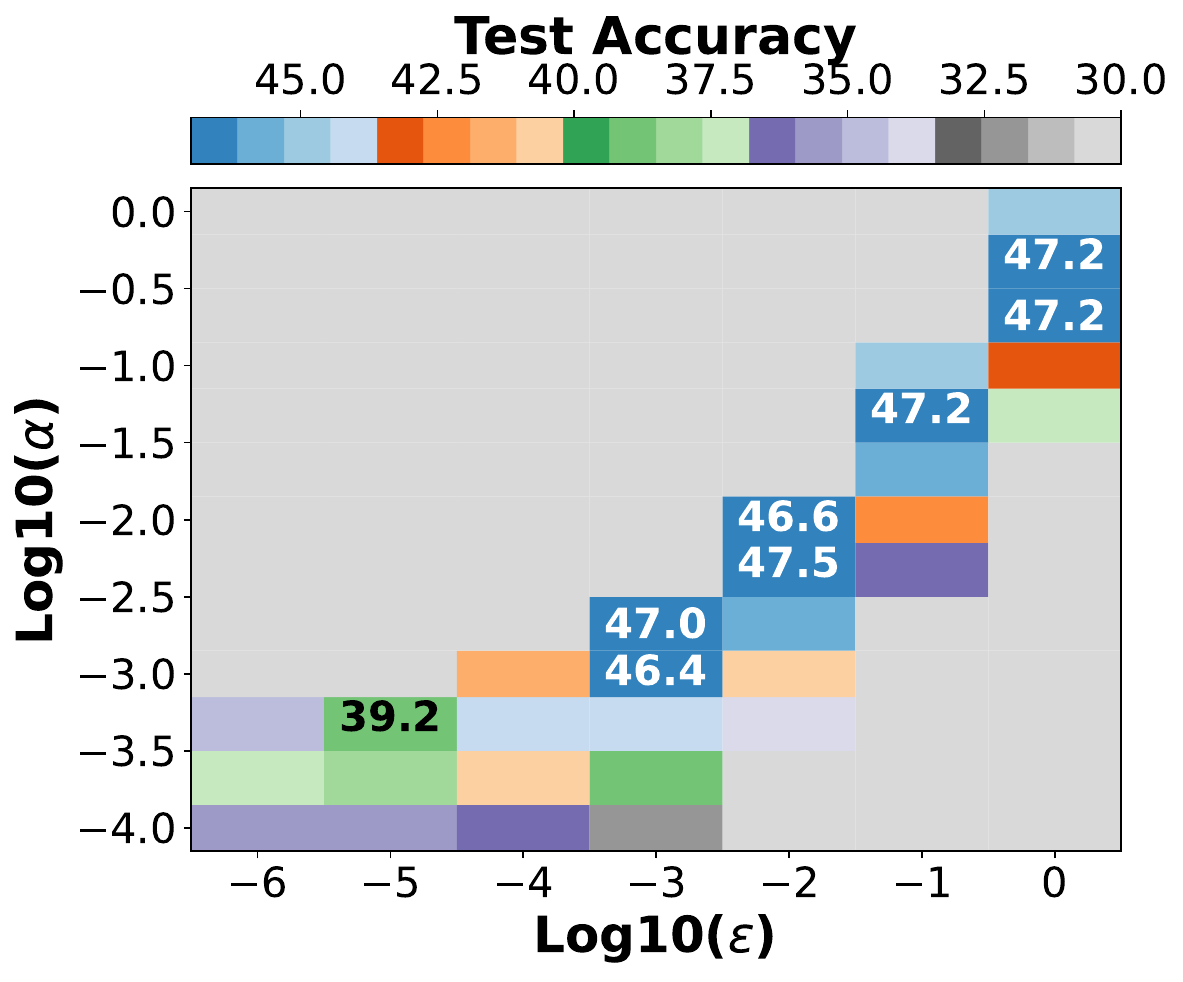} \label{subfig:vgg-imagenet}}\hspace{-2mm}
\subfigure[ VGG11 | Epoch 1]{\includegraphics[width=0.228\textwidth]{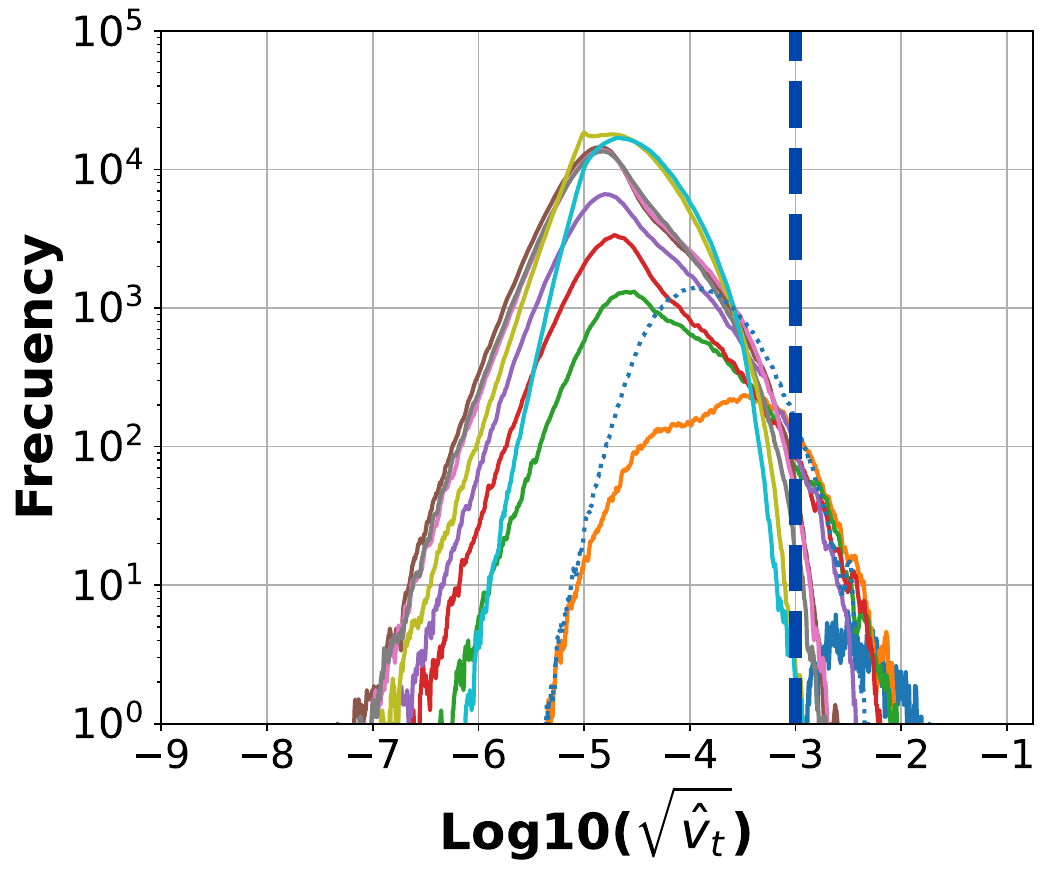} \label{subfig:fig3}}  
\subfigure[AlexNet]{\includegraphics[width=0.226\textwidth]{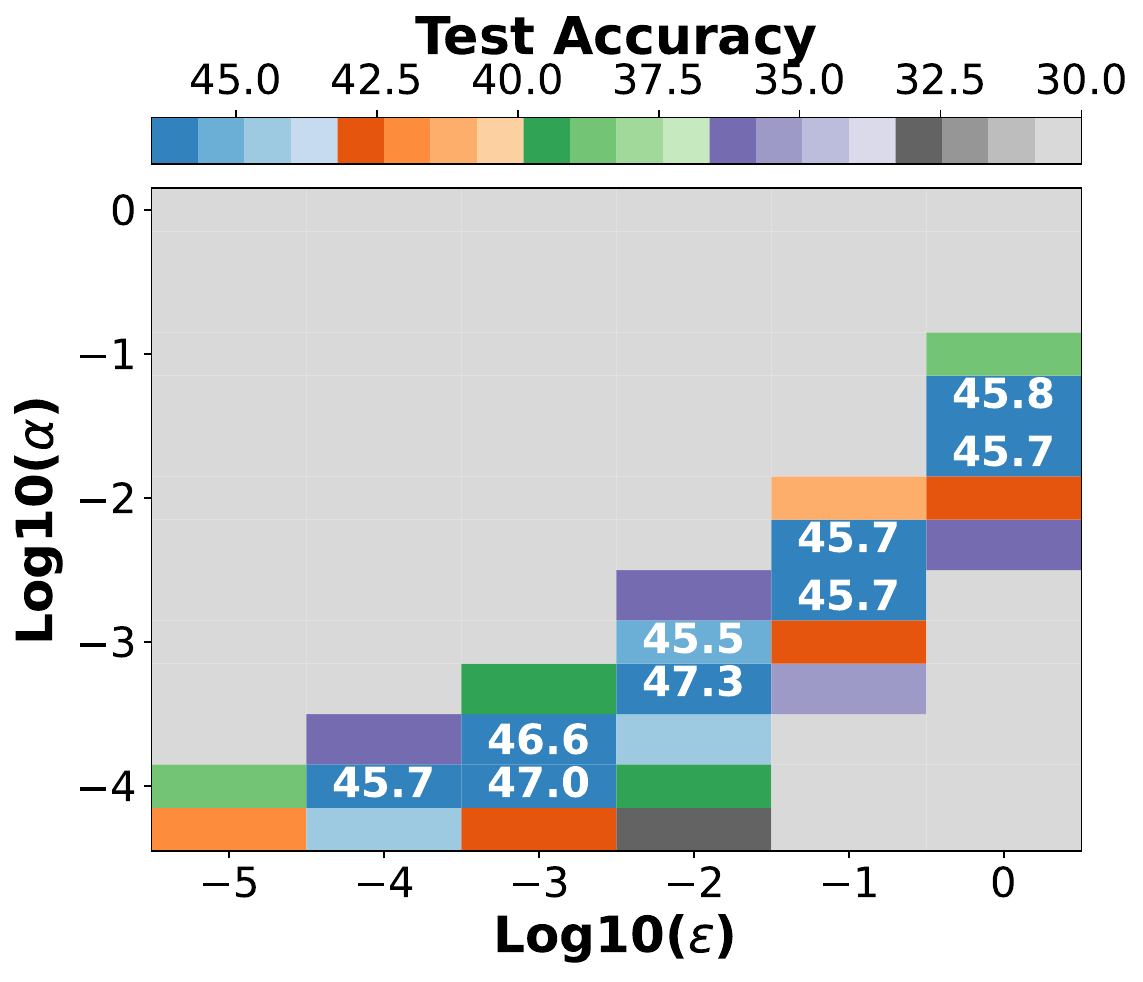} \label{subfig:fig3}}\hspace{-2mm}
\subfigure[AlexNet | Epoch 1]{\includegraphics[width=0.226\textwidth]{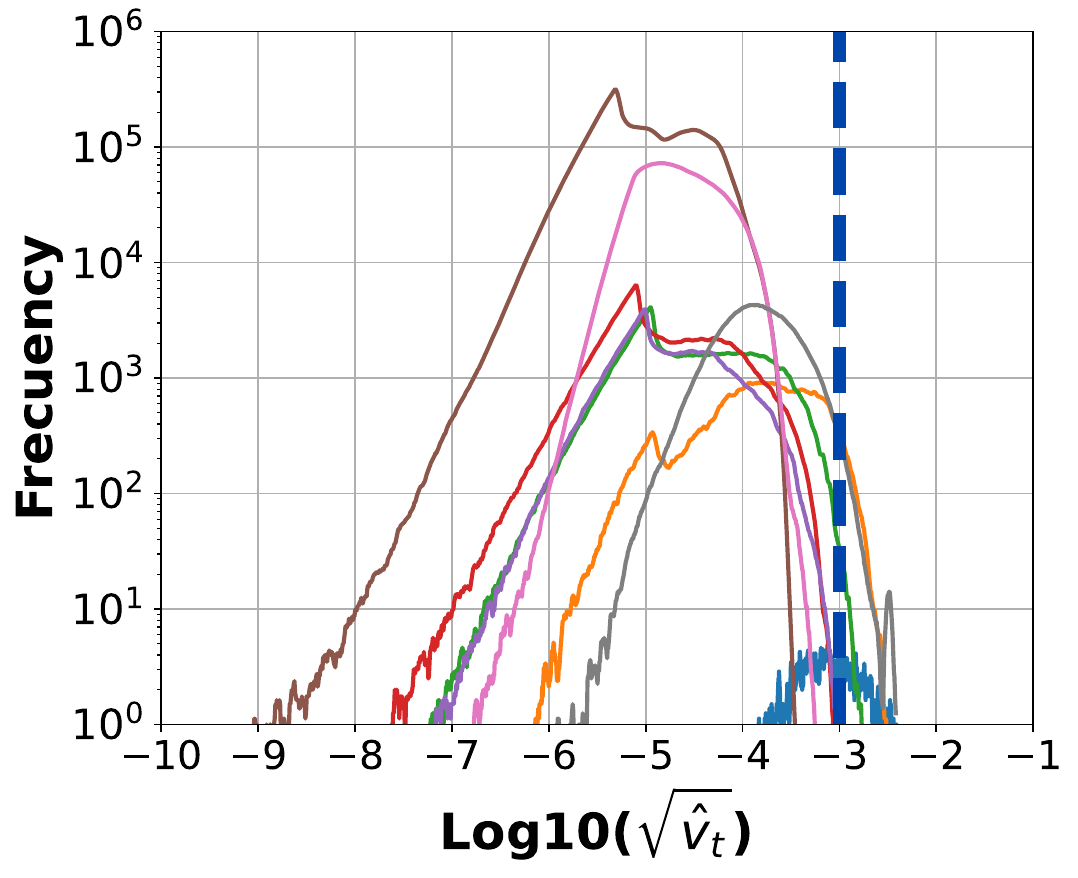} \label{subfig:fig3}}   
\caption{Test accuracy of VGG11 (trained for 45 epochs) and AlexNet (trained for 25 epochs) classifiers on the Tiny ImageNet dataset using the Adam optimizer. Performance by varying hyperparameters $\alpha$ and  $\epsilon$,   and gradient magnitude histograms at first epoch are assessed, where in the gradient magnitude histograms $0.17\%$ and {\color{black} $0.032\%$} of adaptive elements are greater than the chosen optimal values of immutability hyperparameter (vertical dashed blue line) for the respective classifiers.}
\label{subfig:lr-eps-TinyImageNet} 
\end{figure*}

{\color{black} As observed in Fig. \ref{subfig:classif1} for the classification task, Fig. \ref{subfig:lr-eps-DenseNet} illustrates the same linear pattern between $\alpha$ and $\epsilon$ after training AlexNet and DenseNet models. This pattern reveals that the Adam optimizer behaves similarly to SGD+Momentum with a single constant learning rate.

Before analyzing overfitting, it is worth noting that common practice in the deep learning field suggests that an accuracy gap around $5\%$ is considered acceptable and indicative of good generalization, while an accuracy gap of $10\%-15\%$ or more is often a sign of strong overfitting, although this can vary depending on the specific application and dataset. For the classification tasks depicted in Fig. \ref{subfig:classif1} and Fig. \ref{subfig:lr-eps-DenseNet}, we can observe in Fig. \ref{Fig18} that there is no continuous increase in  their corresponding test loss performance, indicating no overfitting issues. Particularly, in the region of interest, the accuracy gaps are around $5\%$ for smaller models (VGG11 and AlexNet), and for larger models (ResNet34 and DenseNet121), the accuracy gaps range from $3.4\%$  to $7.1\%$. However, when training for a higher number of epochs, such as the 150 epochs used in \cite{zhuang2020adabelief} and reproduced in Fig. \ref{Fig21}, the test loss exhibits a consistent increase at certain points in the progression, or even the final accuracy gap increases. Therefore, when employing 150 epochs, it is recommended to use the early stopping technique to avoid overfitting issues, as presented in Fig. \ref{Fig22}. Regardless of these three cases, a similar linear pattern is evident, where the Adam optimizer behaves similarly to SGD+Momentum with an approximately single learning rate $\alpha_t \approx \alpha/ \epsilon$, where increasing $\alpha$ in the same proportion as $\epsilon$ gives the same $\alpha_t$.

Examining the training and test loss progression for the language processing task, as illustrated in Fig. \ref{Fig19} and Fig. \ref{Fig20} in conjunction with the experiment shown in Fig. \ref{subfig:NLP1}, where LSTM models are trained using the Adam optimizer, demonstrates that there are no signs of overfitting. Both training and test losses consistently decrease over the course of 200 epochs, reflecting stable model performance.
}

Regarding other adaptive optimizers shown in the Fig. \ref{subfig:lr-eps-AdaBelief}, in the context of the classification task, the RMSprop algorithm, with an optimal immutability hyperparameter $\epsilon$, behaves similarly to the basic SGD algorithm throughout the entire training process. This is because the update direction in RMSprop is determined by the current gradient and its adaptive learning rate is approximately constant $\alpha_t \approx \alpha/ \epsilon$. On the other hand, the AdaBelief and AdaMomentum optimizers with $\alpha_t \approx \alpha/ \epsilon$ tend to behave as the SGD+Momentum optimizer  (Eq. \ref{eq:Mom1}, where $\mu = \gamma = \beta_1$).

For the language modeling task, as depicted in the Fig. \ref{subfig:NLP-optimizer}, given the largest optimal values of immutability hyperparameter $\epsilon$ corresponding to the RMSprop, AdaBelief and AdaMomentum optimizers, $85.73\%$, $99.31\%$ and  $91.42\%$ of adaptive elements are respectively greater than the chosen immutability values from start of training. Furthermore, if the optimal immutability hyperparameter $\epsilon$ is smaller value, which also guarantees best performance (refer to the leftmost column of the Fig. \ref{subfig:NLP-optimizer}), there will be a greater number of dominant adaptive elements up to $100\%$, meaning a fully adaptive algorithm. 

\clearpage

\begin{figure*}
\subfigure[Transformer]{\includegraphics[width=0.237\textwidth]{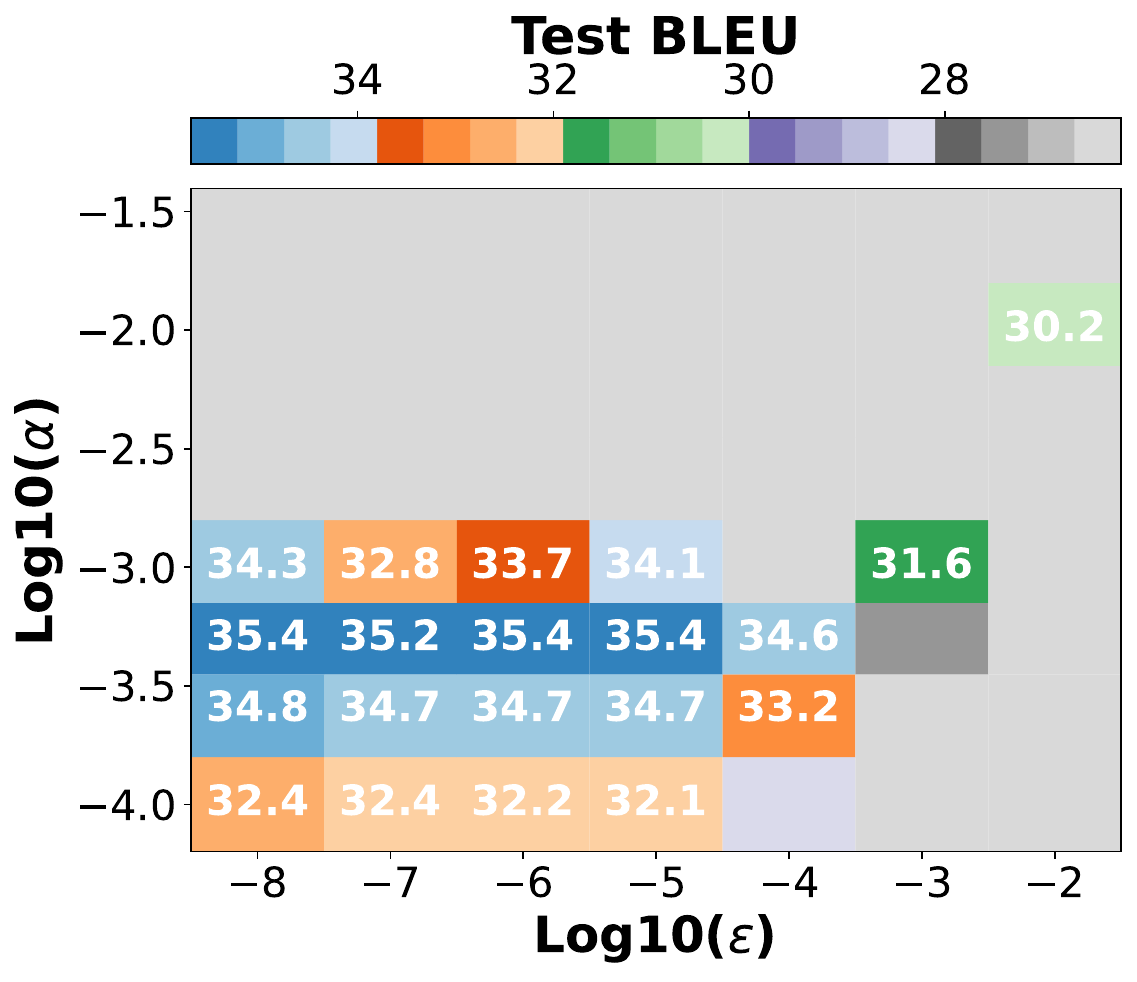} \label{subfig:fig3}} \hspace{-2mm}
\subfigure[Transformer - From epoch 1 to epoch 50]{\includegraphics[width=0.74\textwidth]{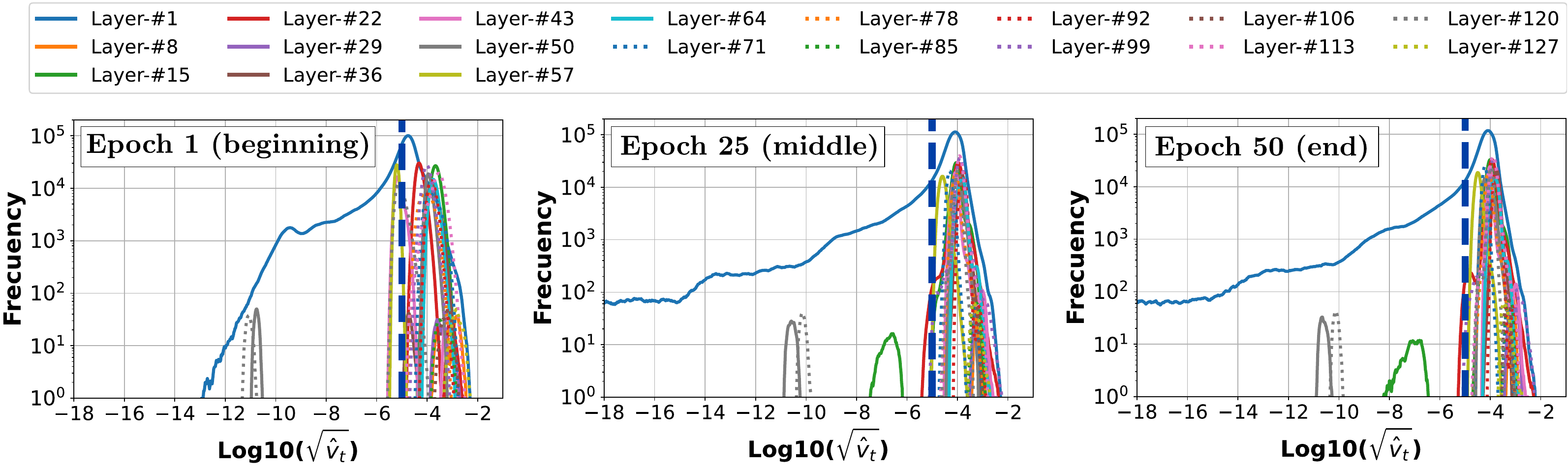} \label{subfig:fig3}} 
\caption{\color{black} Test BLEU \cite{papineni2002bleu} of Transformer model (trained for 50 epochs) on the  IWSTL’14 dataset using the Adam optimizer. In the first column, performance is evaluated by varying learning rate hyperparameter $\alpha$ and  immutability hyperparameter $\epsilon$, where higher BLEU  corresponds to better performance. In the second to fourth column, progress of gradient magnitude histograms is presented. At the initial epoch and the final epoch, the percentages of adaptive elements that are greater than the chosen highest optimal values of immutability hyperparameter (vertical dashed blue line) are $85.36 \%$  and $96.68 \%$, respectively.}
\label{subfig:Transformer} 
\end{figure*}

\begin{figure*}
\centering
\subfigure[VGG11 (trained for 45 epochs)]{\includegraphics[width=0.48\textwidth]{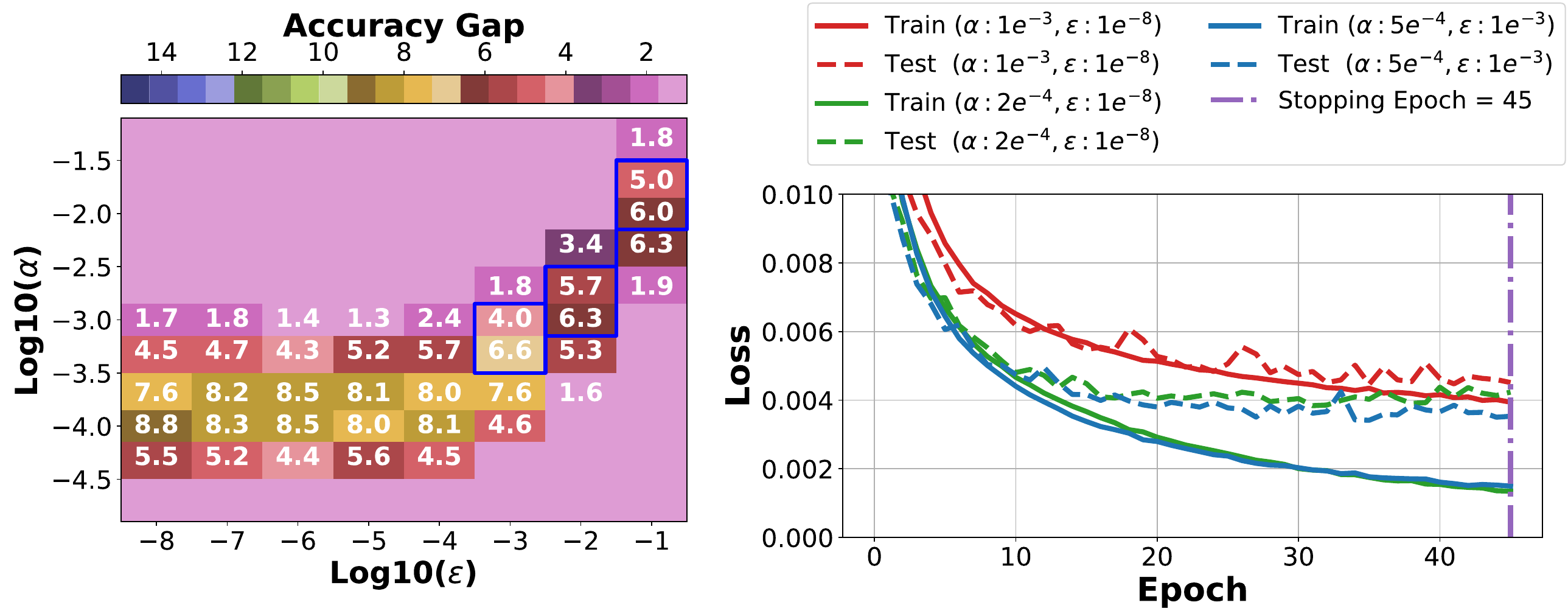}}
\subfigure[ResNet34 (trained for 60 epochs)]{\includegraphics[width=0.48\textwidth]{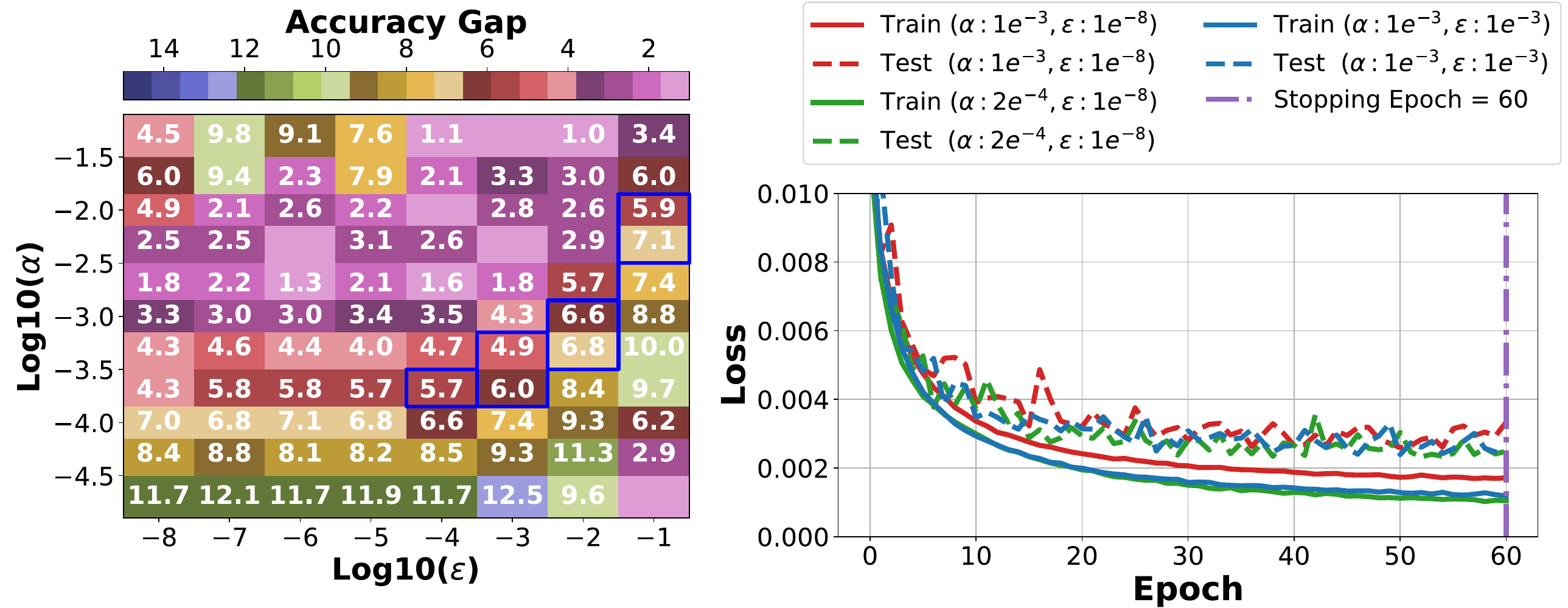}}  \\ 
\subfigure[AlexNet (trained for 25 epochs)]{\includegraphics[width=0.48\textwidth]{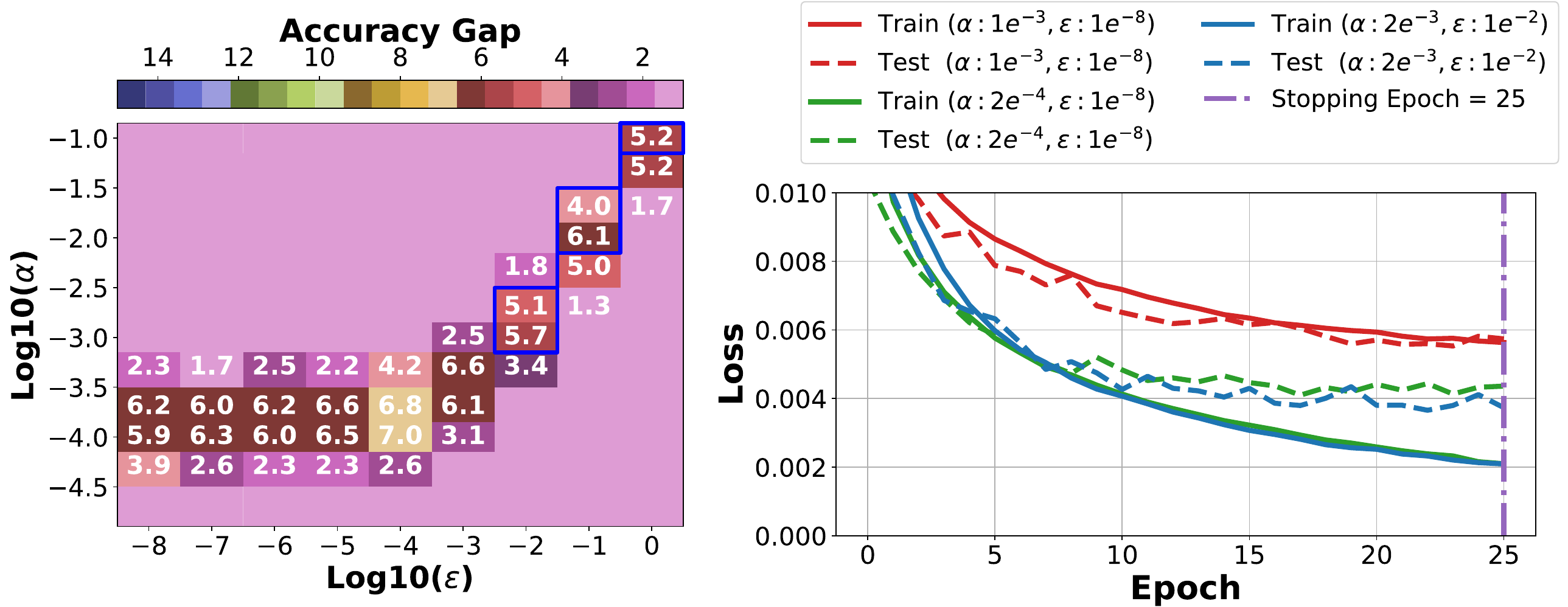}}  
\subfigure[DenseNet121  (trained for 60 epochs)]{\includegraphics[width=0.48\textwidth]{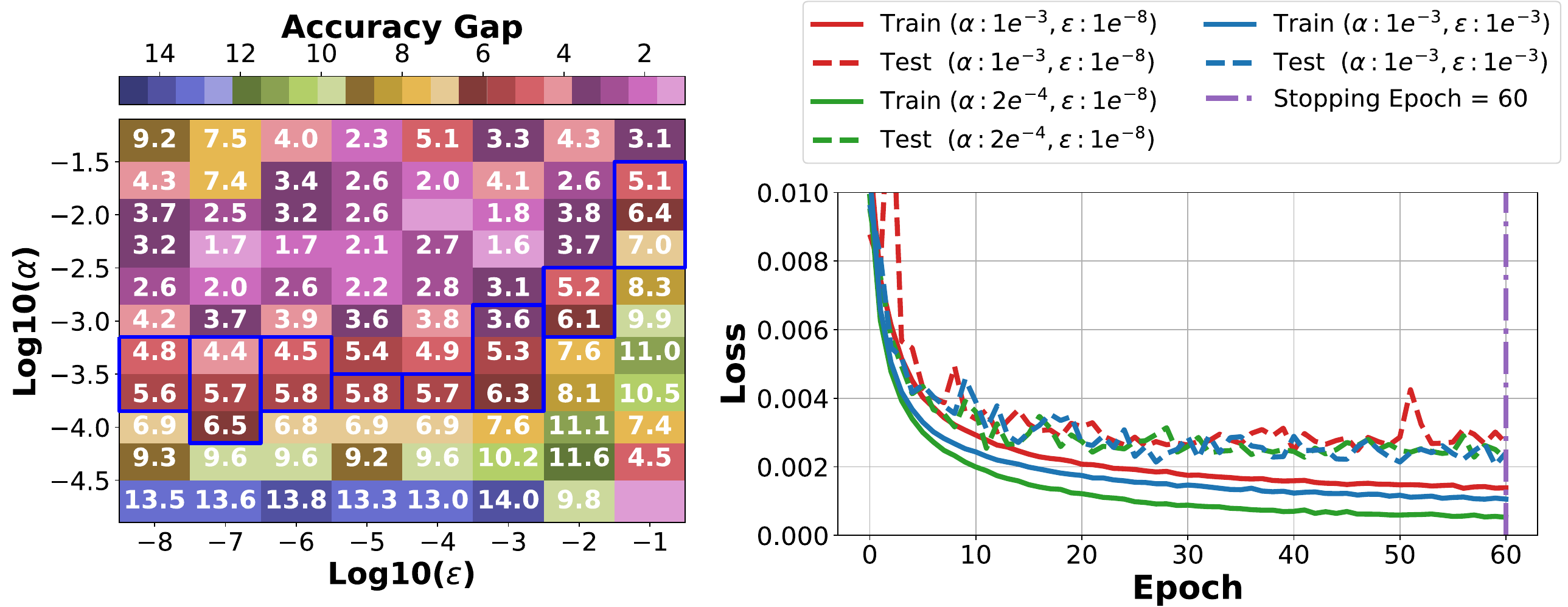}}  
\caption{VGG11, ResNet34, AlexNet and DenseNet121 on CIFAR-10 dataset with Adam optimizer, varying learning rate $\alpha$ and  immutability $\epsilon$. \textbf{(Left)} Heatmap of accuracy gap, corresponding to Fig. \ref{subfig:classif1} and Fig. \ref{subfig:lr-eps-DenseNet},  with the optimal region of  $\epsilon$ highlighted by a blue grid. \textbf{(Right)} Training and test loss progression. \label{Fig18}}
\end{figure*}

\clearpage

\setcounter{figure}{18} 
\begin{figure}[h]
\vspace{6mm}
\centering
{\includegraphics[width=0.475\textwidth]{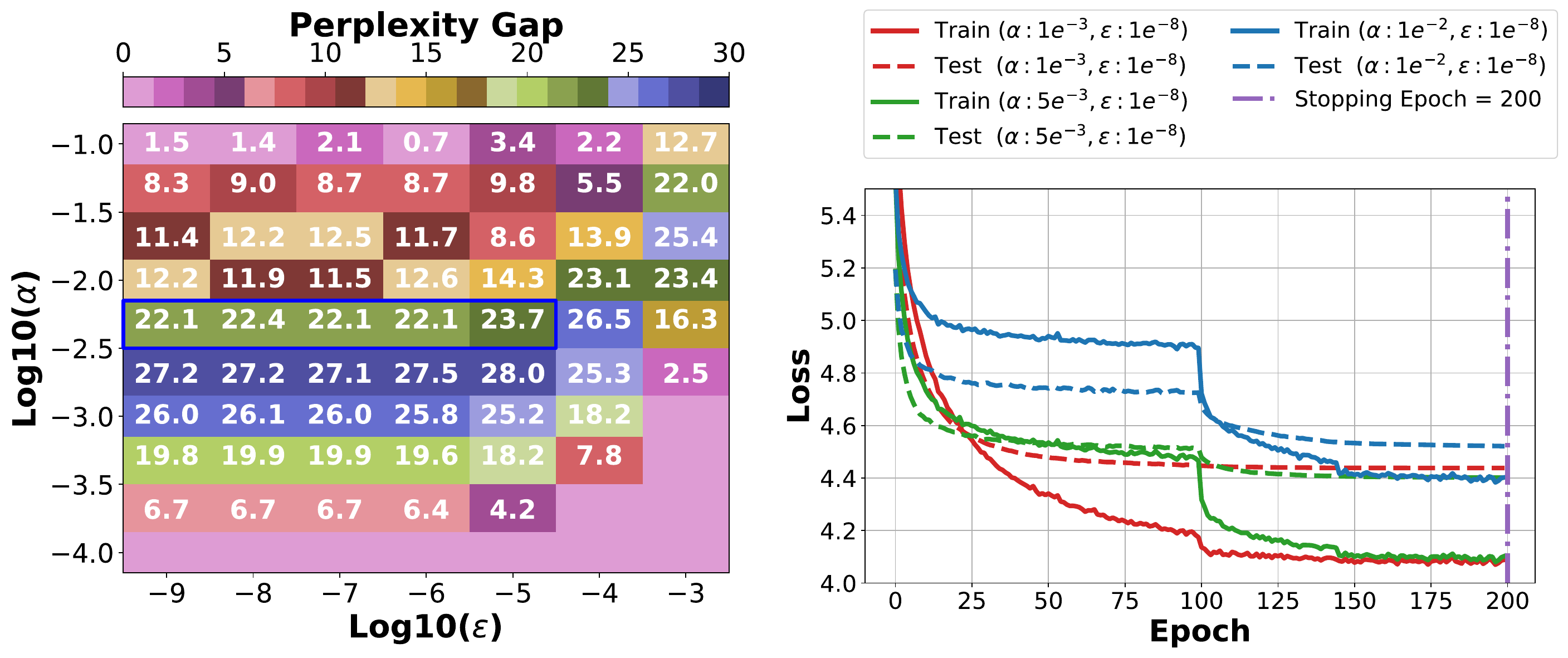}}  
\caption{1-Layer LSTM model (trained for 200 epochs) on Penn TreeBank  dataset with
Adam optimizer, varying learning rate $\alpha$ and  immutability. \textbf{(Left)} Heatmap of perplexity gap, corresponding to Fig. \ref{subfig:NLP1}, with the optimal region of  $\epsilon$ highlighted by a blue grid. \textbf{(Right)} Training and test loss progression.\label{Fig19}}
\end{figure}

\setcounter{figure}{20} 
\begin{figure}[h]
\centering
\subfigure[VGG11 (150 epochs)]{\includegraphics[width=0.22\textwidth]{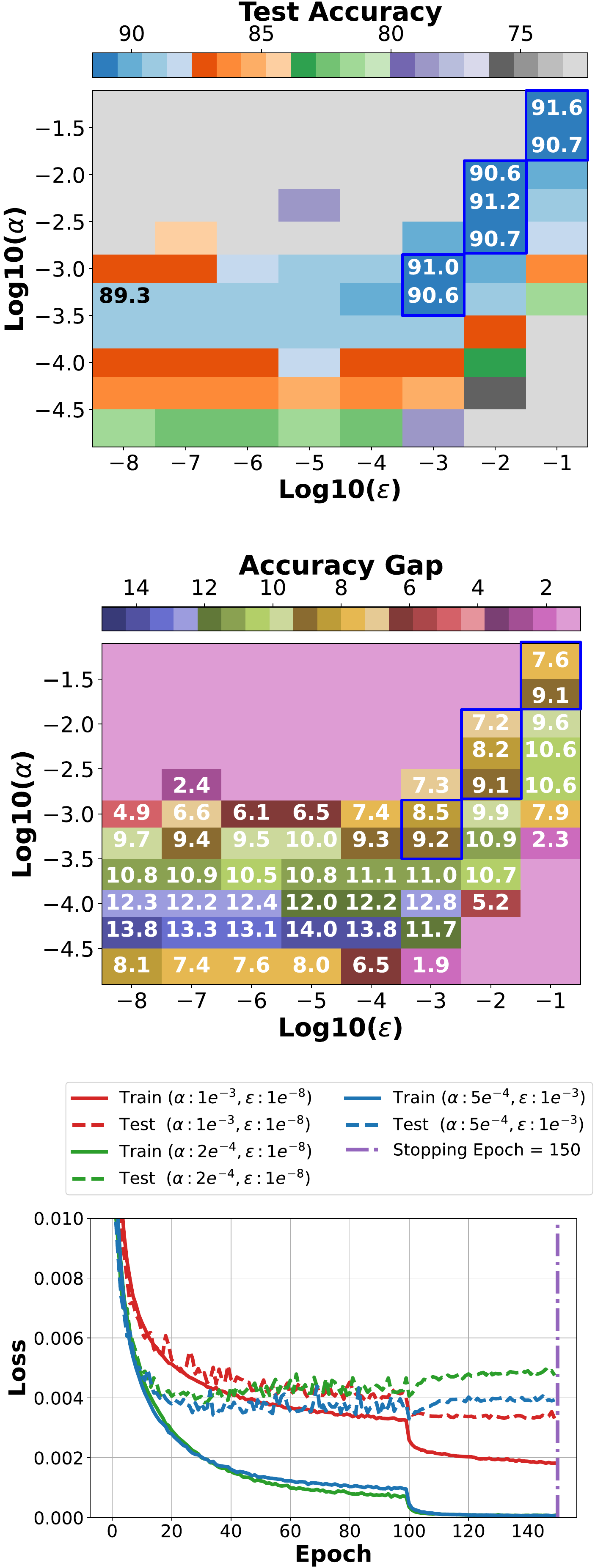} \label{Fig21a}}
\subfigure[ResNet34 (150 epochs)]{\includegraphics[width=0.22\textwidth]{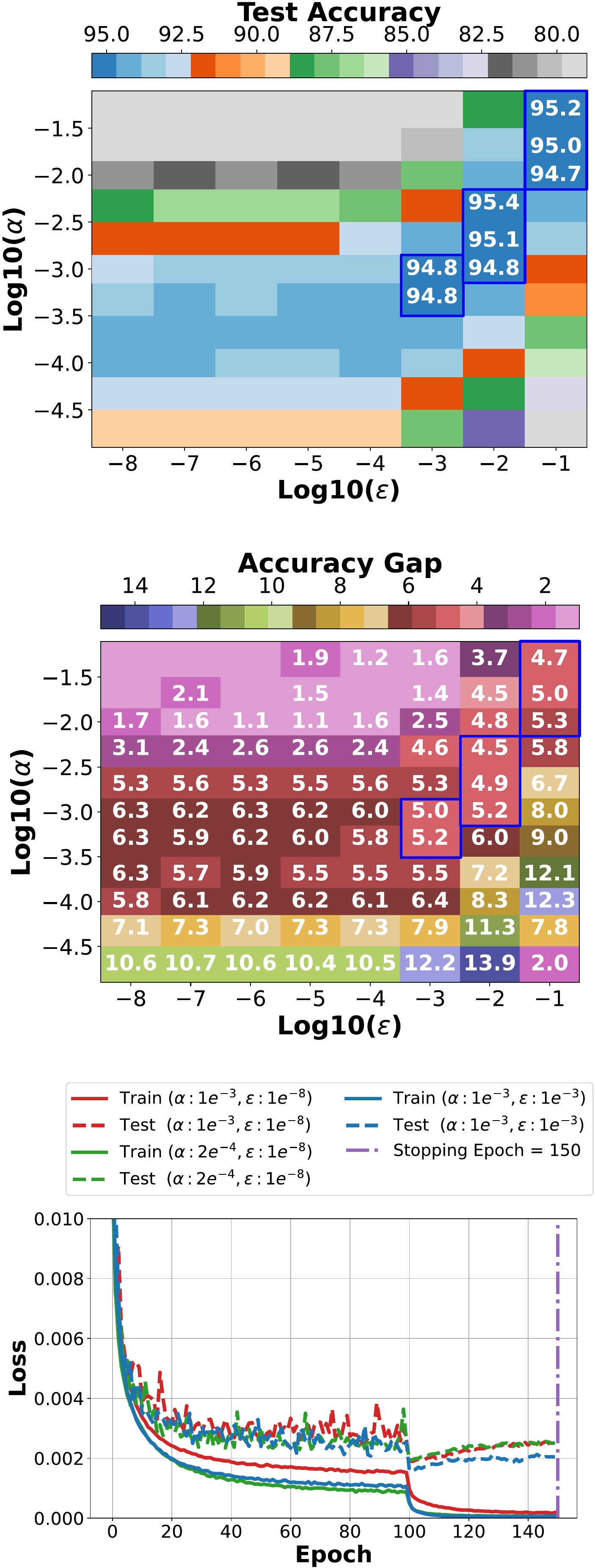}} 
\caption{VGG11 and ResNet34 (trained for 150 epochs) on CIFAR-10 dataset with
Adam optimizer, varying learning rate $\alpha$ and  immutability $\epsilon$.  Optimal region of  $\epsilon$ highlighted by a blue grid. \label{Fig21}}
\end{figure}

\setcounter{figure}{19} 
\begin{figure}[h]
\centering
{\includegraphics[width=0.475\textwidth]{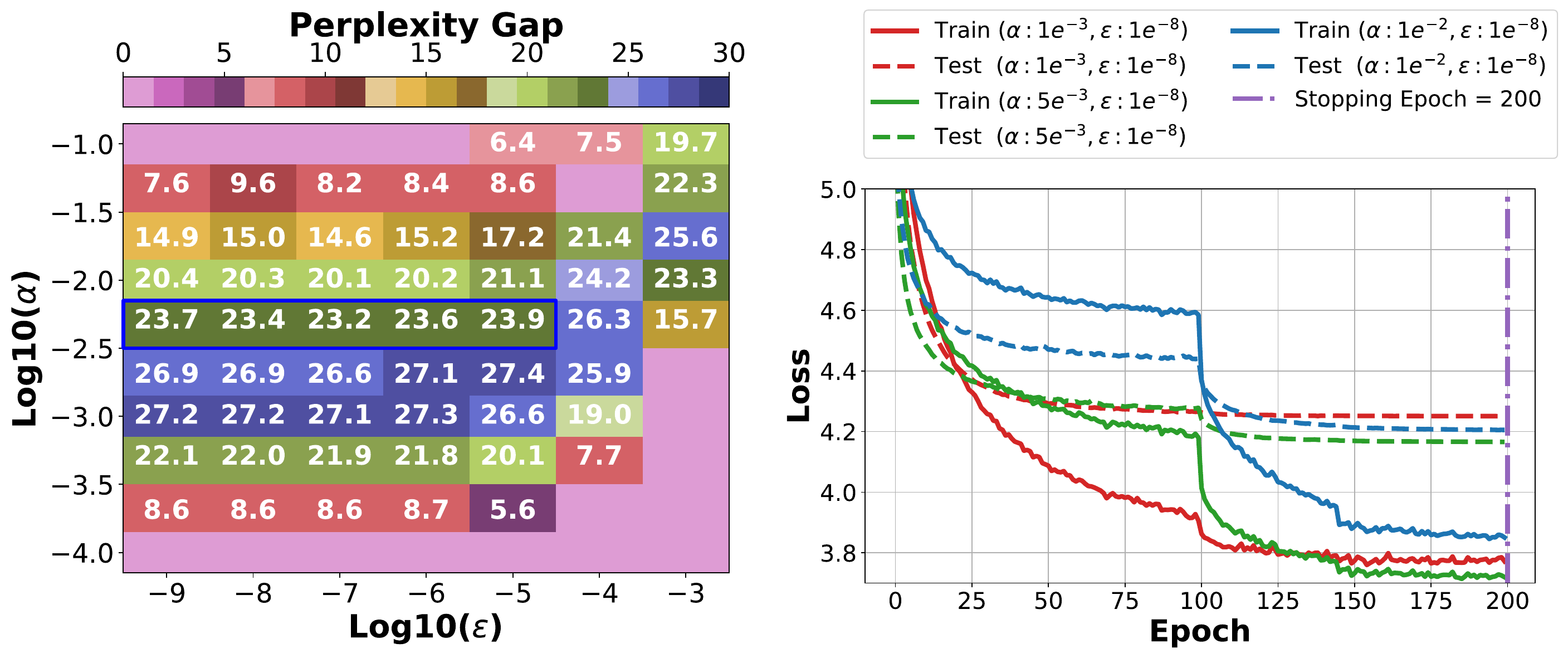}}  
\caption{2-Layer LSTM model (trained for 200 epochs) on Penn TreeBank  dataset with
Adam optimizer. \textbf{(Left)} Heatmap of perplexity gap, corresponding to Fig. \ref{subfig:NLP1}, with the optimal region of  $\epsilon$ highlighted by a blue grid. \textbf{(Right)} Training and test loss progression.\label{Fig20}}
\end{figure}

\setcounter{figure}{21} 
\begin{figure}[h]
\centering
\subfigure[VGG11]{\includegraphics[width=0.22\textwidth]{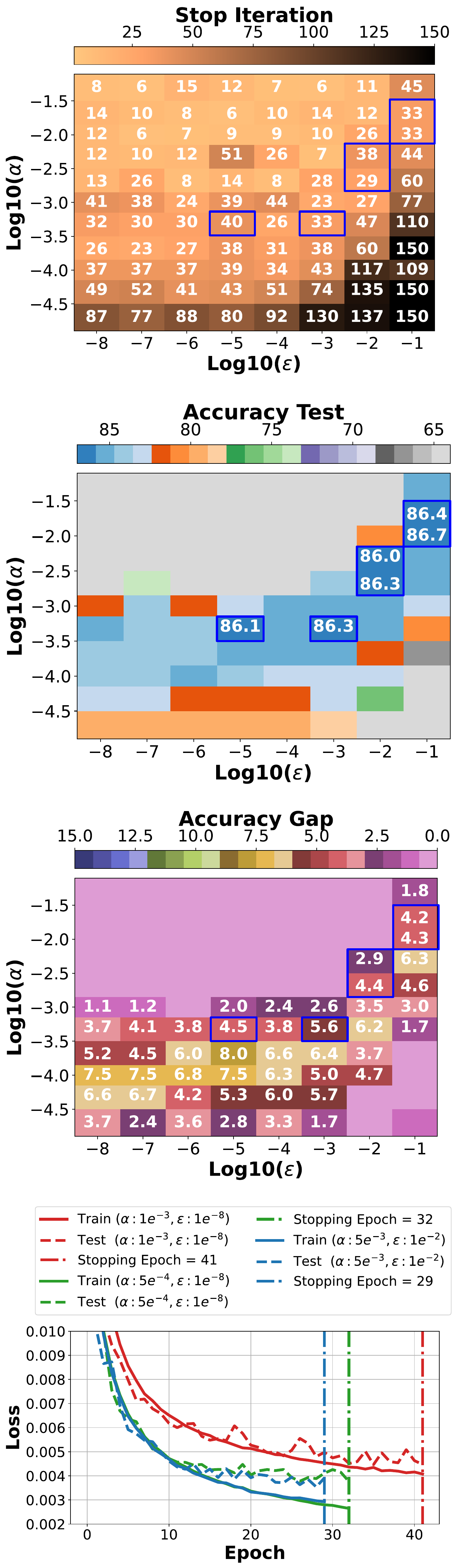}}
\subfigure[ResNet34]{\includegraphics[width=0.22\textwidth]{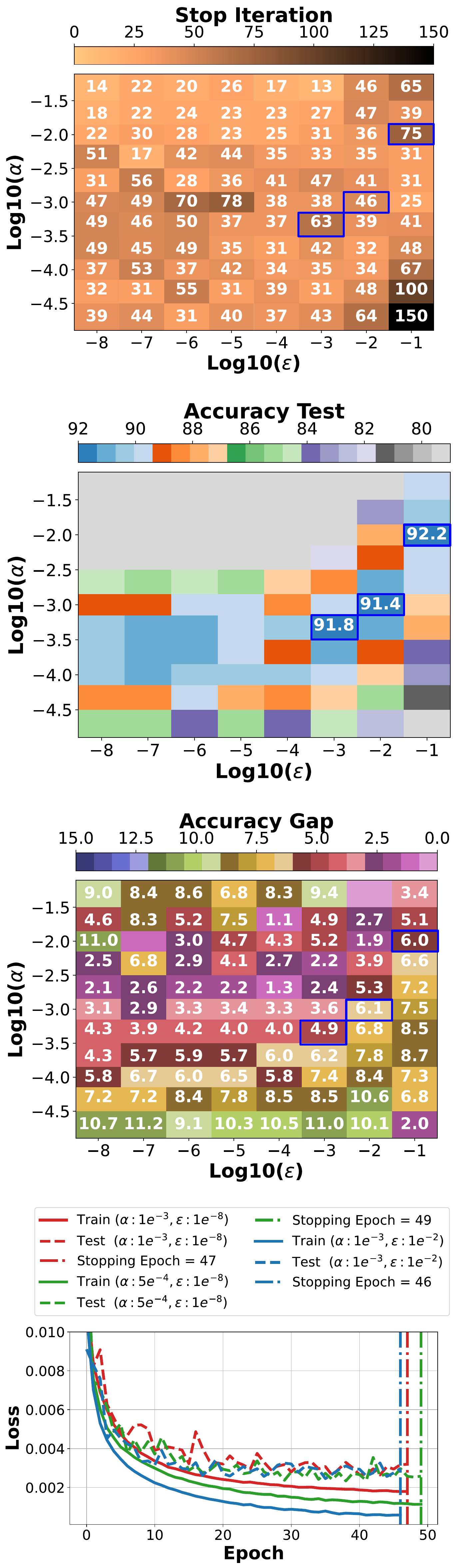}}    
\caption{VGG11 and ResNet34 trained on CIFAR-10 dataset \textbf{with early stopping} and  Adam optimizer. Optimal region of  $\epsilon$ highlighted by a blue grid.\label{Fig22}}
\end{figure}

\clearpage

\begin{figure*}
\centering
\subfigure[RMSprop]{\includegraphics[width=0.237\textwidth]{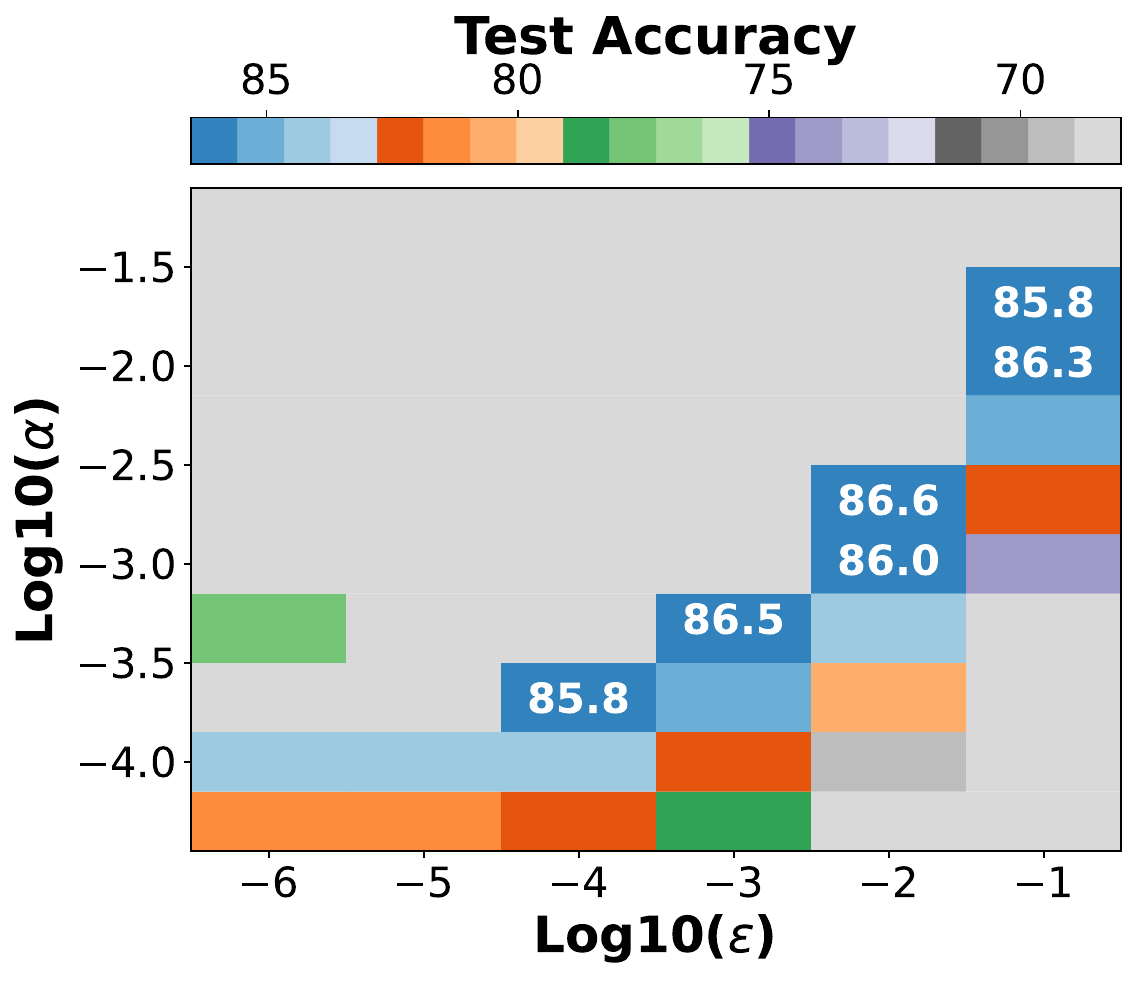} \label{subfig:fig3}} \hspace{-2mm}
\subfigure[RMSprop | From epoch 1 to epoch 45]{\includegraphics[width=0.74\textwidth]{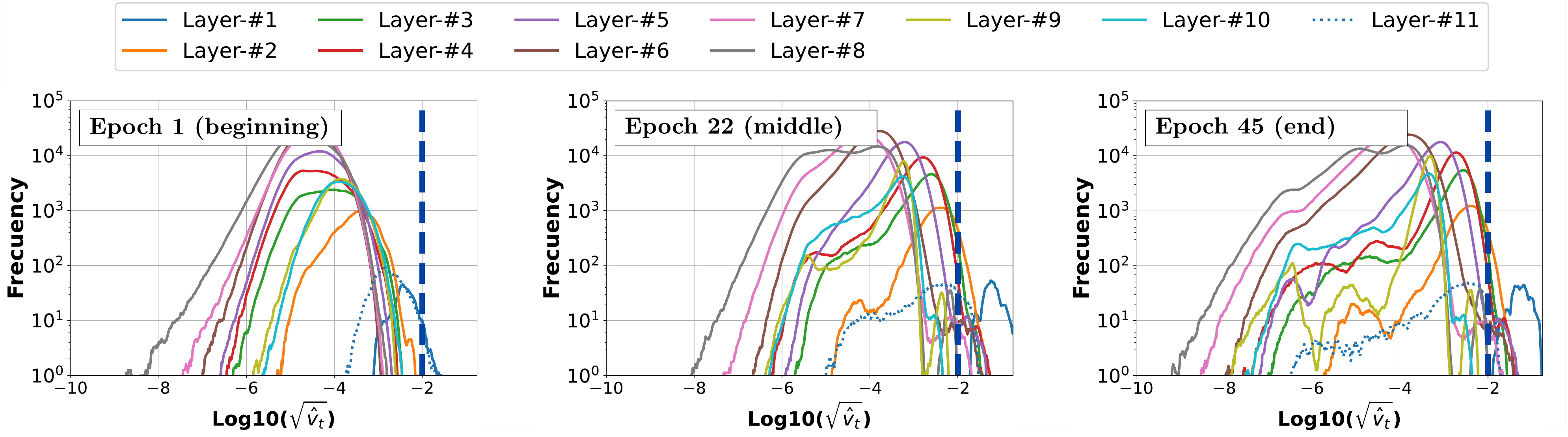} \label{subfig:fig3}} 
 \vspace{2.5mm} \\  
\subfigure[AdaBelief]{\includegraphics[width=0.237\textwidth]{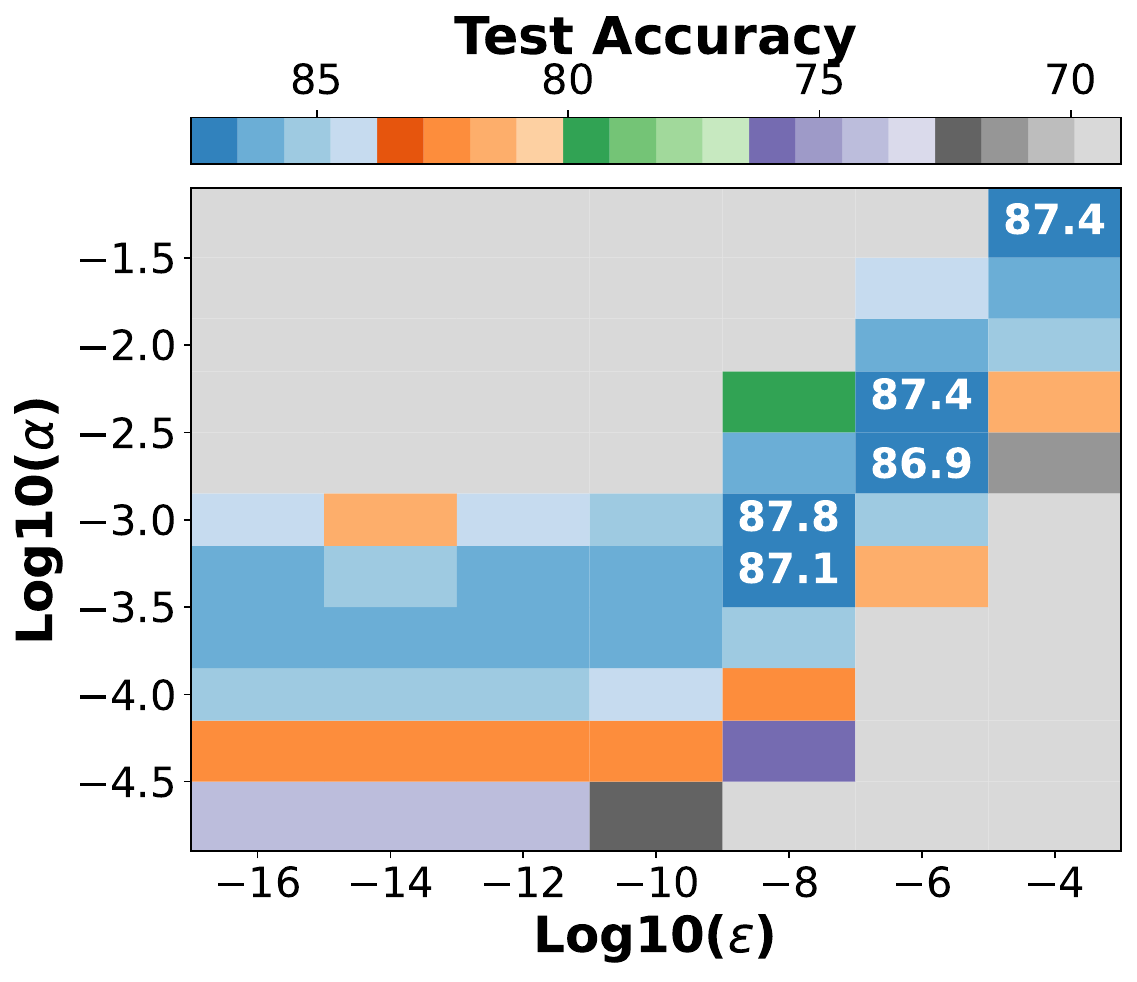} \label{subfig:fig3}} \hspace{-2mm}
\subfigure[AdaBelief | From epoch 1 to epoch 45]{\includegraphics[width=0.74\textwidth]{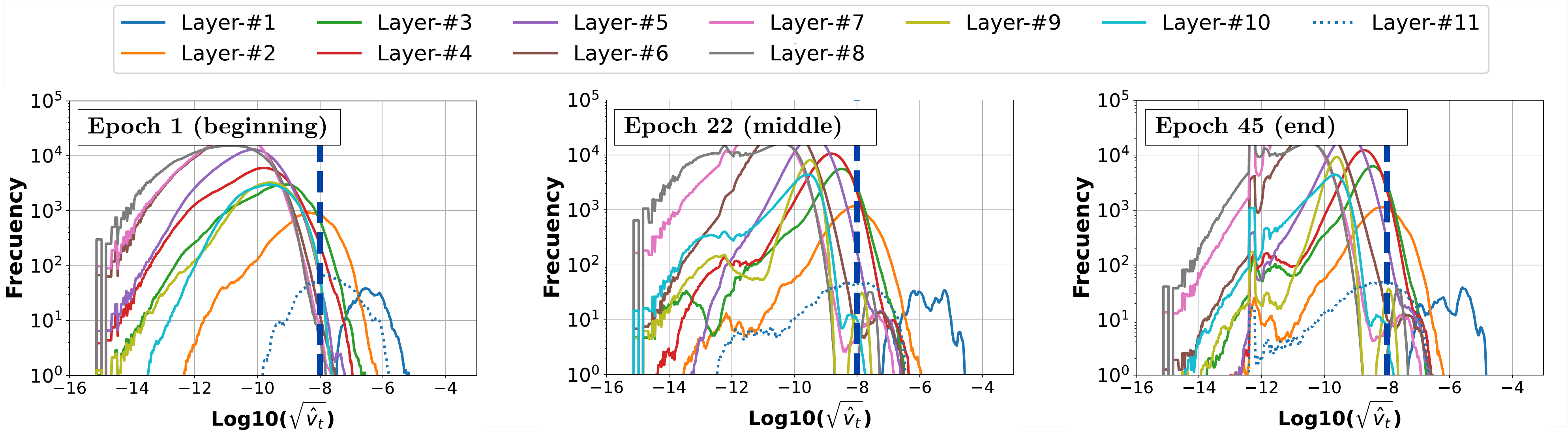} \label{subfig:fig3}} 
 \vspace{2.5mm} \\  
\subfigure[AdaMomentum]{\includegraphics[width=0.237\textwidth]{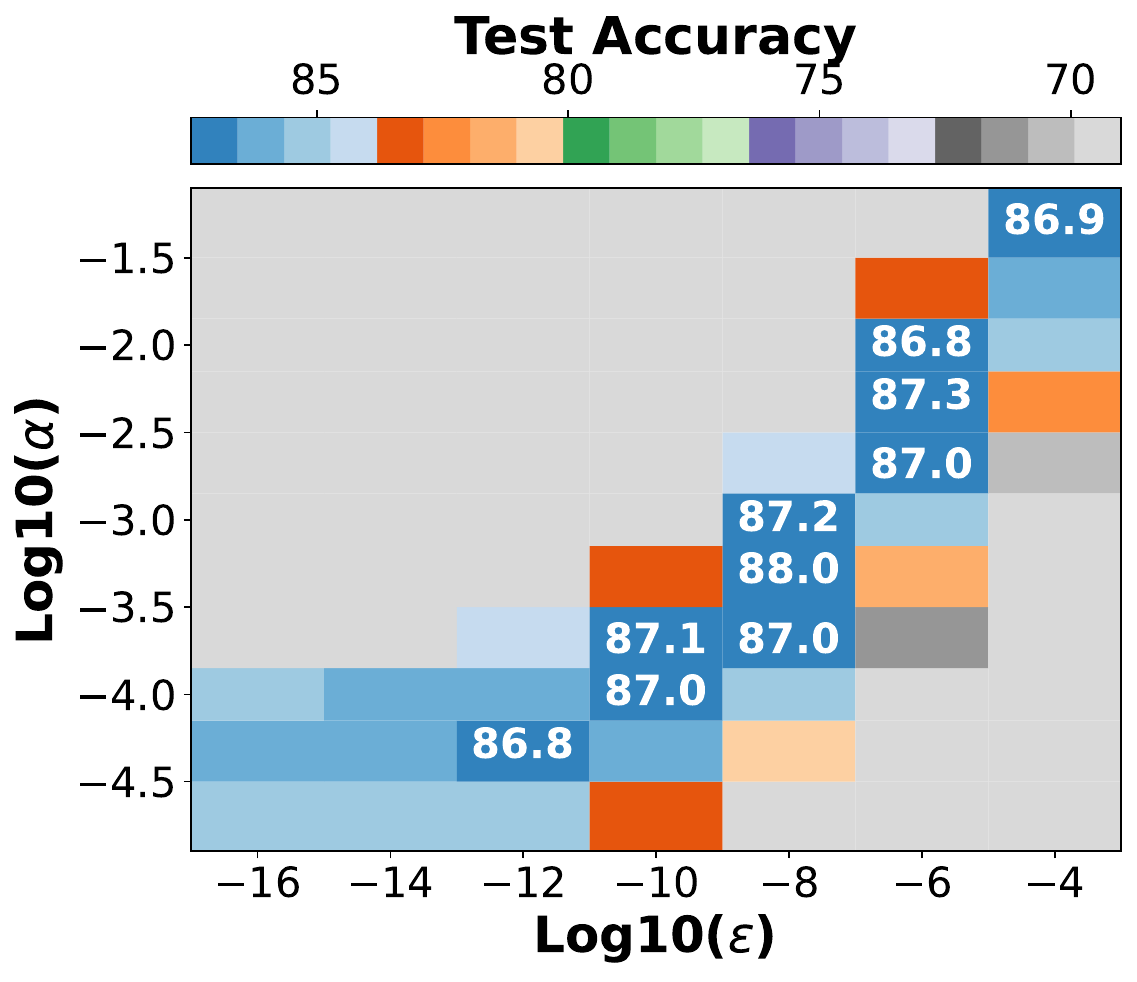} \label{subfig:fig3}}\hspace{-2mm}
\subfigure[AdaMomentum  | From epoch 1 to epoch 45]{\includegraphics[width=0.74\textwidth]{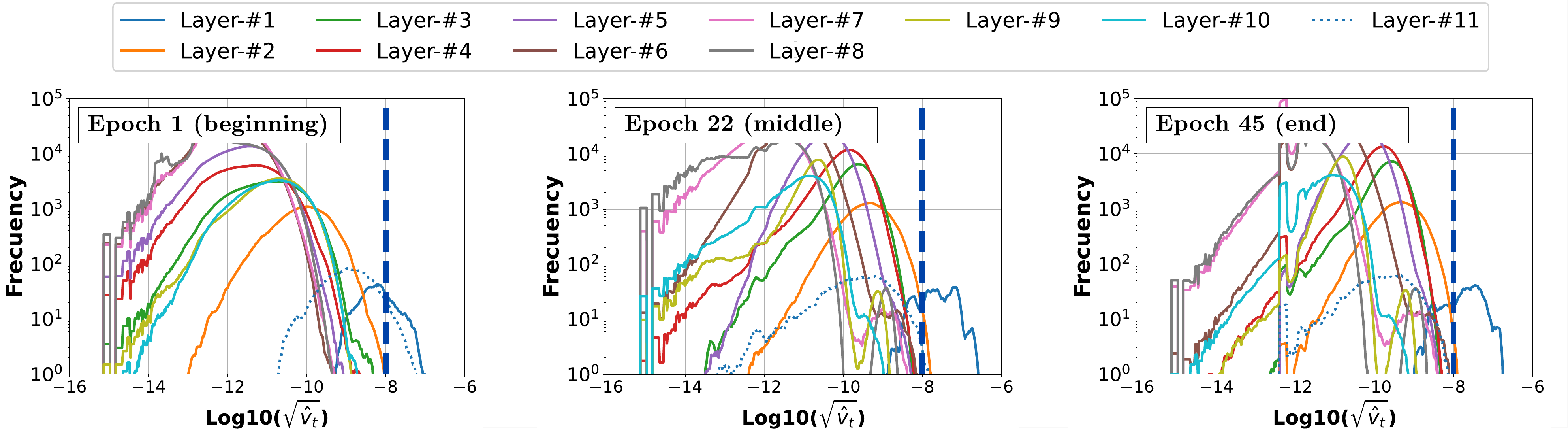} \label{subfig:fig3}} 
\\ \vspace{2mm}
\caption{Test accuracy of VGG11 classifier (trained for 45 epochs) on the CIFAR-10 dataset with RMSprop, AdaBelief and AdaMomentum optimizers. In the first column, performance is evaluated by varying learning rate hyperparameter $\alpha$ and  immutability hyperparameter $\epsilon$. In the second to fourth column, progress of gradient magnitude histograms is presented, where at first epoch $0.0051\%$, $0.45\%$ and  $0.007\%$ of adaptive elements are greater than the chosen optimal values of immutability hyperparameter (vertical dashed blue line) for the respective adaptive optimizers.\label{subfig:lr-eps-AdaBelief}}
\end{figure*}

\begin{figure*}
\vspace{4.5mm}
\centering 
\subfigure[RMSprop]{\includegraphics[width=0.237\textwidth]{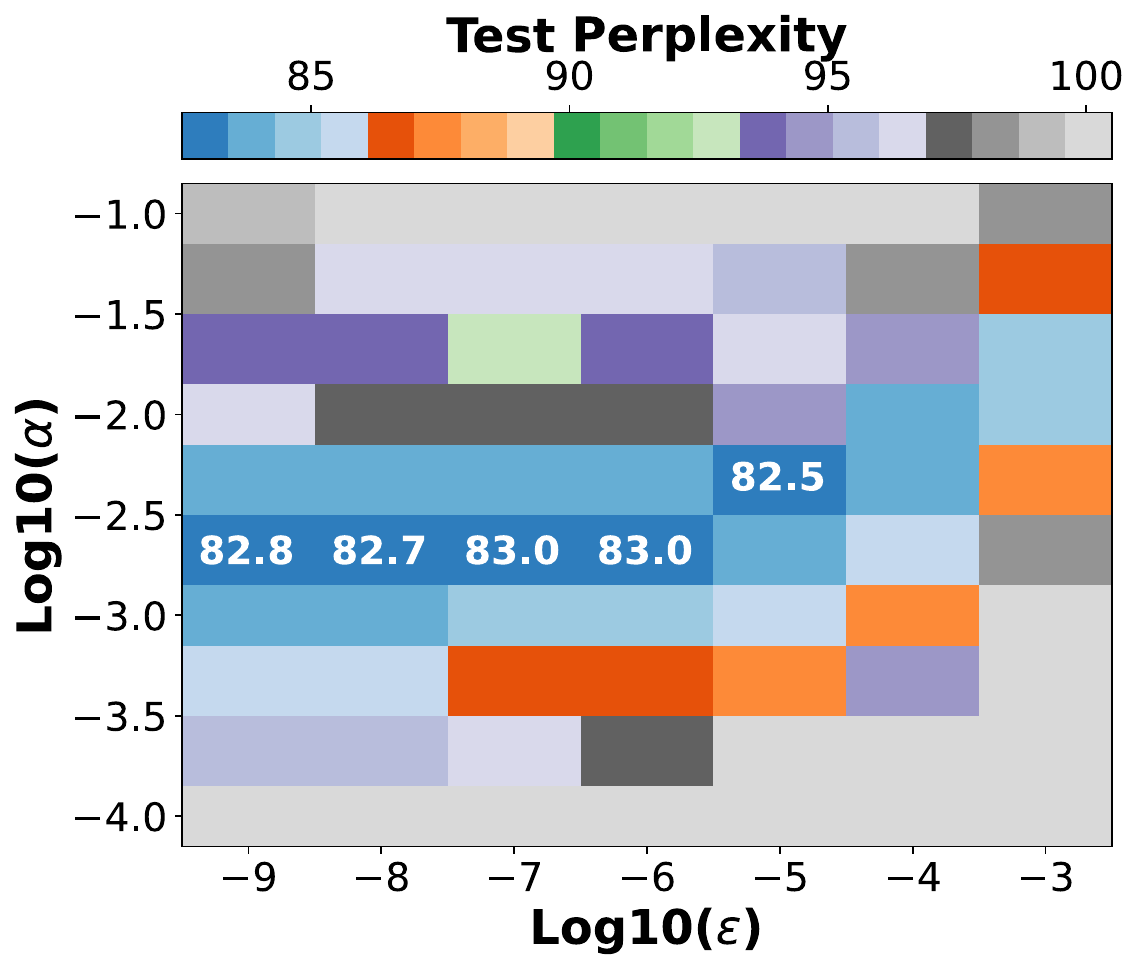} \label{subfig:NLP-RMSprop}}
\subfigure[RMSprop | From epoch 1  to epoch 200]{\includegraphics[width=0.74\textwidth]{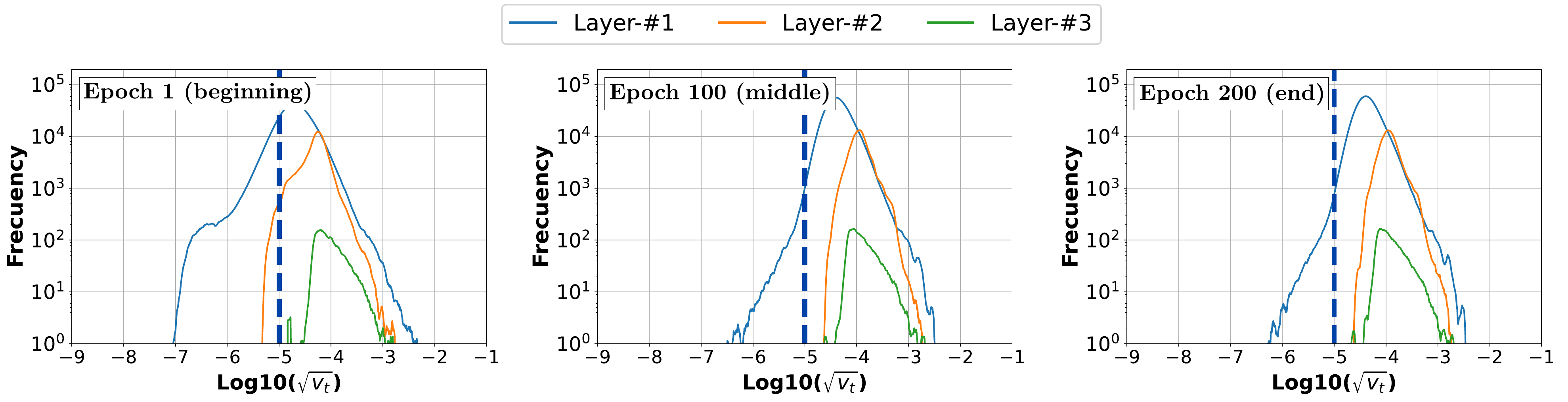} \label{subfig:fig3}} 
 \\
\subfigure[AdaBelief]{\includegraphics[width=0.237\textwidth]{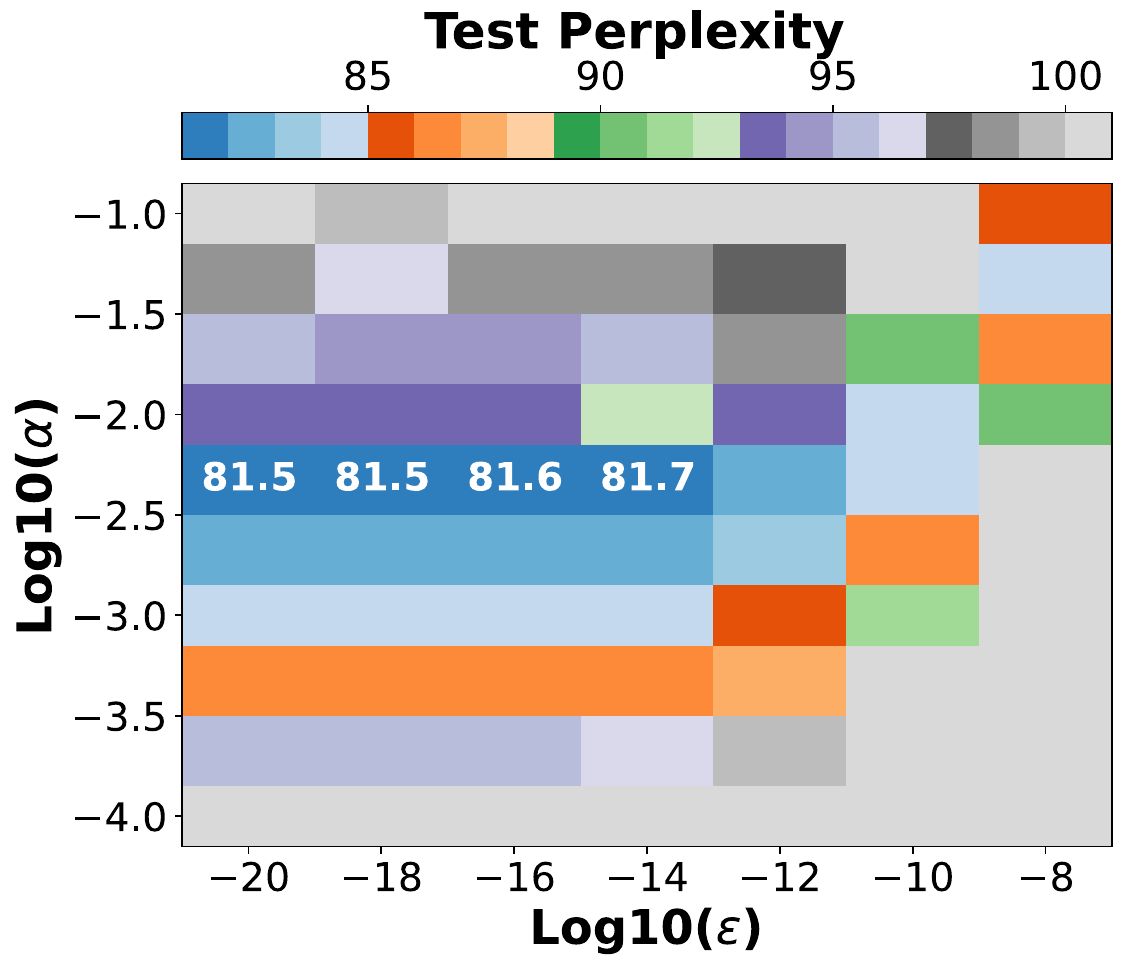} \label{subfig:NLP-AdaBelief}}
\subfigure[AdaBelief | From epoch 1  to epoch 200]{\includegraphics[width=0.74\textwidth]{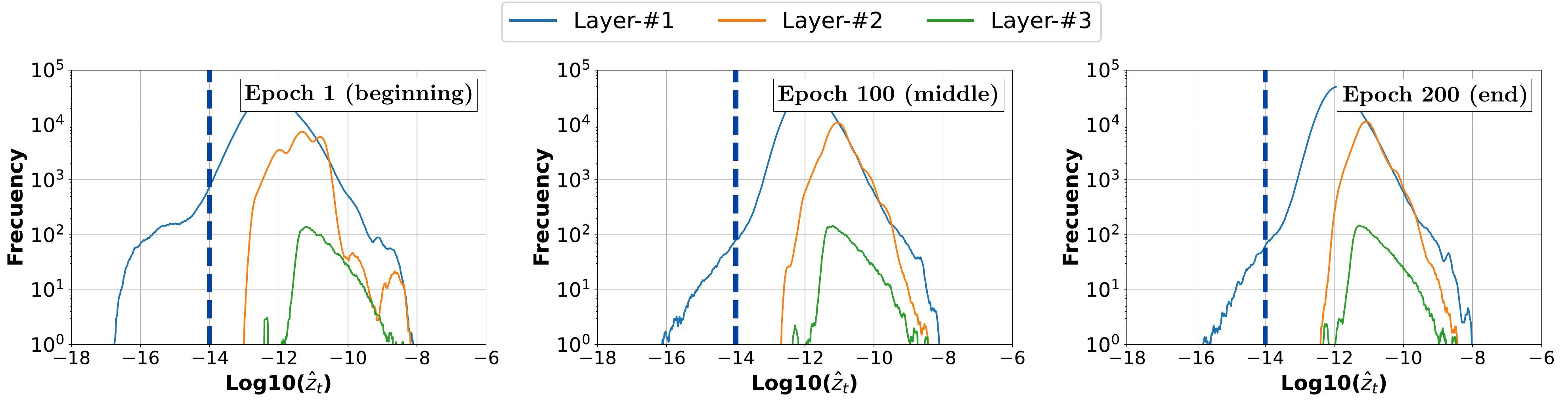} \label{subfig:fig3}} 
 \\
\subfigure[AdaMomentum]{\includegraphics[width=0.237\textwidth]{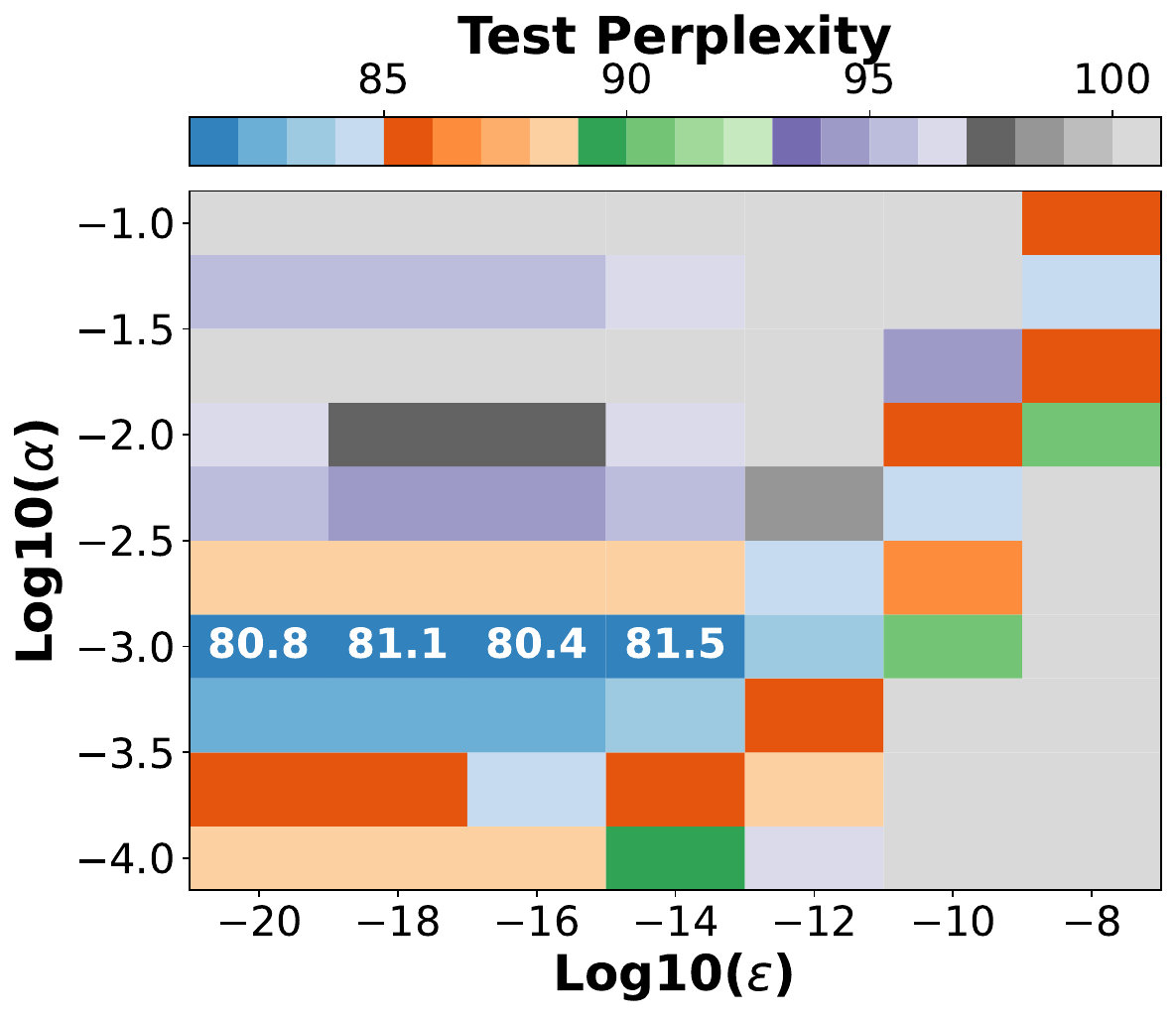} \label{subfig:NLP-AdamMom}}  
\subfigure[AdaMomentum | From epoch 1  to epoch 200]{\includegraphics[width=0.74\textwidth]{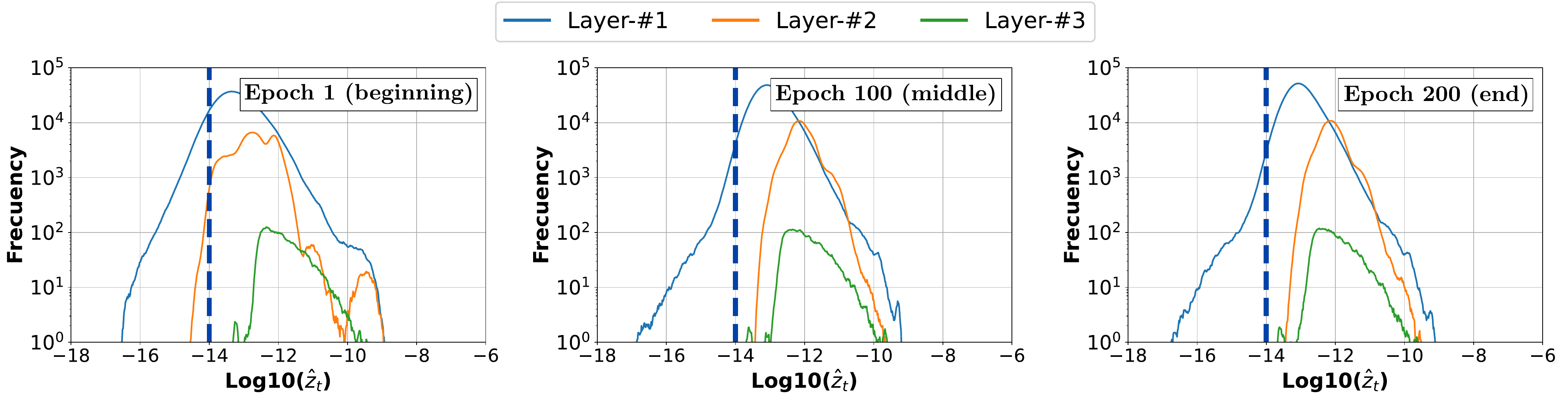} \label{subfig:fig3}} 
  \\ \vspace{2mm}
\caption{Test perplexity of 1-Layer LSTM model (trained for 200 epochs) on the Penn TreeBank dataset with RMSprop, AdaBelief and AdaMomentum.  In the first column, performance is evaluated by varying learning rate hyperparameter $\alpha$ and  immutability hyperparameter $\epsilon$, where lower perplexity  corresponds to better performance. In the second to fourth column, progress of gradient magnitude histograms is presented, where at first epoch $85.73\%$, $99.31\%$ and  $91.42\%$ of adaptive elements are greater than the chosen highest optimal values of immutability hyperparameter (vertical dashed blue line) for the respective adaptive optimizers.\label{subfig:NLP-optimizer}} 
\end{figure*}

\clearpage

\onecolumn
\begin{table}
\vspace{-1.5mm}
\setlength\extrarowheight{5pt}
\addtolength{\tabcolsep}{-3pt}
\caption{Summary of adaptive learning rates. For the chosen value of hyperparameter $\epsilon$ is contemplated two extreme cases (fully adaptable and fully immutable), where each element of estimated adaptive learning rate $\alpha_t$ can be self-adjusting or approximately constant.  \label{tab:FullCases}
\vspace{1.5mm}}
\centering
\footnotesize
\begin{tabular}{c|l|l|l}
\hline 
$\textbf{Optimizer}$  & $~~\textbf{Adaptive Learning Rate}$  & $~~\textbf{Fully Adaptable Case}$  & ~~$\textbf{Fully Immutable Case}$\tabularnewline [1ex]
\hline 
\hline 
$\text{\textbf{AdaGrad}}$ & $\begin{array}{l}
v_{t}=v_{t-1}+g_{t}^{2}\\
\alpha_{t}=\dfrac{\alpha}{\sqrt{v_{t}}+\epsilon} \\ 
\end{array}$ & $\begin{array}{l}
\hat{z}_t = \sqrt{v_{t}} \\
\text{If}~~\big(\hat{z}_t \gg\epsilon\big):
~~~\alpha_{t}\approx\dfrac{\alpha}{\sqrt{v_{t}}} \vspace{1.25mm}
\end{array}$ & $\begin{array}{l}
\hat{z}_t = \sqrt{v_{t}} \\
\text{If}~~\big(\hat{z}_{t}\ll\epsilon\big):
~~~\alpha_{t}\approx\dfrac{\alpha}{\epsilon} \vspace{1.25mm}
\end{array}$\tabularnewline
\hline 
$\text{\textbf{RMSprop}}$ & $\begin{array}{l}
v_{t}=\beta\cdot v_{t-1}+(1-\beta)\cdot g_{t}^{2}\\
\alpha_{t}=\dfrac{\alpha}{\sqrt{v_{t}}+\epsilon}  \\ 
\end{array}$ & $\begin{array}{l}
\hat{z}_t = \sqrt{v_{t}} \\
\text{If}~~\big(\hat{z}_{t}\gg\epsilon\big):~~~\alpha_{t}\approx\dfrac{\alpha}{\sqrt{v_{t}}}  \vspace{1.25mm}
\end{array}$ & $\begin{array}{l}
\hat{z}_t = \sqrt{v_{t}} \\
\text{If}~~\big(\hat{z}_{t}\ll\epsilon\big):~~~\alpha_{t}\approx\dfrac{\alpha}{\epsilon}  \vspace{1.25mm}
\end{array}$\tabularnewline
\hline 
$\text{\textbf{Adam}}$ & $\begin{array}{l}
v_{t}=\beta_{2}\cdot v_{t-1}+(1-\beta_{2})\cdot g_{t}^{2}\\
\hat{{v}}_{t}=v_{t}/(1-\beta_{2}^{t})\\
\alpha_{t}=\dfrac{\alpha}{\sqrt{\hat{{v}_{t}}}+\epsilon}  \vspace{1.25mm}
\end{array}$ & $\begin{array}{l}
\hat{z}_t = \sqrt{\hat{v}_{t}} \\
\text{If}~~\big(\hat{z}_{t}\gg\epsilon\big):\\
~~~~~~~\alpha_{t}\approx\dfrac{\alpha}{\sqrt{\hat{{v}_{t}}}}
\end{array}$ & $\begin{array}{l}
\hat{z}_t = \sqrt{\hat{v}_{t}} \\
\text{If}~~\big(\hat{z}_{t}\ll\epsilon\big):\\
~~~~~~~\alpha_{t}\approx\dfrac{\alpha}{\epsilon}
\end{array}$\tabularnewline
\hline 
$\text{\textbf{DiffGrad}}$ & $\begin{array}{l}
\xi_{t}=1/(1+e^{-|g_{t-1}-g_{t}|}),\\
\text{where }0.5\leq\xi_{t}\leq1\\
v_{t}=\beta_{2}\cdot v_{t-1}+(1-\beta_{2})\cdot g_{t}^{2}\\
\hat{{v}}_{t}=v_{t}/(1-\beta_{2}^{t})\\
\alpha_{t}=\dfrac{\alpha\cdot\xi_{t}}{\sqrt{\hat{{v}_{t}}}+\epsilon} \vspace{1.25mm}
\end{array}$ & $\begin{array}{l}
\hat{z}_t = \sqrt{\hat{v}_{t}} \\
\text{If}~~\big(\hat{z}_{t}\gg\epsilon\big):\vspace{1.25mm}\\
~~~~~~~\alpha_{t}\approx\dfrac{\alpha\cdot\xi_{t}}{\sqrt{\hat{{v}_{t}}}}\vspace{2mm}\\
\\
\end{array}$ & $\begin{array}{l}
\hat{z}_t = \sqrt{\hat{v}_{t}} \\
\text{If}~~\big(\hat{z}_{t}\ll\epsilon\big): \vspace{1.25mm}\\
~~~~~~~\dfrac{\alpha}{2\cdot\epsilon}\leq\alpha_{t}\leq\dfrac{\alpha}{\epsilon}, \vspace{2mm}\\
~~~~~~~\text{where \ensuremath{\alpha_{t}\approx\dfrac{\alpha\cdot\xi_{t}}{\epsilon}}}
\end{array}$\tabularnewline
\hline 
$\textbf{AdaMod}$ & $\begin{array}{l}
v_{t}=\beta_{2}\cdot v_{t-1}+(1-\beta_{2})\cdot g_{t}^{2}\\
\hat{{v}}_{t}=v_{t}/(1-\beta_{2}^{t})\\
n_{t}=\dfrac{\alpha}{\sqrt{\hat{{v}_{t}}}+\epsilon}\\
s_{t}=\beta_{3}\cdot s_{t-1}+(1-\beta_{3})\cdot n_{t}\\
\alpha_{t}=\min\{n_{t},s_{t}\}  \vspace{1.25mm}
\end{array}$ & $\begin{array}{l}
\hat{z}_t = \sqrt{\hat{v}_{t}} \\
\text{If}~~\big(\hat{z}_{t}\gg\epsilon\big):\\
~~~~~~~n_{t}\approx\dfrac{\alpha}{\sqrt{\hat{{v}_{t}}}}\\
~~~~~~~s_{t}=\beta_{3}\cdot s_{t-1}+(1-\beta_{3})\cdot n_{t}\\
~~~~~~~\alpha_{t}=\min\{n_{t},s_{t}\}
\end{array}$ & $\begin{array}{l}
\hat{z}_t = \sqrt{\hat{v}_{t}} \\
\text{If}~~\big(\hat{z}_{t}\ll\epsilon\big):\\
~~~~~~~n_{t}\approx\dfrac{\alpha}{\epsilon}\\
~~~~~~~\alpha_{t}=(1-\beta_{3}^{t})\cdot\dfrac{\alpha}{\epsilon}\\
\\
\end{array}$\tabularnewline
\hline 
$\text{\textbf{AdaBelief}}$\protect\footnotemark & $\begin{array}{l}
v_{t}=\beta_{2}\cdot v_{t-1}+(1-\beta_{2})\cdot(g_{t}-m_{t})^{2}\\
s_{t}=v_t+ (1-\beta_2^t) \cdot \epsilon / (1-\beta_2)\\
\hat{{s}}_{t}=s_{t}/(1-\beta_{2}^{t})= \hat{{v}}_{t} + \epsilon / (1-\beta_2)\\
\alpha_{t}=\dfrac{\alpha}{\sqrt{\hat{{s}_{t}}}+\epsilon}  \vspace{1.25mm}
\end{array}$ & $\begin{array}{l}
\hat{z}_t = (1-\beta_2) \cdot \hat{v}_t \\
\text{If}~~\big(\hat{z}_{t}\gg\epsilon\big):\\
~~~~~~~s_{t}\approx\beta_{2}\cdot s_{t-1}+(1-\beta_{2})\cdot(g_{t}-m_{t})^{2}\\
~~~~~~~\alpha_{t}\approx\dfrac{\alpha}{\sqrt{\hat{{s}_{t}}}}  \vspace{1.25mm}
\end{array}$ & $\begin{array}{l}
\hat{z}_t = (1-\beta_2) \cdot \hat{v}_t \\
\text{If}~~\big(\hat{z}_{t}\ll\epsilon \big):\\
~~~~~~~s_{t}\approx\beta_{2}\cdot s_{t-1}+\epsilon\\
~~~~~~~\alpha_{t}\approx\dfrac{\alpha}{\sqrt{\epsilon/(1-\beta_{2})}+\epsilon}  \vspace{1.25mm}
\end{array}$\tabularnewline
\hline 
$\textbf{MADGRAD}$ & $\begin{array}{l}
\lambda_{t}=\alpha\sqrt{t+1}\\
v_{t}=v_{t-1}+\lambda_{t}\cdot g_{t}^{2}\\
\alpha_{t}=\dfrac{1}{\sqrt[3]{v_{t}}+\epsilon} \vspace{1.25mm}
\end{array}$ & $\begin{array}{l}
\hat{z}_t =  \sqrt[3]{v_{t}} \\
\text{If}~~\big(\hat{z}_{t}\gg\epsilon\big):\vspace{1.25mm}\\
~~~~~~~\alpha_{t}\approx\dfrac{1}{\sqrt[3]{v_{t}}} \vspace{1.25mm}
\end{array}$ & $\begin{array}{l}
\hat{z}_t = \sqrt[3]{v_{t}} \\
\text{If}~~\big(\hat{z}_{t}\ll\epsilon\big):\vspace{1.25mm}\\
~~~~~~~\alpha_{t}\approx\dfrac{1}{\epsilon}
\end{array}$\tabularnewline
\hline 
$\begin{array}{c}
\textbf{EAdam}\footref{refnote}
\end{array}$ & $\begin{array}{l}
v_{t}=\beta_{2}\cdot v_{t-1}+(1-\beta_{2})\cdot g_{t}^{2}\\
s_{t}=v_t+ (1-\beta_2^t) \cdot \epsilon / (1-\beta_2)\\
\hat{{s}}_{t}=s_{t}/(1-\beta_{2}^{t}) = \hat{{v}}_{t} + \epsilon / (1-\beta_2)\\
\alpha_{t}=\dfrac{\alpha}{\sqrt{\hat{{s}_{t}}}} \vspace{1.25mm}
\end{array}$ & $\begin{array}{l}
\hat{z}_t = (1-\beta_2) \cdot \hat{v}_t \\
\text{If}~~\big(\hat{z}_{t}\gg\epsilon \big):\\
~~~~~~~s_{t}\approx\beta_{2}\cdot s_{t-1}+(1-\beta_{2})\cdot g_{t}^{2}\\
~~~~~~~\alpha_{t}\approx\dfrac{\alpha}{\sqrt{\hat{{s}_{t}}}} \vspace{1.25mm}
\end{array}$ & $\begin{array}{l}
\hat{z}_t = (1-\beta_2) \cdot \hat{v}_t \\
\text{If}~~\big(\hat{z}_{t}\ll\epsilon\big):\\
~~~~~~~s_{t}\approx\beta_{2}\cdot s_{t-1}+\epsilon\\
~~~~~~~\alpha_{t}\approx\dfrac{\alpha}{\sqrt{\epsilon/(1-\beta_{2})}}\vspace{1.25mm}
\end{array}$\tabularnewline
\hline 
$\begin{array}{c}
\textbf{AdaMo-} \vspace{-1.5mm}\\
\textbf{mentum}\footref{refnote}
\end{array}$ & $\begin{array}{l}
v_{t}=\beta_{2}\cdot v_{t-1}+(1-\beta_{2})\cdot m_{t}^{2}\\
s_{t}=v_t+ (1-\beta_2^t) \cdot \epsilon / (1-\beta_2)\\
\hat{{s}}_{t}=s_{t}/(1-\beta_{2}^{t}) = \hat{{v}}_{t} + \epsilon / (1-\beta_2)\\
\alpha_{t}=\dfrac{\alpha}{\sqrt{\hat{{s}_{t}}}} \vspace{1.25mm}
\end{array}$ & $\begin{array}{l}
\hat{z}_t = (1-\beta_2) \cdot \hat{v}_t \\
\text{If}~~\big(\hat{z}_{t}\gg\epsilon\big):\\
~~~~~~~s_{t}\approx\beta_{2}\cdot s_{t-1}+(1-\beta_{2})\cdot m_{t}^{2}\\
~~~~~~~\alpha_{t}\approx\dfrac{\alpha}{\sqrt{\hat{{s}_{t}}}} \vspace{1.25mm}
\end{array}$ & $\begin{array}{l}
\hat{z}_t = (1-\beta_2) \cdot \hat{v}_t \\
\text{If}~~\big(\hat{z}_{t}\ll\epsilon\big):\\
~~~~~~~s_{t}\approx\beta_{2}\cdot s_{t-1}+\epsilon\\
~~~~~~~\alpha_{t}\approx\dfrac{\alpha}{\sqrt{\epsilon/(1-\beta_{2})}}\vspace{1.25mm}
\end{array}$\tabularnewline
\hline 
\end{tabular}
\vspace{-35mm}
\end{table}
\footnotetext{Equations of AdaBelief, Eadam and AdaMomentum were rearranged in order to provide a general algorithm, presented in Section \ref{optimal-search-space}, that computes a search range for the immutability hyperparameter $\epsilon$ of many adaptive optimizers.\label{refnote}}

\twocolumn

\end{document}